\documentclass[10pt,twocolumn,letterpaper]{article}

\usepackage{cvpr}              
\usepackage{comment}
\usepackage{tabu}

\definecolor{cvprblue}{rgb}{0.21,0.49,0.74}
\usepackage[pagebackref,breaklinks,colorlinks,allcolors=cvprblue]{hyperref}

\def\paperID{6058} 
\def\confName{CVPR}
\def\confYear{2025}

\title{Multi-Resolution Pathology-Language Pre-training Model with \\ Text-Guided Visual Representation}

\author{Shahad Albastaki$^{1}$,~Anabia Sohail$^{1}$,~Iyyakutti Iyappan Ganapathi$^{1}$,~Basit Alawode$^{1}$,\\~Asim Khan$^{1}$,~Sajid Javed$^{1,*}$,~Naoufel Werghi$^{1}$,~Mohammed Bennamoun$^{3}$,~Arif Mahmood$^{2}$\\$^{1}$Department of Computer Science, $^{*}$ARIC, Khalifa University of Science and Technology, UAE\\$^{2}$Information Technology University of the Punjab, Pakistan,~$^{3}$University of the Western Australia\\
}

\begin{document}
\maketitle
\vspace{-2mm}
\begin{abstract}
In Computational Pathology (CPath), the introduction of Vision-Language Models (VLMs) has opened new avenues for research, focusing primarily on aligning image-text pairs at a single magnification level. 
However, this approach might not be sufficient for tasks like cancer subtype classification, tissue phenotyping, and survival analysis due to the limited level of detail that a single-resolution image can provide. 
Addressing this, we propose a novel multi-resolution paradigm leveraging Whole Slide Images (WSIs) to extract histology patches at multiple resolutions and generate corresponding textual descriptions through advanced CPath VLM.
We introduce visual-textual alignment at multiple resolutions as well as cross-resolution alignment to establish more effective text-guided visual representations.
Cross-resolution alignment using a multi-modal encoder enhances the model's ability to capture context from multiple resolutions in histology images.
Our model aims to capture a broader range of information, supported by novel loss functions, enriches feature representation, improves discriminative ability, and enhances generalization across different resolutions. 
Pre-trained on a comprehensive TCGA dataset with 34 million image-language pairs at various resolutions, our fine-tuned model outperforms State-Of-The-Art (SOTA) counterparts across multiple datasets and tasks, demonstrating its effectiveness in CPath.
The code is available on GitHub at: \textcolor{magenta}{https://github.com/BasitAlawode/MR-PLIP}.
%
\end{abstract}

\vspace{-8mm}
\section{Introduction}
\label{sec:intro}
\vspace{-1mm}
Cancer diagnosis, outcomes prediction, and the development of treatment strategies often depend on microscopic examination of tissue samples, which is considered the cornerstone of medical practice \cite{abels2019computational, kim2022application, cui2021artificial, verghese2023computational, shastry2022cancer, cifci2023ai}. 
The introduction of digital scanning technology has enabled the creation of high-resolution tissue images, commonly referred to as Whole Slide Images (WSIs) \cite{zhang2019pathologist, cruz2017accurate, cornish2012whole,hanna2020whole, pantanowitz2011review}.
Experienced pathologists analyze these WSIs to identify established cancer subtypes and explore potential new forms of the disease \cite{pantanowitz2011review, hanna2020whole, cornish2012whole}.
This approach has led to the establishment of CPath as a specialized area of study \cite{hosseini2024computational, campanella2019clinical, echle2021deep, berbis2023computational, song2023artificial}.

\begin{figure}[t!]
\centering
\includegraphics[height=2.5in, width=0.90\linewidth]{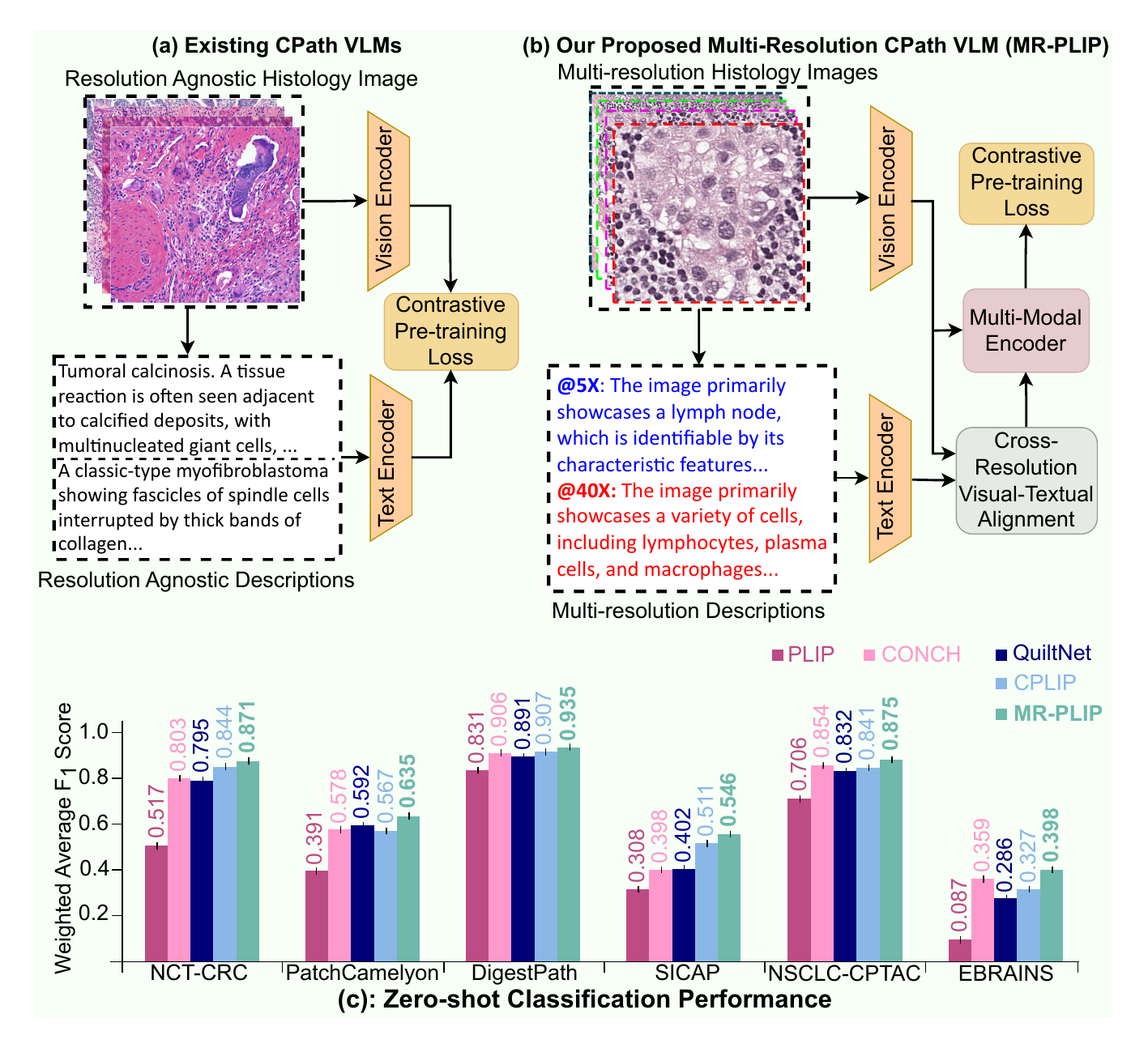}
\vspace{-2mm}
 \caption{\textbf{(a)} Existing CPath VLMs \cite{huang2023visual, lu2023visual}. \textbf{(b)} Our Proposed Multi-Resolution Pathology-Language Pre-training (MR-PLIP) model, and \textbf{(c)} shows the zero-shot classification performance comparison of MR-PLIP and SOTA VLMs in terms of weighted average $F_{1}$ score on publicly available benchmark datasets. 
MR-PLIP outperforms existing SOTA methods.}
\label{fig1_intro}
\vspace{-6mm}
\end{figure}

The advent of Vision-Language (VL) models has further advanced the field of CPath by using the innate zero-shot learning abilities of the foundational models \cite{radford2021learning,seyfioglu2024quilt,javed2024cplip}.
Recent VL models designed for CPath, such as PLIP \cite{huang2023visual}, MI-Zero \cite{lu2023visual}, QuiltNet \cite{ikezogwo2024quilt}, and CPLIP \cite{javed2024cplip}, among others, have been developed to pair images and texts through a contrastive loss mechanism as shown in Fig. \ref{fig1_intro} (a) \cite{radford2021learning, liu2021self}. 
This method facilitates unsupervised learning of features by bringing representations of similar visual-textual content closer together and separating dissimilar ones \cite{radford2021learning}.
These innovations have demonstrated promising results in tasks such as classifying cancer subtypes and segmenting them at the WSI-level based on CPath techniques \cite{lu2023towards, zhang2024biomedclip, acosta2022multimodal}.

\begin{figure}[t!]
\centering
\includegraphics[width=\linewidth]{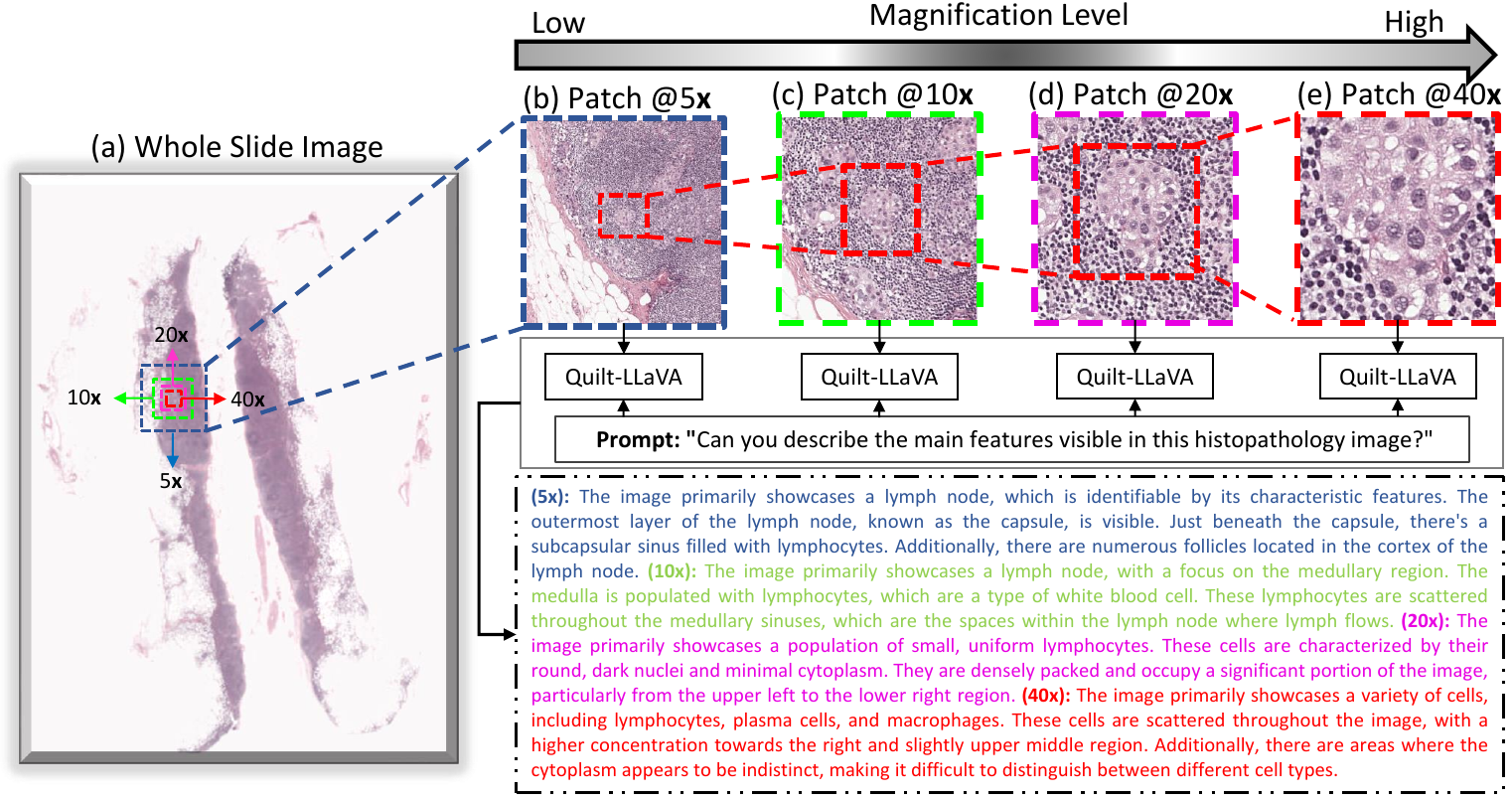}
\vspace{-1.7\baselineskip}
\caption{Analysis of multi-resolution input WSI {\bf(a)} using Quilt-LLaVA \cite{seyfioglu2024quilt}.
Exemplar histology patches {\bf(b)-(e)} are shown at different magnifications, demonstrating how higher magnification (5$\times$ to 40$\times$) shifts focus from contextual to detailed information. 
Textual descriptions generated by QuiltLLaVA vary, reflecting the change in textual details observed at each magnification level.}
\label{fig1}
\vspace{-.7cm}
\end{figure}




In the field of clinical diagnostics, pathologists often engage in the evaluation of WSIs through a process including multiple scales \cite{ding2020multi, bejnordi2015multi, hashimoto2020multi, abels2019computational, zhang2019pathologist, chen2022fast}.
This multi-scale analysis is crucial, as it involves the integration of both overarching (i.e., viewing the WSI at the lowest level of magnification) and detailed (i.e., viewing the WSI at the highest level of magnification) viewpoints \cite{cruz2017accurate,cornish2012whole, hanna2020whole, bolhasani2020histopathological}. 
Such a thorough approach enables pathologists to accurately distinguish between various types of cancer, such as differentiating invasive ductal carcinoma from invasive lobular carcinoma, as well as identifying tumor-infiltrating lymphocytes \cite{aresta2019bach,bolhasani2020histopathological}.
In the context of overarching information, the examination focuses on the overall architectural layout of the tissue samples, assessing the arrangement of each tissue component to pinpoint areas of healthy and potential malignant structures, including ductal carcinoma. 
For the detailed aspects, attention is focused on specific areas within the WSI, inspecting the tissue at a closer, high-magnification level to scrutinize tumor epithelial cells and assess the tumor based on its immediate cellular environment \cite{van2021hooknet}.
Another instance where the blend of overarching and detailed observations is beneficial is in examining the spatial arrangement of immune cells, which can indicate inflammation within or around the tumor, and the formation of tertiary lymphoid structures as a response to cancer.

\begin{figure}[t!]
\centering
\includegraphics[width=0.90\linewidth]{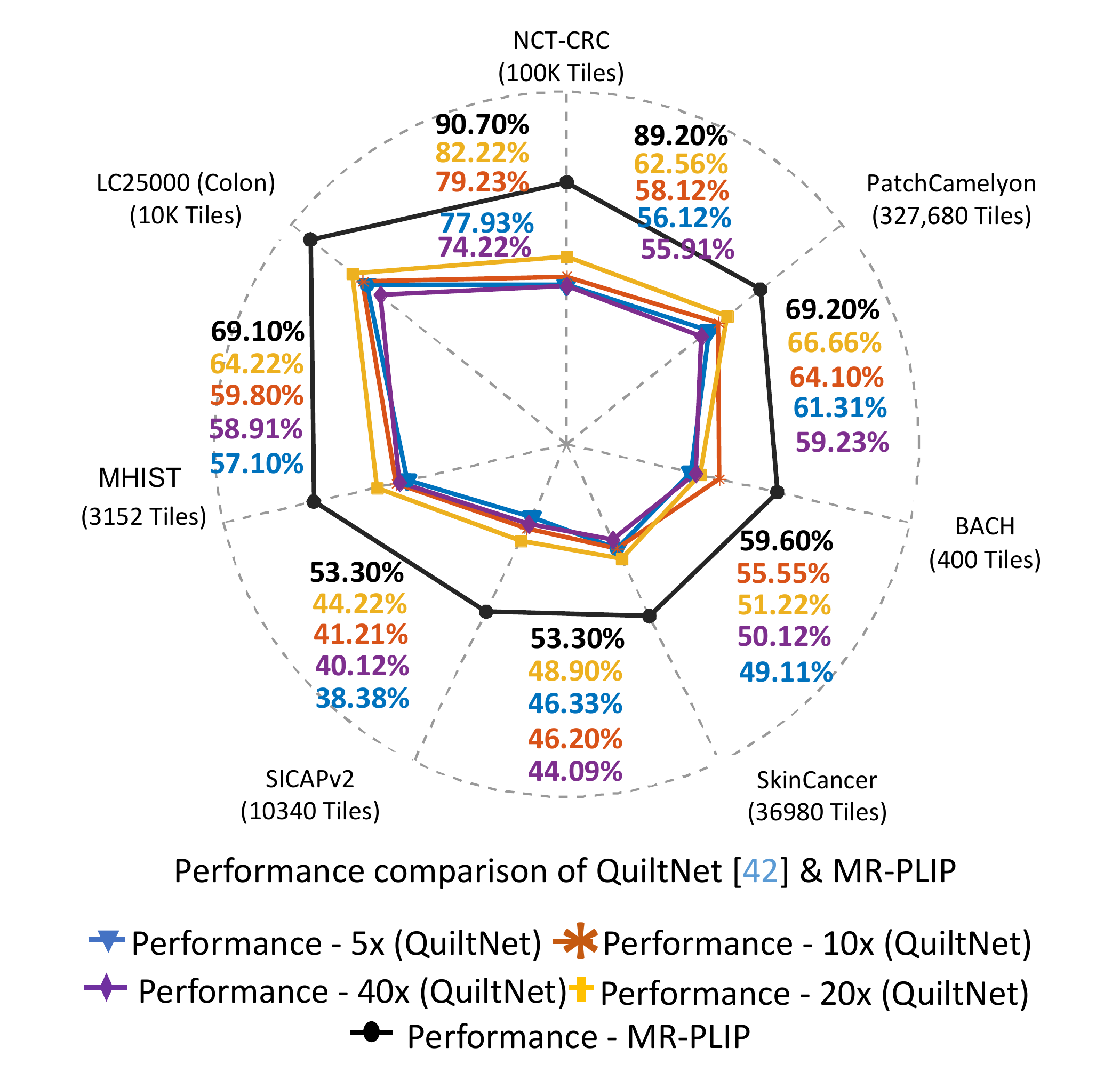}
\vspace{-4mm}
\caption{Comparative analysis of zero-shot tile-based classification balanced accuracy of QuiltNet~\cite{ikezogwo2024quilt} and our proposed MR-PLIP across seven independent datasets.
Both models, pre-trained on the TCGA dataset using 34M image-text pairs, evaluate the impact of fine-tuning language encoders at varying magnification levels$-$5$\times$, 10$\times$, 20$\times$, and 40$\times$.
The variations in performance across these magnifications highlight the necessity for a multi-resolution VLM in CPath to enhance generalization capabilities.}
\label{fig2}
\vspace{-.7cm}
\end{figure}

Most contemporary VLMs in histopathology primarily use histology images extracted from WSI of a single resolution, which might restrict their capability to adequately convey the essential broad and detailed perspectives for optimal analysis \cite{lu2023visual, lu2023towards, huang2023visual}. 
An illustration of this, as shown in Fig. \ref{fig1}, is provided by the SOTA VLM, Quilt-LLaVA \cite{seyfioglu2024quilt}, which demonstrates how, with increasing magnification levels, the amount of textual descriptions derived from the input histology patch varies. 
This decline is attributed to the loss of contextual information at higher magnifications. 
Additionally, pivotal cues, such as those indicating lymph node and lymphocytes, may only be visible at specific magnifications, highlighting the Quilt-LLaVA model's considerable dependency on certain magnification levels for generating accurate textual descriptions, a limitation that might be seen as a drawback.

To explore this issue deeper, we fine-tuned SOTA CPath models, including QuiltNet \cite{ikezogwo2024quilt}, MI-Zero \cite{lu2023visual}, PLIP \cite{huang2023visual}, and BioCLIP \cite{zhang2024biomedclip}, across magnification levels of 5$\times$, 10$\times$, 20$\times$, and 40$\times$, using 20,000 WSIs (comprising 34 million patches) from the TCGA dataset \cite{weinstein2013cancer}.
These models are assessed through zero-shot settings across seven benchmark datasets for tile-based classification, as depicted in Fig. \ref{fig2} and the supplementary material. 
Excluding our top-scoring model MR-PLIP, Fig. \ref{fig2} shows that 20$\times$ is virtually the best, coming first in 13 out of 14 trials. 
The 10$\times$  consistently ranks either first or second in 8 out of 14 trials, while the extreme magnifications of 5$\times$ and 40$\times$ typically land in the last or third position in 10 out of 14 trials. 
These statistics highlight that the magnifications 20$\times$ and 10$\times$, striking a balance between detail and context, achieve optimal performance. 
Our intuition is that integrating the 5$\times$  and 40$\times$ alongside the 20$\times$ and 10$\times$ in VL models will further leverage the complementarity of the four magnification levels.
We suggest, therefore, that synchronizing visual-textual concepts across multiple resolutions enhances their efficacy for diverse CPath applications and their overall generalization (Figs. \ref{fig1_intro} (b)-(c)).

To concretize this advocacy, we present the \textbf{Multi-Resolution Pathology-Language Image Pre-training} (MR-PLIP) model, pre-trained on 34 million histology patches of varied resolutions from the TCGA cohort. 
Our work's contributions are multi-faceted: (\textbf{1}) {\bf We first} noted the performance inconsistency of SOTA VLMs in CPath across different resolutions (see Fig. \ref{fig2}), alongside the variations in content and quantity of textual descriptions across these resolutions (Fig. \ref{fig1}). 
These findings indicate a generalization shortfall in current SOTA VLMs in CPath. 
(\textbf{2}) {\bf Second}, we autonomously generate textual descriptions for histology images at multiple resolutions, which are employed in the training of MR-PLIP (Sec.~\ref{extraction}).
(\textbf{3}) {\bf Third}, we introduce an effective strategy for aligning histology images at various resolutions with their corresponding multi-resolution textual descriptions through the Cross-Resolution Visual-Textual Alignment (CVTA) module, aiming to enhance MR-PLIP's generalization capability across different resolution levels (Sec.~\ref{align}). 
We propose a multi-resolution textual description bag containing all relevant keywords from the textual descriptions. 
In the CVTA module, histology images of different resolutions are matched with this bag using contrastive loss. 
(\textbf{4}) {\bf Lastly}, we merge visual and textual information into discriminative features at various resolutions using a newly developed multi-modal encoder, subsequently refined via a novel Multi-Resolution Text-guided Visual feature representation Alignment (MRTVA) loss (Sec.~\ref{cross-align}). 
Extensive evaluations on 26 publicly available histopathology benchmark datasets including zero-shot, linear probing, and full fine-tuning for a variety of CPath tasks such as tile-level and WSI-level classification and segmentation, and nuclei segmentation reveal significant performance enhancements and generalization across various resolutions, surpassing existing SOTA foundation models in CPath.

The remainder of this paper is organized as follows: Sec. \ref{sec:literature} reviews related work. 
Sec. \ref{sec:method} presents our proposed algorithm. 
Sec. \ref{sec:results} presents the experiments results, and Sec. \ref{sec:conclusion} concludes our work.

\vspace{-3mm}
\section{Literature Review}
\label{sec:literature}
\vspace{-.1cm}
In recent years, foundation models and Multiple Instance Learning (MIL)-based methods have become the mainstream in CPath \cite{srinidhi2021deep, hosseini2024computational, kapse2024si, li2024dynamic, tang2024feature, ilse2018attention, chen2024towards}.
These paradigms have advanced the performance of several CPath tasks including diagnosis, subtyping, prognosis, and more \cite{tang2024feature, kapse2024si, tokunaga2019adaptive, ding2020multi, bejnordi2015multi,hashimoto2020multi, abels2019computational}.
Below, we discuss these paradigms in detail.

\begin{figure*}[t!]
\centering
\includegraphics[width=0.90\linewidth]{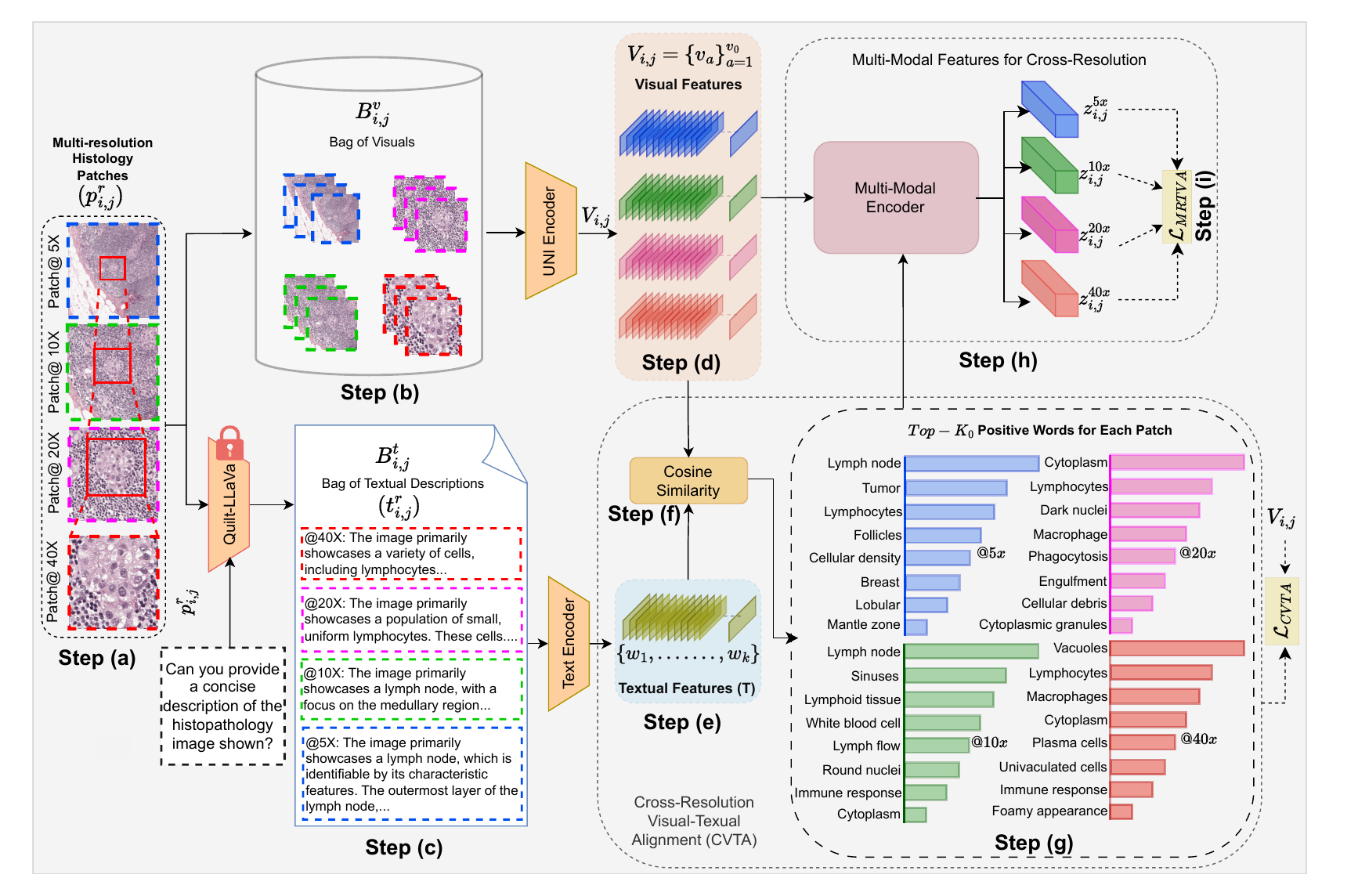}
\vspace{-0.80\baselineskip}
\caption{Outline of the proposed MR-PLIP algorithm workflow. 
Starting with step {\bf (a)}, where multi-resolution histology patches are extracted.
In step {\bf (b)}, these patches are compiled for each resolution into a visual bag. 
Step {\bf (c)} involves the extraction of textual descriptions for the patches using Quilt-LLaVA \cite{seyfioglu2024quilt}. 
Steps {\bf (d)-(f)} are focused on generating matching embeddings for the visual and textual data via visual encoder (UNI \cite{chen2024towards}) and text encoder (QuiltNet \cite{ikezogwo2024quilt}), respectively. 
The high similarity between the visual and textual embeddings allows for the identification of the top-$k_{o}$ positive words for each patch in step {\bf (g)}. 
Lastly, steps {\bf (h)-(i)} process these top-$k_{o}$ words and visual features through a multi-modal encoder, designed to uncover spatial interrelations in the multi-modal context.}
\label{fig3}
\vspace{-0.6cm}
\end{figure*}

\noindent \textbf{Foundation Models:} The recent literature in CPath primarily centers around the development of foundation models \cite{chanda2024new,song2023artificial}.
In CPath, two main types of foundation models have emerged: VLMs and vision-only models.

The \textbf{VLMs} in CPath integrate textual descriptions with visual data to guide the training of vision-language encoders using a contrastive learning \cite{ikezogwo2024quilt, lu2023visual, huang2023visual, zhang2024biomedclip}. 
This approach aligns visually and textually similar concepts while distinguishing dissimilar ones \cite{radford2021learning}.
These foundational models, which support multi-modal learning of vision and text encoders, have been highly effective for a range of CPath applications. 
BioCLIP was trained on 15 million image-text pairs, spanning a variety of medical imaging tasks \cite{zhang2024biomedclip}.
PLIP model, pre-trained on 208K pathology vision-language pairs sourced from Medical Twitter, has exhibited improved zero-shot learning capabilities for numerous CPath tasks \cite{huang2023visual}.
Other VLMs introduced in CPath include MI-Zero \cite{lu2023visual}, CONCH \cite{lu2023visual}, QuiltNet \cite{ikezogwo2024quilt}, and CPLIP \cite{javed2024cplip}.
Li \textit{et al.} developed FiVE, integrating fine-grained visual-semantic details \cite{li2024generalizable}, and Yin \textit{et al.} proposed a prompting VLM \cite{yin2024prompting}.
Shi \textit{et al.} recently developed a dual-scale VLM for WSI classification, focusing on 5$\times$ and 10$\times$ magnifications \cite{shi2024vila}.
The latest VLMs in CPath such as PathChat \cite{lu2024multimodal} Quilt-LLAVA \cite{seyfioglu2024quilt}, specifically designed for visual question-answering tasks.
\textit{Despite their notable performance, multi-resolution pre-training has yet to be comprehensively explored in these models.}

Unlike image-text contrastive pre-training, \textbf{vision-only foundation models} employ Self-Supervised Learning (SSL) paradigm using contrastive loss solely to histopathology images, leveraging large-scale datasets for pre-training. 
Recently, several large-scale vision-only foundation models are introduced in the literature, such as CTransPath \cite{wang2022transformer}, UNI \cite{chen2024towards}, and GigaPath \cite{xu2024whole} etc.
These models are pre-trained using ViT-based architectures and the DINOV2 SSL strategy \cite{dosovitskiy2020image, kang2023benchmarking, caron2021emerging}. 
All of these models have demonstrated exceptional performance in various downstream CPath tasks. 
These models have shown exceptional performance across a variety of downstream CPath tasks.
\textit{In contrast, our proposed MR-PLIP model, pre-trained on 20K WSIs (34 million histology patches), stands out by incorporating multi-resolution histology images along with visual-textual feature interactions, a key element that is absent in the aforementioned foundation models.}

\noindent \textbf{MIL-based Methods:} These methods utilize patches as instances extracted from WSI to estimate the slide-level representations utilized for many CPath tasks \cite{srinidhi2021deep, hosseini2024computational}.
These methods, however, depend significantly on a large collection of WSIs that are labeled only at the slide-level, which is often difficult to obtain for rare diseases. Additionally, because these models only learn from the original slide, they are susceptible to variations in data distribution, resulting in less effective generalization performance \cite{li2021multi, kapse2024si, srinidhi2021deep, ilse2018attention}.

Recently, several successful MIL-based methods are proposed for histology image analysis \cite{hashimoto2020multi, li2024dynamic, tokunaga2019adaptive}.
For instance, Hashimoto \textit{et al.} proposed multi-resolution MIL \cite{hashimoto2020multi}.  
Li \textit{et al.} proposed DSMIL that combines information from two distinct magnification levels for WSI classification \cite{li2021dual}.
Tokunaga \textit{et al.} introduced an adaptive weighting multi-resolution framework for semantic segmentation \cite{tokunaga2019adaptive}.
Rijthoven \textit{et al.} presented HookNet, that processes images at two different resolutions to capture both contextual and detailed information for WSI segmentation \cite{van2021hooknet}. 
Ilse \textit{et al.} proposed ABMIL method, which has extensively been employed in CPath for WSI classification \cite{ilse2018attention}. 
Other successful MIL-based methods include HIPT \cite{chen2022scaling}, H2T \cite{vu2023handcrafted}, WiKG \cite{li2024dynamic}, and SI-MIL \cite{kapse2024si}.
\textit{Importantly, none of these models utilize language supervision, a feature central to the approach outlined in the current study.}
\vspace{-.1cm}

\section{Proposed Multi-Resolution PLIP Model}
\label{sec:method}
\vspace{-2mm}
The schematic illustration of our proposed MR-PLIP algorithm is shown in Fig. \ref{fig3}.
MR-PLIP algorithm consists of four main steps: the extraction of patch-text pairs at multiple resolutions (Sec.\ref{extraction} \& Figs.\ref{fig3} (a)-(e)), a module for aligning these pairs across different visual-textual resolutions (Sec.\ref{align} \& Figs. \ref{fig3} (f)-(g)), a multi-modal encoder designed for the fusion of visual-textual features (Sec.\ref{cross-align}, Fig. \ref{fig3} (h)), and the alignment of these features across various resolutions (Sec. \ref{cross-align} \& Fig. \ref{fig3} (i))).
\vspace{-.2cm}

\subsection{Generating Multi-Resolution Histology Bags: Image Patches and Textual Descriptions}
\label{extraction} 
\vspace{-.2cm}
We begin by selecting $N=20,000$ WSIs $\{W_{i}\}_{i=1}^N$, from the TCGA dataset~\cite{weinstein2013cancer}, a key repository in histopathology. 
We extract $n$ patches $\{p_{i,j}^{r}\}_{j=1}^n$ at varying resolution levels $r\in \{5\times, 10\times, 20\times, 40\times\}$.
Our first step involved extracting 20 random patches from each WSI at a 5$\times$ magnification level (2 $\mu m$/pixel), with each patch sized at $512 \times 512$ pixels for uniformity. 
To further explore these images, we processed the initial patches to higher magnifications of 10$\times$, 20$\times$, and 40$\times$, corresponding to 1 $\mu m$/pixel, 0.5 $\mu m$/pixel, and 0.25 $\mu m$/pixel, respectively.
This resulted in four 10$\times$ patches, 16 20$\times$ patches, and 64 40$\times$ patches from each original 5$\times$ patch.
This approach allows us to examine the cellular structures in greater detail across different scales.

To organize this extensive dataset, we established a parent-child relationship among the patches, where each lower resolution patch was linked as a `parent' to four `child' patches at the next higher resolution.
This strategy resulted in a cumulative total of 34 million patches across all levels of magnification, significantly enforcing the \textcolor{black}{multi-resolution capabilities} of SOTA VLMs in CPath.
Furthermore, to integrate textual information with our image data, we used the \textcolor{black}{Quilt-LLaVA \cite{seyfioglu2024quilt} and QuiltNet \cite{ikezogwo2024quilt} models, pre-trained on large-scale histopathology image-text pairs} from various sources, including YouTube videos.
This step allowed us to extract textual descriptions $t_{i,j}^{r}$ for each patch $p_{i,j}^{r}$, enriching our dataset with information that could lead to more detailed interpretations of the histopathological images. 
Through this careful and organized process, we aim to enhance our understanding of histopathological imagery by using both visual and textual data at different resolutions to provide significant insights in CPath.

The set of all patches ($10\times$, $20\times$, and $40\times$) corresponding to the same parent $5\times$ patch $p_{i,j}^{5\times}$ is considered to represent a single visual bag $B^{v}_{i,j}$ where $|B^{v}_{i,j}|=v_{o}$ and $v_{o}$=85 in our case.
For each patch $p_{i,j}^{r} \in B^{v}_{i,j}$, we extract a visual feature vector $v_{i,j}^{r} \in \mathcal{R}^{d}$ using a \textcolor{black}{UNI vision encoder (ViT-L/16) \cite{chen2024towards}} and generate $V_{i,j}=\{v_a \}_{a=1}^{v_o}$.
\textcolor{black}{We also generate a textual description of each patch $p_{i,j}^{r} \in B^{v}_{i,j}$, by harnessing the existing CPath VL model Quilt-LLaVA \cite{seyfioglu2024quilt} using the input prompt: ``\textit{Can you describe the main features visible in this histopathology image?} ''.}
For each visual bag, we generate a bag of textual descriptions $B^{t}_{i,j}$ by concatenating all the textual descriptions corresponding to the patches in $B^{v}_{i,j}$.
We use a text encoder of the QuiltNet \cite{ikezogwo2024quilt} to obtain textual representations from $B^{t}_{i,j}$ corresponding to different keywords $T=\{w_b\}_{b=1}^{k}$, where $k$ are the words in the textual bag and $w_{b}$ is the feature vector of $b$-th word.
\vspace{-.2cm}
\subsection{Visual-Textual Feature Alignment Across Resolutions for Top $k_0$ Words Extraction}
\label{align}
\vspace{-.2cm}
Within the textual bag $B^{t}_{i,j}$, only a subset of the textual descriptions will directly relate to a specific patch when viewed at a higher resolution. Additionally, these textual descriptions are produced via unsupervised methods and might include irrelevant keywords that do not accurately relate to a given patch. 
Consequently, it becomes necessary to adapt the visual and text encoders to function effectively across varying resolutions. 

\noindent To address this, we introduce the Cross-Resolution Visual-Textual Alignment (CVTA) module, which is designed to calibrate both encoders to work in harmony across different levels of magnification.
To achieve this, for each visual feature represented by $v_a$, we identify $k_{o}< k$ positive keywords $w_{b}^{+}$ from the corresponding $T_{i,j}$. This identification is based on maximizing the cosine similarity between the word feature $w_b \in T_{i,j}$ and the visual feature $v_a \in V_{i,j}$, calculated as the dot product $w_{b}^{\top}v_{a}/||w_b||_2||v_a||_2$.
The rationale behind this selection process is that not all textual features in $B^{t}_{i,j}$ are relevant to each visual feature $v_a$; those that are not relevant are treated as negative keywords. We then apply a contrastive loss function that is minimized by training with these identified positive and negative keyword pairs as shown:
\vspace{-.4cm}
\begin{equation}
\mathcal{L}_{\textrm{CVTA}}=-\frac{1}{v_{o}}\sum_{a=1}^{v_{0}}\Bigg(\frac{1}{k_0}\sum_{k_o}  \log \frac{\exp(v_{a}^{\top}w_{b}^{+}/\tau)}{\sum_{b=1}^{k}(\exp(v_{a}^{\top}w_{b}/\tau))} \Bigg),
\label{eqn1}
\end{equation}
\vspace{-.3cm}

\noindent where $\tau$ represents a learnable temperature hyper-parameter that modulates the concentration of the output distribution, initially set to 0.07.
\vspace{-.2cm}
\subsection{Integrating Multi-Modal Features for Cross-Resolution Alignment}
\label{cross-align}
\vspace{-.2cm}
Although the vision encoder has demonstrated its effectiveness at implicitly learning visual representations as shown in previous studies~\cite{kang2023benchmarking,lu2023towards, ikezogwo2024quilt}, it struggles with identifying the intrinsic contextual relationships across multiple resolutions. This limitation impedes the performance of \textcolor{black}{VLMs}, particularly affecting downstream tasks in CPath. 
Specifically, this gap is evident in current CPath \textcolor{black}{VL} pre-training paradigms~\cite{lu2023towards, ikezogwo2024quilt, huang2023visual, lu2023visual}, which primarily concentrate on aligning visual and textual data. These models often overlook the role of text in guiding the learning of contextual representations at different resolutions, thereby missing out on crucial contextual details.

To address these challenges, we propose the multi-modal cross-resolution features alignment approach, a novel contextual information-aware pre-training task that enhances the ability of models to capture context from multiple resolutions in histopathology patches.
This is achieved through precise text-driven guidance, which facilitates the alignment of multi-modal data pairs. 
Specifically, our model enables the visual features obtained from lower magnification levels, like $v^{5\times}_{i,j}$, to inform and contextualize the features from patches at higher magnification levels, such as  $v^{10\times}_{i,j}$.
Consequently, text features are extracted and interact with the visual features through the multi-modal encoder, resulting in a text-guided visual representation, $z^{r}_{i,j}$.

\vspace{-.2cm}
\subsubsection{\bf Text-guided Visual Representation}
\vspace{-.1cm}
For every patch $p^{r}_{i,j}$, we extract a visual feature $v^{r}_{i,j}$ and $k_{o}$ textual features $w^{+}_{b} \in T_{i,j}$, representing all the positive keywords, using visual and text encoders trained through the CVTA module.
These visual and textual features are then concatenated and fed into a multi-modal encoder, specifically trained to achieve cross-resolution feature alignment, as in~\cite{li2022mplug, ALBEF}:
\vspace{-.4cm}
\begin{equation}
~~~~~z^{r}_{i,j}=E_{mm}(v^{r}_{i,j},w^{+}_{1},w^{+}_{2},\cdot \cdot w^{+}_{k_{o}}),
\label{eqn2}
\end{equation}
\vspace{-.6cm}

\noindent where $E_{mm}(\cdot,\cdot)$ denotes the multi-modal encoder that produces the output $z^{r}_{i,j}$, with $r$ indicating the resolution of the input patch and $r \in R$ representing the complete set of possible resolutions.
The text-guided representation $z^{r}_{i,j}$ facilitates the alignment of visual features with their corresponding optimal textual representations. However, it doesn't guarantee consistency across different resolutions, such as between parent and child resolution levels. 
Achieving this consistency is essential for maintaining coherence between the overall structure and the more detailed aspects within histology patches.
To enforce this consistency, we apply a Multi-Resolution Text-guided Visual feature representations Alignment (MRTVA) loss function derived from the SimSiam framework~\cite{chen2021exploring}:
\vspace{-8mm}

\begin{equation}
\mathcal{L}_{MRTVA} (h^{p}_{i,j}, g^{c}_{i,j})= -\sum_{p,c \in R, p \ne c}\Big(\frac{h^{p}_{i,j}}{||h^{p}_{i,j}||_{2}}\cdot\frac{g^{c}_{i,j}}{||g^{c}_{i,j}||_{2}} \Big),
\label{eqn3}
\end{equation}

\noindent where $p$ represents the parent patch at a lower magnification level, and $c$ is the child patch at a higher magnification level, which is key for achieving cross-resolution alignment. This alignment between the text-guided visual features of parent and child patches, as illustrated in Fig. \ref{fig3} (h) as well as in the supplementary material, is critical because it preserves contextual information across the parent-child hierarchy. 
Moreover, $h_{i,j}=p_{d}(p_{j}(z_{i,j}))$ and $g_{i,j}= p_{j}(z_{i,j})$, where $p_{d}$ and $p_{j}$ are the projection and prediction heads, respectively, as used in~\cite{chenempirical,grill2020bootstrap,chen2020simple}. 
Minimizing the $\mathcal{L}_{MRTVA}$ loss across multi-resolution text-guided visual features essentially reduces to minimizing the mean square error, adjusted by a scale factor, thus ensuring the alignment of multi-modal features across various resolutions. 
Following \cite{chen2021exploring,grill2020bootstrap}, we introduce a symmetric loss defined as:
\vspace{-6mm}

\begin{equation}
\begin{split}
\mathcal{L}_{MRTVA}= \frac{1}{2} [\mathcal{L}_{MRTVA} (h^{p}_{i,j}, S_{g}(g^{c}_{i,j}))+ \\
\mathcal{L}_{MRTVA} (g^{p}_{i,j}, S_{g}(h^{c}_{i,j}))],
\end{split}
\label{eqn4}
\end{equation}

\noindent Here, $S_{g}(\cdot)$ denotes the stop-gradient operation, a technique employed to prevent model collapse during training \cite{chen2021exploring}.
\vspace{-.2cm}
\subsection{Pre-Training Tasks}
\vspace{-.2cm}
Building on the approaches outlined in~\cite{li2022mplug, ALBEF, Ye_2023_ICCV}, we integrate four pre-training tasks: Image-Text Contrastive (ITC) learning, Image-Text Matching (ITM), Masked Language Modeling (MLM), and Prefix Language Modeling (PLM). 

In the ITC task, we align the image features with textual features extracted by unimodal encoders, following the methodologies described in~\cite{li2022mplug, ALBEF}. 
During the ITM task, the goal is to ascertain whether there is a match between an image and text using the derived cross-modal representations~\cite{li2022mplug, ALBEF}. 
For MLM, we mask 15\% of the text tokens at random, and the model must predict the concealed words relying on the cross-modal representations~\cite{li2022mplug, ALBEF}. 
In PLM, based on a given image, we generate captions and predict the ensuing text segment in alignment with the cross-modal context~\cite{li2022mplug, ALBEF}. 
For the sake of simplicity, we combine these loss functions into a single baseline loss function $\mathcal{L}_{bl}=ITC+ITM+MLM+PLM$. 
Consequently, the full pre-training objective for our MR-PLIP model is formulated as follows: 
\vspace{-2mm}

\vspace{-2mm}
\begin{equation}
\mathcal{L}_t  =  \mathcal{L}_{bl}+\mathcal{L}_{CVTA}+\mathcal{L}_{MRTVA}.
\label{eqn5}
\vspace{-2mm}
\end{equation}

\begin{table*}[t!]
\centering
\caption{Zero-shot tile and WSI levels classification performance comparison of MR-PLIP and SOTA VLMs in terms of weighted average $F_{1}$ score with Prompts Ensembling (PE) and no PE (nPE) \cite{lu2023towards}. 
MR-PLIP outperforms existing models on all datasets.}
\vspace{-0.5\baselineskip}
\scalebox{0.62}{
    \begin{tabular}{|c|c|c|c|c|c|c|c|c|c|c|c|c|c|c|c|c|}
   \hline
   \textbf{Tile Datasets} &\multicolumn{2} {c|} {CLIP}&\multicolumn{2} {c|} {BioCLIP}&\multicolumn{2} {c|} {PLIP}&\multicolumn{2} {c|} {MI-Zero}&\multicolumn{2} {c|} {CONCH}&\multicolumn{2} {c|} {QuiltNet}&\multicolumn{2} {c|} {CPLIP}&\multicolumn{2} {c|} {MR-PLIP} \\
    \hline
    &PE&nPE&PE&nPE&PE&nPE&PE&nPE&PE&nPE&PE&nPE&PE&nPE&PE&nPE\\
    Databiox \cite{bolhasani2020histopathological}&0.162&0.121&0.286&0.231&0.384&0.331&0.458&0.401&0.473&0.416&0.433&0.397&\underline{0.487}&\underline{0.462}&\textbf{0.532}&\textbf{0.498}\\
    BACH \cite{iciar2018grand}&0.230&0.201&0.202&0.176&0.381&0.336&0.441&0.4031&0.522&0.483&0.541&0.485&\underline{0.563}&\underline{0.528}&\textbf{0.605}&\textbf{0.587}\\
    PatchCamelyon \cite{veeling2018rotation} &0.255&0.223&0.302&0.266&0.391&0.362&0.542&0.510&0.578&\underline{0.543}&\underline{0.592}&0.551&0.567&0.531&\textbf{0.635}&\textbf{0.604}\\
    Osteo \cite{arunachalam2019viable}&0.302&0.244&0.371&0.326&0.452&0.409&0.531&0.498&0.577&0.535&\underline{0.581}&0.521&0.580&\underline{0.544}&\textbf{0.656}&\textbf{0.606}\\
    SkinCancer \cite{kriegsmann2022deep}&0.256&0.231&0.322&0.271&0.393&0.355&0.436&0.392&\underline{0.462}&0.433&0.402&0.367&0.451&\underline{0.476}&\textbf{0.513}&\textbf{0.479}\\
    MHIST \cite{wei2021petri}&0.333&0.2887&0.388&0.331&0.451&0.431&0.523&0.497&0.546&0.502&\underline{0.572}&0.521&0.571&\underline{0.541}&\textbf{0.643}&\textbf{0.596}\\
    RenalCell \cite{brummer2022integrative}&0.200&0.176&0.244&0.219&0.353&0.314&0.503&0.455&0.512&0.467&0.503&0.487&\underline{0.522}&\underline{0.491}&\textbf{0.568}&\textbf{0.531}\\
    NCT-CRC \cite{kather2018100}&0.247&0.185&0.533&0.372&0.517&0.687&0.786&0.536&0.803&0.542&0.795&0.553&\underline{0.844}&\underline{0.681}&\textbf{0.871}&\textbf{0.832}\\
    LC25000Lung \cite{borkowski2019lung}&0.361&0.321&0.431&0.402&0.558&0.502&0.802&0.716&\underline{0.805}&0.735&0.781&0.730&0.800&\underline{0.752}&\textbf{0.853}&\textbf{0.824}\\
    LC25000Colon \cite{borkowski2019lung}&0.402&0.351&0.453&0.403&0.578&0.524&0.868&0.802&\underline{0.871}&0.822&0.866&0.806&0.842&\underline{0.791}&\textbf{0.882}&\textbf{0.851}\\
    DigestPath \cite{da2022digestpath}&0.151&0.123&0.501&0.622&0.831&0.782&0.902&0.861&0.906&\underline{0.866}&0.891&0.803&\underline{0.907}&0.861&\textbf{0.935}&\textbf{0.902}\\
    SICAP \cite{silva2021self}&0.183&0.140&0.438&0.352&0.308&0.226&0.387&0.235&0.398&0.241&0.402&0.363&\underline{0.511}&\underline{0.388}&\textbf{0.546}&\textbf{0.487}\\
    WSSS4LUAD \cite{han2022wsss4luad} &0.196&0.195&0.575&0.452&0.636&0.408&0.697&0.582&0.705&0.590&0.701&0.673&\textbf{0.882}&\underline{0.791}&\underline{0.864}&\textbf{0.831}\\
    \hline
    \textbf{WSI Datasets}&\multicolumn{2} {c|} {CLIP}&\multicolumn{2} {c|} {BioCLIP}&\multicolumn{2} {c|} {PLIP}&\multicolumn{2} {c|} {MI-Zero}&\multicolumn{2} {c|} {CONCH}&\multicolumn{2} {c|} {QuiltNet}&\multicolumn{2} {c|} {CPLIP}&\multicolumn{2} {c|} {MR-PLIP} \\
    \hline
    &PE&nPE&PE&nPE&PE&nPE&PE&nPE&PE&nPE&PE&nPE&PE&nPE&PE&nPE\\
    CAMELYON16 \cite{bejnordi2017diagnostic}&0.175&0.198&0.402&0.377&0.416&0.442&0.522&0.461&0.538&0.476&0.562&0.504&\underline{0.632}&\underline{0.587}&\textbf{0.664}&\textbf{0.632}\\
    CAMELYON17 \cite{bandi2018detection}&0.171&0.143&0.287&0.322&0.401&0.386&0.455&0.421&0.501&0.441&0.489&0.446&\underline{0.518}&\underline{0.473}&\textbf{0.581}&\textbf{0.548}\\
    NSCLC-CPTAC \cite{edwards2015cptac} &0.202&0.146&0.275&0.311&0.706&0.661&0.804&0.753&0.854&0.806&0.832&0.773&0.841&\underline{0.809}&\textbf{0.875}&\textbf{0.842}\\
    EBRAINS \cite{thomas2022digital} &0.029&0.020&0.124&0.093&0.087&0.013&0.331&0.253&0.359&\underline{0.304}&0.286&0.229&\underline{0.327}&0.256&\textbf{0.398}&\textbf{0.332}\\
\hline
      \end{tabular}}
    \label{table1}
\vspace{-.3cm}
\end{table*}

\begin{table*}[t!]
\centering
\caption{Performance comparison of proposed MR-PLIP with existing SOTA foundation models. 
Tile-level classification performance is reported using linear probe evaluations, while WSI-level classification results are based on weakly-supervised learning. 
BA represents balanced accuracy and $F_{1}$ is the weighted $F_{1}$ score. 
MR-PLIP achieves top performance across most datasets.}
\vspace{-0.6\baselineskip}
\makebox[\linewidth]{
\scalebox{0.62}{
\begin{tabular}{ c| c c| c c| c c| c c |c c |c c|cc|cc|cc|}
\hline
\textbf{Datasets} & \multicolumn{2}{c|}{CONCH} & \multicolumn{2}{c|}{QuiltNet} & \multicolumn{2}{c|}{UNI}& \multicolumn{2}{c|}{REMEDIS}&\multicolumn{2}{c|}{Virchow} &\multicolumn{2}{c|}{CHIEF}&\multicolumn{2}{c|}{CTransPath}&\multicolumn{2}{c|}{GigaPath}&\multicolumn{2}{c|}{MR-PLIP}\\
(Tile-level)& BA & $F_{1}$ & BA & $F_{1}$ & BA & $F_{1}$ & BA & $F_{1}$ & BA & $F_{1}$& BA & $F_{1}$&BA&$F_{1}$&BA&$F_{1}$&BA&$F_{1}$\\
\hline
NCT-CRC&0.938&0.955&0.922&0.947&0.874&0.875&0.787&0.802&\underline{0.960}&\underline{0.968}&0.844&0.856&0.845&0.867&-&-&\textbf{0.965}&\textbf{0.976}\\
PatchCamelyon&0.866&0.869&0.822&0.831&0.901&0.930&0.805&0.822&\underline{0.933}&\underline{0.933}&0.833&0.851&0.911&0.935&-&-&\textbf{0.955}&\textbf{0.961}\\
WILDS-CAM17&0.911&0.925&0.861&0.877&\textbf{0.983}&\textbf{0.983}&0.926&0.926&0.971&0.971&0.901&0.922&0.960&0.960&-&-&\underline{0.975}&\underline{0.980}\\
MHIST&0.791&0.807&0.802&0.823&\underline{0.856}&\underline{0.881}&0.781&0.807&0.831&0.836&0.791&0.813&0.811&0.826&-&-&\textbf{0.876}&\textbf{0.915}\\
SICAP&0.711&0.745&0.722&0.767&0.826&0.841&0.806&0.811&\underline{0.855}&\underline{0.873}&0.771&0.783&0.678&0.747&-&-&\textbf{0.886}&\textbf{0.905}\\
WSSS4LUAD&0.811&0.825&0.805&0.812&0.831&0.835&0.769&0.782&\underline{0.866}&\underline{0.873}&0.812&0.828&0.844&0.857&-&-&\textbf{0.887}&\textbf{0.896}\\
BACH&0.856&0.871&0.833&0.861&\underline{0.925}&\underline{0.926}&0.863&0.864&0.915&0.920&0.847&0.863&0.875&0.872&-&-&\textbf{0.945}&\textbf{0.966}\\
UniToPatho&0.451&0.467&0.446&0.457&0.504&0.533&0.446&0.473&\underline{0.557}&\underline{0.574}&0.405&0.416&0.432&0.481&-&-&\textbf{0.605}&\textbf{0.622}\\
\hline
\textbf{Datasets} & \multicolumn{2}{c|}{CONCH} & \multicolumn{2}{c|}{QuiltNet} & \multicolumn{2}{c|}{UNI}& \multicolumn{2}{c|}{REMEDIS}&\multicolumn{2}{c|}{Virchow} &\multicolumn{2}{c|}{CHIEF}&\multicolumn{2}{c|}{CTransPath}&\multicolumn{2}{c|}{GigaPath}&\multicolumn{2}{c|}{MR-PLIP}\\    
(WSI-level)& BA & $F_{1}$ & BA & $F_{1}$ & BA & $F_{1}$ & BA & $F_{1}$ & BA & $F_{1}$& BA & $F_{1}$&BA&$F_{1}$&BA&$F_{1}$&BA&$F_{1}$\\
\hline
CAM16&0.881&0.902&0.902&0.922&\underline{0.957}&0.961&0.930&0.923&0.951&0.913&0.944&0.952&0.897&0.907&\textbf{0.967}&\underline{0.960}&0.950&\textbf{0.966}\\
RCC-DHMC&0.856&0.866&0.851&0.864&0.919&0.926&0.865&0.877&\underline{0.922}&0.931&0.897&0.901&0.804&0.883&0.921&\underline{0.936}&\textbf{0.941}&\textbf{0.952}\\
HunCRC&\underline{0.681}&0.721&0.702&0.722&0.643&\textbf{0.824}&0.604&\underline{0.787}&0.621&0.667&0.651&0.643&0.556&0.728&0.641&0.667&\textbf{0.688}&0.701\\
BRCA-BRACS&\underline{0.723}&\underline{0.748}&0.718&0.725&0.687&0.691&0.676&0.696&0.708&0.722&0.656&0.666&0.639&0.648&0.704&0.715&\textbf{0.741}&\textbf{0.767}\\
PANDA&0.702&0.733&0.722&0.744&\underline{0.757}&\underline{0.809}&0.711&0.766&0.728&0.741&0.724&0.745&0.691&0.752&0.744&0.789&\textbf{0.786}&\textbf{0.816}\\
EBRAINS&0.687&0.717&0.655&0.666&0.675&\underline{0.746}&0.382&0.471&\underline{0.701}&0.723&0.688&0.706&0.514&0.597&0.687&0.704&\textbf{0.745}&\textbf{0.763}\\
NSCLC-CPTAC&0.881&0.902&0.877&0.900&0.904&0.935&0.841&0.866&\underline{0.923}&\underline{0.936}&0.922&0.934&0.877&0.895&0.900&0.915&\textbf{0.930}&\textbf{0.955}\\
\hline     
\end{tabular}
}}
\label{table6_new}
\vspace{-5mm}
\end{table*}

\section{Experimental Evaluations}
\label{sec:results}
\vspace{-.2cm}
We thoroughly evaluated the proposed MR-PLIP \textcolor{black}{model} through rigorous experimental testing on a diverse range of tasks.
These comprehensive assessments spanned multiple modalities and levels of analysis, specifically: tile-based classification, WSI-level classification, WSI-level segmentation, nuclei segmentation, and cross-model retrieval. 
Across all experiments, we used zero-shot, linear \textcolor{black}{probe}, and fine-tuned evaluative protocols to quantify model performance. 
The extensive evaluations validate MR-PLIP's effectiveness and versatility for various precision medicine applications.
\vspace{-.3cm}

\subsection{Training and Implementation Details}
\vspace{-.1cm}
We pre-trained MR-PLIP model on TCGA \cite{weinstein2013cancer} data by extracting 34 million multi-resolution histology patches from 20K WSIs, each of dimension $512 \times 512$ pixels. 
The architecture of MR-PLIP model consists of a frozen Quilt-LLaVA \cite{seyfioglu2024quilt}, a UNI vision encoder (ViT-L/16-224) \cite{chen2024towards}, a QuiltNet text encoder (GPT-2/77) \cite{ikezogwo2024quilt}, and a multi-modal encoder \cite{ALBEF, li2022mplug}.
Our model is pre-trained using a batch size of 64 and 50 epochs with AdamW optimizer~\cite{adam2014method}.
We used the optimal number of positive keywords ($k_{o}$) at 9. 
For a fair comparison, we reported balanced accuracy and weighted average $F_{1}$ scores for classification tasks similar to existing CPath methods \cite{chen2024towards, lu2023visual}. 
\vspace{-.3cm}
\subsection{Downstream Histopathology Datasets}
\vspace{-.2cm}
We performed our experiments on five different CPath tasks across \textcolor{black}{26} independent datasets.
We performed \textbf{\textit{tile-level}} classification tasks on 15 distinct datasets including  Databiox~\cite{bolhasani2020histopathological}, \textcolor{black}{WILDS-CAM17 \cite{bandi2018detection,koh2021wilds}}, \textcolor{black}{UniToPatho \cite{barbano2021unitopatho}}, BACH~\cite{iciar2018grand}, PatchCamelyon~\cite{veeling2018rotation}, Osteo~\cite{arunachalam2019viable}, SkinCancer~\cite{kriegsmann2022deep}, MHIST~\cite{wei2021petri}, RenalCell~\cite{brummer2022integrative}, WSSS4LUAD~\cite{han2022wsss4luad}, DigestPath~\cite{da2022digestpath}, SICAP~\cite{silva2021self},  LC25000Lung~\cite{borkowski2019lung}, NCT-CRC~\cite{kather2018100}, and LC25000Colon~\cite{borkowski2019lung}.
We performed \textbf{\textit{WSI-level}} classification tasks on eight independent datasets: CAMELYON16 (CAM16)~\cite{bejnordi2017diagnostic}, CAMELYON-17 (CAM17)~\cite{bandi2018detection}, \textcolor{black}{NSCLC-CPTAC \cite{edwards2015cptac}, \textcolor{black}{RCC-DHMC \cite{zhu2021development}},  \textcolor{black}{BRCA-BRACS \cite{brancati2022bracs}},  \textcolor{black}{HunCRC \cite{pataki2022huncrc}},  \textcolor{black}{PANDA \cite{bulten2022artificial}}, 
and EBRAINS \cite{roetzer2022digital, thomas2022digital}}, all of which provide only slide-level labels. 
Following the approach of CONCH~\cite{lu2023towards} and MI-Zero~\cite{lu2023visual}, we employed a top-$K$ pooling operator for this classification task. 
\textit{More details are provided in supplementary material.}
\vspace{-.2cm}

\subsection{SOTA Methods for Comparison}
\vspace{-.1cm}
We assessed the performance of our MR-PLIP algorithm against \textcolor{black}{seven} other SOTA methods in zero-shot classification task. 
These methods include CLIP~\cite{radford2021learning}, PLIP~\cite{huang2023visual}, MI-Zero~\cite{lu2023visual}, BioCLIP \cite{zhang2024biomedclip}, CONCH~\cite{lu2023towards}, QuiltNet~\cite{ikezogwo2024quilt}, \textcolor{black}{and CPLIP \cite{javed2024cplip}}. 
In addition to comparing with SOTA CPath VLMs, our comparison extends to SOTA \textcolor{black}{vision only foundation models} based on SSL that use only an image encoder, namely CTransPath~\cite{wang2022transformer}, DinoSSLPath~\cite{kang2023benchmarking}, \textcolor{black}{UNI \cite{chen2024towards}, CHIEF \cite{wang2024pathology}, GigaPath \cite{xu2024whole}, Virchow \cite{vorontsov2024foundation}, and REMEDIS \cite{azizi2023robust}.}
For a fair comparison, we used the official source codes for all methods, maintaining consistent settings for testing splits and inference prompts. 
Our zero-shot evaluation was conducted using the testing split for each dataset, while for linear and full-fine-tuning evaluations (see supplementary material), we used both the training and testing splits of each dataset \textcolor{black}{as suggested in UNI \cite{chen2024towards}}.

\vspace{-.3cm}
\subsection{Zero-shot Transfer to Histopathology Images}
\vspace{-.2cm}
We conduct zero-shot evaluations on various datasets with our trained vision, text, and multi-modal encoders. 
For a specific test histology patch, we introduce our zero-shot classification approach.
We first select $k_{o}$ positive keywords from a dictionary of textual descriptions collected during the training process of the CVTA module (Sec. \ref{align}).
Then, these positive keywords and the visual features are input to the multi-modal encoder to obtain \textcolor{black}{text-guided} visual features, which are then used to match with the testing prompts to predict the class label.
\textcolor{black}{We use similar dataset-specific testing prompts as suggested in CONCH \cite{lu2023towards}, and QuiltNet \cite{ikezogwo2024quilt} and reported the performance in terms of with and without prompts ensembling.}\\
\noindent \textbf{Tile-level Classification Results:} 
Table~\ref{table1} displays the results of zero-shot tile-level classification across \textcolor{black}{17} datasets, evaluating the weighted average $F_{1}$ score and comparing it against \textcolor{black}{seven} leading CPath VLMs. 
Our MR-PLIP \textcolor{black}{model} has consistently outperformed all other methods across these datasets by significant margins \textcolor{black}{except WSSS4LUAD in prompts ensembling evaluation.
CPLIP emerged as the runner-up in all datasets, while CONCH and QuiltNet claimed the second positions in three datasets. }
These results demonstrate the effectiveness of incorporating multi-resolution image-text pairings in our MR-PLIP algorithm.

\noindent \textbf{WSI-level Classification Results:} 
We conduct WSI-level classification in line with \textcolor{black}{other SOTA methods}. 
Tiles are extracted from the tissue sections of the test WSI, and their embeddings are obtained via our trained vision encoder. 
These are then matched with testing prompts to determine the class label for each tile. 
For the WSI-level label, we compile similarity scores across tiles using top-1, 5, 10, 50, and 100 pooling operators. 
The class that achieves the highest cumulative score is assigned as the WSI class. 
Table~\ref{table1} (last \textcolor{black}{four} rows) presents the zero-shot WSI-level classification results on \textcolor{black}{four} datasets and benchmarks them against \textcolor{black}{seven} SOTA methods. Across all datasets, our MR-PLIP algorithm consistently outperforms other methods by a substantial margin, affirming the advanced performance of the MR-PLIP model over the second-ranking CPLIP.
\vspace{-3mm}

\subsection{Linear Probe Evaluations}
\vspace{-2mm}
Following the approach of SOTA VLMs in CPath \cite{lu2023visual,ikezogwo2024quilt}, we conducted downstream analysis by freezing the weights of our proposed MR-PLIP model and subsequently training linear layers for supervised tile-level classification tasks. 
We obtained text-guided visual features by inputting an image alongside its most closely matching text description from a predefined prompt set. 
A downstream linear classifier was then trained on these features to evaluate the quality of the representations learned by our MR-PLIP model. 
For the linear probe, we used the L-BFGS solver \cite{zhu1997algorithm} with a maximum of 1K iterations. 
Table \ref{table6_new} presents a comparison of MR-PLIP with SOTA foundation models on eight different tile-level datasets. 
MR-PLIP achieved the best performance across all datasets except for WILDS-CAM17, where UNI remains the top performer. 
This underscores the advantages of our proposed MR-PLIP model.
\vspace{-2mm}
\subsection{Weakly Supervised Evaluations}
\vspace{-2mm}
We also evaluated the text-guided visual representations learned by MR-PLIP in weakly-supervised settings across seven diverse WSI classification datasets. 
MR-PLIP was used to extract text-guided visual features from each patch, after which the ABMIL method \cite{ilse2018attention} was employed for feature aggregation and MIL-based classification, as done in other SOTA methods \cite{wang2022transformer, wang2024pathology, azizi2023robust, vorontsov2024foundation, chen2024towards}.
For training, we used the AdamW optimizer with a cosine learning rate scheduler, a learning rate of $1 \times 10^{-4}$, cross-entropy loss, and a maximum of 20 epochs. 
To ensure fair comparisons, we followed the experimental protocols of existing SOTA methods for WSI classification tasks \cite{chen2024towards}.
If official data folds were not available, the WSI datasets were case-stratified and label-stratified into train-validation-test splits as suggested by UNI \cite{chen2024towards}. 
Table \ref{table6_new} compares MR-PLIP with other foundation models based on balanced accuracy and $F_{1}$ score. 
Our results show that MR-PLIP outperforms existing models by a significant margin, highlighting the benefits of explicitly incorporating multi-resolution image-text features.
\vspace{-3mm}
\subsection{Ablation Studies}
\vspace{-2mm}

\begin{table}[t!]
\centering
\caption{Ablation study of MR-PLIP's zero-shot classification performance (weighted $F_{1}$ score) for various loss term combinations.}
\vspace{-.2cm}
\scalebox{0.58}{
\begin{tabular}{ |c|c|c|c|c|c|c|}
\hline
Loss&CAM16&CPTAC&SICAP&DigestPath&Databiox&NCT-CRC \\
\hline
1. $\mathcal{L}_{bl}$&0.567&0.796&0.487&0.831&0.429&0.786 \\
2. $\mathcal{L}_{bl}+\mathcal{L}_{CVTA}$&\underline{0.614}&0.812&0.518&0.873&\underline{0.466}&0.829\\
3. $\mathcal{L}_{bl}+\mathcal{L}_{MRTVA}$&0.599&\underline{0.846}&\underline{0.527}&\underline{0.881}&0.451&\underline{0.851}\\
4. $\mathcal{L}_{bl}+\mathcal{L}_{CVTA}+\mathcal{L}_{MRTVA}$&\textbf{0.664}&\textbf{0.875}&\textbf{0.546}&\textbf{0.935}&\textbf{0.532}&\textbf{0.871}\\
\hline
\end{tabular} 
}
\label{table5}
\vspace{-5mm}
\end{table}

\noindent \textbf{1. Impact of Loss Terms (Table \ref{table5})}
Table~\ref{table5} presents our analysis of the impact that each individual loss term has on the overall performance. We observed that the combination of both $\mathcal{L}_{CVTA}$ and $\mathcal{L}_{MRTVA}$ in Eq. (\ref{eqn5}) significantly improves the performance of zero-shot classification when compared to the use of either 
$\mathcal{L}_{bl}$ alone or one of the two proposed losses in isolation. 
Additionally, $\mathcal{L}_{MRTVA}$ outperforms $\mathcal{L}_{CVTA}$ in four datasets, demonstrating the effectiveness of using the multi-modal encoder. The superior performance of MRTVA can be attributed to its ability not just to merge textual and visual representations but also to align features across different resolutions in a multi-modal context, thereby boosting overall performance.

\noindent \textbf{2. Importance of Parent-Child Hierarchy (Table \ref{table9})}
An experiment was conducted without enforcing the parent-child hierarchy in Eq. (\ref{eqn3}), meaning that all parent and child representations were aligned across consecutive resolution levels without considering any hierarchy. 
Table \ref{table9} compares the zero-shot classification performance of MR-PLIP on six datasets, with and without the parent-child hierarchy. 
The results show a significant performance improvement when the parent-child hierarchy was applied. 
This is because not all child representations overlap with all parent representations; rather, only children within a specific hierarchy share content with their corresponding parents. 
Thus, aligning representations across multiple resolutions is more effective when using a parent-child hierarchy.
\textit{More ablations are provided in our supplementary material.}
\vspace{-1mm}
\subsection{Computational Time}
We implemented our model on six NVIDIA A100 GPUs.
During inference, the MR-PLIP algorithm must generate image-text embedding before matching. 
On CAM16 dataset, for WSI-level classification, MR-PLIP averaged 4.51 minutes per WSI. This is in comparison to other VLMs such as MI-Zero, which took 3.00 minutes, BioCLIP at 2.70 minutes, and PLIP at 2.90 minutes. Thus, in terms of speed at inference, MR-PLIP holds up well against other VLMs.

\begin{table}[t!]
\caption{Ablation study comparing the zero-shot classification performance of MR-PLIP in terms of weighted $F_{1}$ score, with and without enforcing the parent-child hierarchy. }
\vspace{-2mm}
\makebox[\linewidth]{
\scalebox{0.64}{
\begin{tabular}{ |c|c|c|c|c|c|c|}
\hline
Resolutions&CAM16&CPTAC&SICAP&DigestPath&Databiox&NCT-CRC\\
\hline
w Parent-Child&\textbf{0.664}&\textbf{0.875}&\textbf{0.546}&\textbf{0.935}&\textbf{0.532}&\textbf{0.871} \\
w/o Parent-Child&0.610&0.826&0.504&0.853&0.496&0.848\\
\hline
\end{tabular} 
}}
\label{table9}
\vspace{-6mm}
\end{table}

\vspace{-4mm}
\section{Conclusion}
\label{sec:conclusion}
\vspace{-2mm}
In this paper, we introduced the innovative MR-PLIP model, which matches image-text pairs across multiple resolutions, thereby incorporating both broader contexts and more detailed information than traditional single-resolution VLMs in CPath. 
We constructed two types of histology bags: a visual collection consisting of multi-resolution images linked in a hierarchical parent-child manner, and a textual bag filled with aggregated textual descriptions generated at various resolutions via SOTA Quilt-LLaVA. 
Using these bags, we developed two ground breaking modules for cross-resolution visual-textual alignment and for fusing multi-modal features to achieve cross-resolution alignment. 
The first module focuses on pinpointing positive keywords across different resolution levels to train the vision-text encoders, ensuring the image-text representations are accurately aligned. The second module trains the multi-modal encoder to initially deduce text-guided visual features, which facilitate cross-resolution alignment. 
Our pre-training encompassed 34 million image-text pairs across various resolutions. 
Through extensive testing on 26 histopathology datasets, our approach has shown superior performance over existing SOTA methods. 
In the future, we plan to adapt our model for additional medical imaging modalities, including X-rays, CT scans, and MRIs.
\vspace{-4mm}
\section{Acknowledgement}
\vspace{-1mm}
This publication acknowledges the support provided by the Advanced Research and Innovation Center (ARIC), which is jointly funded by Aerospace Holding Company LLC, a wholly-owned subsidiary of Mubadala Investment Company PJSC and Khalifa University for Science and Technology.

{
    \small
    \bibliographystyle{ieeenat_fullname}
    \bibliography{main}

\begin{thebibliography}{107}
\providecommand{\natexlab}[1]{#1}
\providecommand{\url}[1]{\texttt{#1}}
\expandafter\ifx\csname urlstyle\endcsname\relax
  \providecommand{\doi}[1]{doi: #1}\else
  \providecommand{\doi}{doi: \begingroup \urlstyle{rm}\Url}\fi

\bibitem[Abels et~al.(2019)Abels, Pantanowitz, Aeffner, Zarella, van~der Laak,
  Bui, Vemuri, Parwani, Gibbs, Agosto-Arroyo, et~al.]{abels2019computational}
Esther Abels, Liron Pantanowitz, Famke Aeffner, Mark~D Zarella, Jeroen van~der
  Laak, Marilyn~M Bui, Venkata~NP Vemuri, Anil~V Parwani, Jeff Gibbs, Emmanuel
  Agosto-Arroyo, et~al.
\newblock Computational pathology definitions, best practices, and
  recommendations for regulatory guidance: a white paper from the digital
  pathology association.
\newblock \emph{The Journal of pathology}, 249\penalty0 (3):\penalty0 286--294,
  2019.

\bibitem[Achiam et~al.(2023)Achiam, Adler, Agarwal, Ahmad, Akkaya, Aleman,
  Almeida, Altenschmidt, Altman, Anadkat, et~al.]{achiam2023gpt}
Josh Achiam, Steven Adler, Sandhini Agarwal, Lama Ahmad, Ilge Akkaya,
  Florencia~Leoni Aleman, Diogo Almeida, Janko Altenschmidt, Sam Altman,
  Shyamal Anadkat, et~al.
\newblock Gpt-4 technical report.
\newblock \emph{arXiv preprint arXiv:2303.08774}, 2023.

\bibitem[Acosta et~al.(2022)Acosta, Falcone, Rajpurkar, and
  Topol]{acosta2022multimodal}
Juli{\'a}n~N Acosta, Guido~J Falcone, Pranav Rajpurkar, and Eric~J Topol.
\newblock Multimodal biomedical ai.
\newblock \emph{Nature Medicine}, 28\penalty0 (9):\penalty0 1773--1784, 2022.

\bibitem[Adam et~al.(2014)]{adam2014method}
Kingma DP Ba~J Adam et~al.
\newblock A method for stochastic optimization.
\newblock \emph{arXiv preprint arXiv:1412.6980}, 1412, 2014.

\bibitem[Allen(2014)]{allen2014social}
Timothy~Craig Allen.
\newblock Social media: pathologists' force multiplier.
\newblock \emph{Archives of Pathology and Laboratory Medicine}, 138\penalty0
  (8):\penalty0 1000--1001, 2014.

\bibitem[Alsentzer et~al.(2019)Alsentzer, Murphy, Boag, Weng, Jin, Naumann, and
  McDermott]{alsentzer2019publicly}
Emily Alsentzer, John~R Murphy, Willie Boag, Wei-Hung Weng, Di Jin, Tristan
  Naumann, and Matthew McDermott.
\newblock Publicly available clinical bert embeddings.
\newblock \emph{arXiv preprint arXiv:1904.03323}, 2019.

\bibitem[Aresta et~al.(2019)Aresta, Ara{\'u}jo, Kwok, Chennamsetty, Safwan,
  Alex, Marami, Prastawa, Chan, Donovan, et~al.]{aresta2019bach}
Guilherme Aresta, Teresa Ara{\'u}jo, Scotty Kwok, Sai~Saketh Chennamsetty,
  Mohammed Safwan, Varghese Alex, Bahram Marami, Marcel Prastawa, Monica Chan,
  Michael Donovan, et~al.
\newblock Bach: Grand challenge on breast cancer histology images.
\newblock \emph{Medical image analysis}, 56:\penalty0 122--139, 2019.

\bibitem[Arunachalam et~al.(2019)Arunachalam, Mishra, Daescu, Cederberg,
  Rakheja, Sengupta, Leonard, Hallac, and Leavey]{arunachalam2019viable}
Harish~Babu Arunachalam, Rashika Mishra, Ovidiu Daescu, Kevin Cederberg, Dinesh
  Rakheja, Anita Sengupta, David Leonard, Rami Hallac, and Patrick Leavey.
\newblock Viable and necrotic tumor assessment from whole slide images of
  osteosarcoma using machine-learning and deep-learning models.
\newblock \emph{PloS one}, 14\penalty0 (4):\penalty0 e0210706, 2019.

\bibitem[Azizi et~al.(2023)Azizi, Culp, Freyberg, Mustafa, Baur, Kornblith,
  Chen, Tomasev, Mitrovi{\'c}, Strachan, et~al.]{azizi2023robust}
Shekoofeh Azizi, Laura Culp, Jan Freyberg, Basil Mustafa, Sebastien Baur, Simon
  Kornblith, Ting Chen, Nenad Tomasev, Jovana Mitrovi{\'c}, Patricia Strachan,
  et~al.
\newblock Robust and data-efficient generalization of self-supervised machine
  learning for diagnostic imaging.
\newblock \emph{Nature Biomedical Engineering}, 7\penalty0 (6):\penalty0
  756--779, 2023.

\bibitem[Bandi et~al.(2018)Bandi, Geessink, Manson, Van~Dijk, Balkenhol,
  Hermsen, Bejnordi, Lee, Paeng, Zhong, et~al.]{bandi2018detection}
Peter Bandi, Oscar Geessink, Quirine Manson, Marcory Van~Dijk, Maschenka
  Balkenhol, Meyke Hermsen, Babak~Ehteshami Bejnordi, Byungjae Lee, Kyunghyun
  Paeng, Aoxiao Zhong, et~al.
\newblock From detection of individual metastases to classification of lymph
  node status at the patient level: the camelyon17 challenge.
\newblock \emph{IEEE transactions on medical imaging}, 38\penalty0
  (2):\penalty0 550--560, 2018.

\bibitem[Barbano et~al.(2021)Barbano, Perlo, Tartaglione, Fiandrotti, Bertero,
  Cassoni, and Grangetto]{barbano2021unitopatho}
Carlo~Alberto Barbano, Daniele Perlo, Enzo Tartaglione, Attilio Fiandrotti,
  Luca Bertero, Paola Cassoni, and Marco Grangetto.
\newblock Unitopatho, a labeled histopathological dataset for colorectal polyps
  classification and adenoma dysplasia grading.
\newblock In \emph{2021 IEEE International Conference on Image Processing
  (ICIP)}, pages 76--80. IEEE, 2021.

\bibitem[Bejnordi et~al.(2015)Bejnordi, Litjens, Hermsen, Karssemeijer, and
  van~der Laak]{bejnordi2015multi}
Babak~Ehteshami Bejnordi, Geert Litjens, Meyke Hermsen, Nico Karssemeijer, and
  Jeroen~AWM van~der Laak.
\newblock A multi-scale superpixel classification approach to the detection of
  regions of interest in whole slide histopathology images.
\newblock In \emph{Medical Imaging 2015: Digital Pathology}, pages 99--104.
  SPIE, 2015.

\bibitem[Bejnordi et~al.(2017)Bejnordi, Veta, Van~Diest, Van~Ginneken,
  Karssemeijer, Litjens, Van Der~Laak, Hermsen, Manson, Balkenhol,
  et~al.]{bejnordi2017diagnostic}
Babak~Ehteshami Bejnordi, Mitko Veta, Paul~Johannes Van~Diest, Bram
  Van~Ginneken, Nico Karssemeijer, Geert Litjens, Jeroen~AWM Van Der~Laak,
  Meyke Hermsen, Quirine~F Manson, Maschenka Balkenhol, et~al.
\newblock Diagnostic assessment of deep learning algorithms for detection of
  lymph node metastases in women with breast cancer.
\newblock \emph{Jama}, 318\penalty0 (22):\penalty0 2199--2210, 2017.

\bibitem[Berb{\'\i}s et~al.(2023)Berb{\'\i}s, McClintock, Bychkov, Van~der
  Laak, Pantanowitz, Lennerz, Cheng, Delahunt, Egevad, Eloy,
  et~al.]{berbis2023computational}
M~Alvaro Berb{\'\i}s, David~S McClintock, Andrey Bychkov, Jeroen Van~der Laak,
  Liron Pantanowitz, Jochen~K Lennerz, Jerome~Y Cheng, Brett Delahunt, Lars
  Egevad, Catarina Eloy, et~al.
\newblock Computational pathology in 2030: a delphi study forecasting the role
  of ai in pathology within the next decade.
\newblock \emph{EBioMedicine}, 88, 2023.

\bibitem[Bi et~al.(2020)Bi, Li, Wu, Yan, Wang, Huang, Huang, and
  Si]{bi2020palm}
Bin Bi, Chenliang Li, Chen Wu, Ming Yan, Wei Wang, Songfang Huang, Fei Huang,
  and Luo Si.
\newblock Palm: Pre-training an autoencoding\&autoregressive language model for
  context-conditioned generation.
\newblock \emph{arXiv preprint arXiv:2004.07159}, 2020.

\bibitem[Bolhasani et~al.(2020)Bolhasani, Amjadi, Tabatabaeian, and
  Jassbi]{bolhasani2020histopathological}
Hamidreza Bolhasani, Elham Amjadi, Maryam Tabatabaeian, and Somayyeh~Jafarali
  Jassbi.
\newblock A histopathological image dataset for grading breast invasive ductal
  carcinomas.
\newblock \emph{Informatics in Medicine Unlocked}, 19:\penalty0 100341, 2020.

\bibitem[Borkowski et~al.(2019)Borkowski, Bui, Thomas, Wilson, DeLand, and
  Mastorides]{borkowski2019lung}
Andrew~A Borkowski, Marilyn~M Bui, L~Brannon Thomas, Catherine~P Wilson,
  Lauren~A DeLand, and Stephen~M Mastorides.
\newblock Lung and colon cancer histopathological image dataset (lc25000).
\newblock \emph{arXiv preprint arXiv:1912.12142}, 2019.

\bibitem[Brancati et~al.(2022)Brancati, Anniciello, Pati, Riccio, Scognamiglio,
  Jaume, De~Pietro, Di~Bonito, Foncubierta, Botti, et~al.]{brancati2022bracs}
Nadia Brancati, Anna~Maria Anniciello, Pushpak Pati, Daniel Riccio, Giosu{\`e}
  Scognamiglio, Guillaume Jaume, Giuseppe De~Pietro, Maurizio Di~Bonito,
  Antonio Foncubierta, Gerardo Botti, et~al.
\newblock Bracs: A dataset for breast carcinoma subtyping in h\&e histology
  images.
\newblock \emph{Database}, 2022:\penalty0 baac093, 2022.

\bibitem[Brummer et~al.(2022)Brummer, P{\"o}l{\"o}nen, Mustjoki, and
  Br{\"u}ck]{brummer2022integrative}
Otso Brummer, Petri P{\"o}l{\"o}nen, Satu Mustjoki, and Oscar Br{\"u}ck.
\newblock Integrative analysis of histological textures and lymphocyte
  infiltration in renal cell carcinoma using deep learning.
\newblock \emph{bioRxiv}, pages 2022--08, 2022.

\bibitem[Bulten et~al.(2022)Bulten, Kartasalo, Chen, Str{\"o}m, Pinckaers,
  Nagpal, Cai, Steiner, Van~Boven, Vink, et~al.]{bulten2022artificial}
Wouter Bulten, Kimmo Kartasalo, Po-Hsuan~Cameron Chen, Peter Str{\"o}m, Hans
  Pinckaers, Kunal Nagpal, Yuannan Cai, David~F Steiner, Hester Van~Boven,
  Robert Vink, et~al.
\newblock Artificial intelligence for diagnosis and gleason grading of prostate
  cancer: the panda challenge.
\newblock \emph{Nature medicine}, 28\penalty0 (1):\penalty0 154--163, 2022.

\bibitem[Campanella et~al.(2019)Campanella, Hanna, Geneslaw, Miraflor, Werneck
  Krauss~Silva, Busam, Brogi, Reuter, Klimstra, and
  Fuchs]{campanella2019clinical}
Gabriele Campanella, Matthew~G Hanna, Luke Geneslaw, Allen Miraflor, Vitor
  Werneck Krauss~Silva, Klaus~J Busam, Edi Brogi, Victor~E Reuter, David~S
  Klimstra, and Thomas~J Fuchs.
\newblock Clinical-grade computational pathology using weakly supervised deep
  learning on whole slide images.
\newblock \emph{Nature medicine}, 25\penalty0 (8):\penalty0 1301--1309, 2019.

\bibitem[Caron et~al.(2021)Caron, Touvron, Misra, J{\'e}gou, Mairal,
  Bojanowski, and Joulin]{caron2021emerging}
Mathilde Caron, Hugo Touvron, Ishan Misra, Herv{\'e} J{\'e}gou, Julien Mairal,
  Piotr Bojanowski, and Armand Joulin.
\newblock Emerging properties in self-supervised vision transformers.
\newblock In \emph{Proceedings of the IEEE/CVF international conference on
  computer vision}, pages 9650--9660, 2021.

\bibitem[Chanda et~al.(2024)Chanda, Aryal, Soltani, and Ganji]{chanda2024new}
Dibaloke Chanda, Milan Aryal, Nasim~Yahya Soltani, and Masoud Ganji.
\newblock A new era in computational pathology: A survey on foundation and
  vision-language models.
\newblock \emph{arXiv preprint arXiv:2408.14496}, 2024.

\bibitem[Chen et~al.(2022{\natexlab{a}})Chen, Lu, Williamson, Chen, Schaumberg,
  and Mahmood]{chen2022fast}
Chengkuan Chen, Ming~Y Lu, Drew~FK Williamson, Tiffany~Y Chen, Andrew~J
  Schaumberg, and Faisal Mahmood.
\newblock Fast and scalable search of whole-slide images via self-supervised
  deep learning.
\newblock \emph{Nature Biomedical Engineering}, 6\penalty0 (12):\penalty0
  1420--1434, 2022{\natexlab{a}}.

\bibitem[Chen et~al.(2022{\natexlab{b}})Chen, Chen, Li, Chen, Trister,
  Krishnan, and Mahmood]{chen2022scaling}
Richard~J Chen, Chengkuan Chen, Yicong Li, Tiffany~Y Chen, Andrew~D Trister,
  Rahul~G Krishnan, and Faisal Mahmood.
\newblock Scaling vision transformers to gigapixel images via hierarchical
  self-supervised learning.
\newblock In \emph{Proceedings of the IEEE/CVF Conference on Computer Vision
  and Pattern Recognition}, pages 16144--16155, 2022{\natexlab{b}}.

\bibitem[Chen et~al.(2024)Chen, Ding, Lu, Williamson, Jaume, Song, Chen, Zhang,
  Shao, Shaban, et~al.]{chen2024towards}
Richard~J Chen, Tong Ding, Ming~Y Lu, Drew~FK Williamson, Guillaume Jaume,
  Andrew~H Song, Bowen Chen, Andrew Zhang, Daniel Shao, Muhammad Shaban, et~al.
\newblock Towards a general-purpose foundation model for computational
  pathology.
\newblock \emph{Nature Medicine}, 30\penalty0 (3):\penalty0 850--862, 2024.

\bibitem[Chen et~al.(2020)Chen, Kornblith, Norouzi, and Hinton]{chen2020simple}
Ting Chen, Simon Kornblith, Mohammad Norouzi, and Geoffrey Hinton.
\newblock A simple framework for contrastive learning of visual
  representations.
\newblock In \emph{International conference on machine learning}, pages
  1597--1607. PMLR, 2020.

\bibitem[Chen and He(2021)]{chen2021exploring}
Xinlei Chen and Kaiming He.
\newblock Exploring simple siamese representation learning.
\newblock In \emph{Proceedings of the IEEE/CVF conference on computer vision
  and pattern recognition}, pages 15750--15758, 2021.

\bibitem[Chen et~al.()Chen, Xie, and He]{chenempirical}
X Chen, S Xie, and K He.
\newblock An empirical study of training self-supervised vision transformers.
  in 2021 ieee.
\newblock In \emph{CVF International Conference on Computer Vision (ICCV)},
  pages 9620--9629.

\bibitem[Cifci et~al.(2023)Cifci, Veldhuizen, Foersch, and Kather]{cifci2023ai}
Didem Cifci, Gregory~P Veldhuizen, Sebastian Foersch, and Jakob~Nikolas Kather.
\newblock Ai in computational pathology of cancer: Improving diagnostic
  workflows and clinical outcomes?
\newblock \emph{Annual Review of Cancer Biology}, 7:\penalty0 57--71, 2023.

\bibitem[Cornish et~al.(2012)Cornish, Swapp, and Kaplan]{cornish2012whole}
Toby~C Cornish, Ryan~E Swapp, and Keith~J Kaplan.
\newblock Whole-slide imaging: routine pathologic diagnosis.
\newblock \emph{Advances in anatomic pathology}, 19\penalty0 (3):\penalty0
  152--159, 2012.

\bibitem[Cruz-Roa et~al.(2017)Cruz-Roa, Gilmore, Basavanhally, Feldman,
  Ganesan, Shih, Tomaszewski, Gonz{\'a}lez, and Madabhushi]{cruz2017accurate}
Angel Cruz-Roa, Hannah Gilmore, Ajay Basavanhally, Michael Feldman, Shridar
  Ganesan, Natalie~NC Shih, John Tomaszewski, Fabio~A Gonz{\'a}lez, and Anant
  Madabhushi.
\newblock Accurate and reproducible invasive breast cancer detection in
  whole-slide images: A deep learning approach for quantifying tumor extent.
\newblock \emph{Scientific reports}, 7\penalty0 (1):\penalty0 46450, 2017.

\bibitem[Cui and Zhang(2021)]{cui2021artificial}
Miao Cui and David~Y Zhang.
\newblock Artificial intelligence and computational pathology.
\newblock \emph{Laboratory Investigation}, 101\penalty0 (4):\penalty0 412--422,
  2021.

\bibitem[Da et~al.(2022)Da, Huang, Li, Zuo, Zhang, Liu, Chen, Li, Xu, Hu,
  et~al.]{da2022digestpath}
Qian Da, Xiaodi Huang, Zhongyu Li, Yanfei Zuo, Chenbin Zhang, Jingxin Liu, Wen
  Chen, Jiahui Li, Dou Xu, Zhiqiang Hu, et~al.
\newblock Digestpath: A benchmark dataset with challenge review for the
  pathological detection and segmentation of digestive-system.
\newblock \emph{Medical Image Analysis}, 80:\penalty0 102485, 2022.

\bibitem[Devlin et~al.(2018)Devlin, Chang, Lee, and Toutanova]{devlin2018bert}
Jacob Devlin, Ming-Wei Chang, Kenton Lee, and Kristina Toutanova.
\newblock Bert: Pre-training of deep bidirectional transformers for language
  understanding.
\newblock \emph{arXiv preprint arXiv:1810.04805}, 2018.

\bibitem[Ding et~al.(2020)Ding, Pan, Cen, Li, and Chen]{ding2020multi}
Huijun Ding, Zhanpeng Pan, Qian Cen, Yang Li, and Shifeng Chen.
\newblock Multi-scale fully convolutional network for gland segmentation using
  three-class classification.
\newblock \emph{Neurocomputing}, 380:\penalty0 150--161, 2020.

\bibitem[Dosovitskiy et~al.(2020)Dosovitskiy, Beyer, Kolesnikov, Weissenborn,
  Zhai, Unterthiner, Dehghani, Minderer, Heigold, Gelly,
  et~al.]{dosovitskiy2020image}
Alexey Dosovitskiy, Lucas Beyer, Alexander Kolesnikov, Dirk Weissenborn,
  Xiaohua Zhai, Thomas Unterthiner, Mostafa Dehghani, Matthias Minderer, Georg
  Heigold, Sylvain Gelly, et~al.
\newblock An image is worth 16x16 words: Transformers for image recognition at
  scale.
\newblock \emph{arXiv preprint arXiv:2010.11929}, 2020.

\bibitem[Echle et~al.(2021)Echle, Rindtorff, Brinker, Luedde, Pearson, and
  Kather]{echle2021deep}
Amelie Echle, Niklas~Timon Rindtorff, Titus~Josef Brinker, Tom Luedde,
  Alexander~Thomas Pearson, and Jakob~Nikolas Kather.
\newblock Deep learning in cancer pathology: a new generation of clinical
  biomarkers.
\newblock \emph{British journal of cancer}, 124\penalty0 (4):\penalty0
  686--696, 2021.

\bibitem[Edwards et~al.(2015)Edwards, Oberti, Thangudu, Cai, McGarvey, Jacob,
  Madhavan, and Ketchum]{edwards2015cptac}
Nathan~J Edwards, Mauricio Oberti, Ratna~R Thangudu, Shuang Cai, Peter~B
  McGarvey, Shine Jacob, Subha Madhavan, and Karen~A Ketchum.
\newblock The cptac data portal: a resource for cancer proteomics research.
\newblock \emph{Journal of proteome research}, 14\penalty0 (6):\penalty0
  2707--2713, 2015.

\bibitem[Gamper and Rajpoot(2021)]{gamper2021multiple}
Jevgenij Gamper and Nasir Rajpoot.
\newblock Multiple instance captioning: Learning representations from
  histopathology textbooks and articles.
\newblock In \emph{Proceedings of the IEEE/CVF conference on computer vision
  and pattern recognition}, pages 16549--16559, 2021.

\bibitem[Gamper et~al.(2019)Gamper, Alemi~Koohbanani, Benet, Khuram, and
  Rajpoot]{gamper2019pannuke}
Jevgenij Gamper, Navid Alemi~Koohbanani, Ksenija Benet, Ali Khuram, and Nasir
  Rajpoot.
\newblock Pannuke: an open pan-cancer histology dataset for nuclei instance
  segmentation and classification.
\newblock In \emph{Digital Pathology: 15th European Congress, ECDP 2019,
  Warwick, UK, April 10--13, 2019, Proceedings 15}, pages 11--19. Springer,
  2019.

\bibitem[Graham et~al.(2019)Graham, Vu, Raza, Azam, Tsang, Kwak, and
  Rajpoot]{graham2019hover}
Simon Graham, Quoc~Dang Vu, Shan E~Ahmed Raza, Ayesha Azam, Yee~Wah Tsang,
  Jin~Tae Kwak, and Nasir Rajpoot.
\newblock Hover-net: Simultaneous segmentation and classification of nuclei in
  multi-tissue histology images.
\newblock \emph{Medical image analysis}, 58:\penalty0 101563, 2019.

\bibitem[Grill et~al.(2020)Grill, Strub, Altch{\'e}, Tallec, Richemond,
  Buchatskaya, Doersch, Avila~Pires, Guo, Gheshlaghi~Azar,
  et~al.]{grill2020bootstrap}
Jean-Bastien Grill, Florian Strub, Florent Altch{\'e}, Corentin Tallec, Pierre
  Richemond, Elena Buchatskaya, Carl Doersch, Bernardo Avila~Pires, Zhaohan
  Guo, Mohammad Gheshlaghi~Azar, et~al.
\newblock Bootstrap your own latent-a new approach to self-supervised learning.
\newblock \emph{Advances in neural information processing systems},
  33:\penalty0 21271--21284, 2020.

\bibitem[Gu et~al.(2021)Gu, Tinn, Cheng, Lucas, Usuyama, Liu, Naumann, Gao, and
  Poon]{gu2021domain}
Yu Gu, Robert Tinn, Hao Cheng, Michael Lucas, Naoto Usuyama, Xiaodong Liu,
  Tristan Naumann, Jianfeng Gao, and Hoifung Poon.
\newblock Domain-specific language model pretraining for biomedical natural
  language processing.
\newblock \emph{ACM Transactions on Computing for Healthcare (HEALTH)},
  3\penalty0 (1):\penalty0 1--23, 2021.

\bibitem[Han et~al.(2022)Han, Pan, Yan, Lin, Li, Yao, Lv, Shi, Mai, Lin,
  et~al.]{han2022wsss4luad}
Chu Han, Xipeng Pan, Lixu Yan, Huan Lin, Bingbing Li, Su Yao, Shanshan Lv,
  Zhenwei Shi, Jinhai Mai, Jiatai Lin, et~al.
\newblock Wsss4luad: Grand challenge on weakly-supervised tissue semantic
  segmentation for lung adenocarcinoma.
\newblock \emph{arXiv preprint arXiv:2204.06455}, 2022.

\bibitem[Hanna et~al.(2020)Hanna, Parwani, and Sirintrapun]{hanna2020whole}
Matthew~G Hanna, Anil Parwani, and Sahussapont~Joseph Sirintrapun.
\newblock Whole slide imaging: technology and applications.
\newblock \emph{Advances in Anatomic Pathology}, 27\penalty0 (4):\penalty0
  251--259, 2020.

\bibitem[Hashimoto et~al.(2020)Hashimoto, Fukushima, Koga, Takagi, Ko, Kohno,
  Nakaguro, Nakamura, Hontani, and Takeuchi]{hashimoto2020multi}
Noriaki Hashimoto, Daisuke Fukushima, Ryoichi Koga, Yusuke Takagi, Kaho Ko, Kei
  Kohno, Masato Nakaguro, Shigeo Nakamura, Hidekata Hontani, and Ichiro
  Takeuchi.
\newblock Multi-scale domain-adversarial multiple-instance cnn for cancer
  subtype classification with unannotated histopathological images.
\newblock In \emph{Proceedings of the IEEE/CVF conference on computer vision
  and pattern recognition}, pages 3852--3861, 2020.

\bibitem[Hosseini et~al.(2024)Hosseini, Bejnordi, Trinh, Chan, Hasan, Li, Yang,
  Kim, Zhang, Wu, et~al.]{hosseini2024computational}
Mahdi~S Hosseini, Babak~Ehteshami Bejnordi, Vincent Quoc-Huy Trinh, Lyndon
  Chan, Danial Hasan, Xingwen Li, Stephen Yang, Taehyo Kim, Haochen Zhang,
  Theodore Wu, et~al.
\newblock Computational pathology: a survey review and the way forward.
\newblock \emph{Journal of Pathology Informatics}, page 100357, 2024.

\bibitem[Huang et~al.(2023)Huang, Bianchi, Yuksekgonul, Montine, and
  Zou]{huang2023visual}
Zhi Huang, Federico Bianchi, Mert Yuksekgonul, Thomas~J Montine, and James Zou.
\newblock A visual--language foundation model for pathology image analysis
  using medical twitter.
\newblock \emph{Nature medicine}, 29\penalty0 (9):\penalty0 2307--2316, 2023.

\bibitem[ICIAR(2018)]{iciar2018grand}
BACH ICIAR.
\newblock Grand challenge on breast cancer histology images. 2018, 2018.

\bibitem[Ikezogwo et~al.(2024)Ikezogwo, Seyfioglu, Ghezloo, Geva,
  Sheikh~Mohammed, Anand, Krishna, and Shapiro]{ikezogwo2024quilt}
Wisdom Ikezogwo, Saygin Seyfioglu, Fatemeh Ghezloo, Dylan Geva, Fatwir
  Sheikh~Mohammed, Pavan~Kumar Anand, Ranjay Krishna, and Linda Shapiro.
\newblock Quilt-1m: One million image-text pairs for histopathology.
\newblock \emph{Advances in Neural Information Processing Systems}, 36, 2024.

\bibitem[Ilse et~al.(2018)Ilse, Tomczak, and Welling]{ilse2018attention}
Maximilian Ilse, Jakub Tomczak, and Max Welling.
\newblock Attention-based deep multiple instance learning.
\newblock In \emph{International conference on machine learning}, pages
  2127--2136. PMLR, 2018.

\bibitem[Javed et~al.(2024)Javed, Mahmood, Ganapathi, Dharejo, Werghi, and
  Bennamoun]{javed2024cplip}
Sajid Javed, Arif Mahmood, Iyyakutti~Iyappan Ganapathi, Fayaz~Ali Dharejo,
  Naoufel Werghi, and Mohammed Bennamoun.
\newblock Cplip: Zero-shot learning for histopathology with comprehensive
  vision-language alignment.
\newblock In \emph{Proceedings of the IEEE/CVF Conference on Computer Vision
  and Pattern Recognition}, pages 11450--11459, 2024.

\bibitem[Johnson et~al.(2016)Johnson, Pollard, Shen, Lehman, Feng, Ghassemi,
  Moody, Szolovits, Anthony~Celi, and Mark]{johnson2016mimic}
Alistair~EW Johnson, Tom~J Pollard, Lu Shen, Li-wei~H Lehman, Mengling Feng,
  Mohammad Ghassemi, Benjamin Moody, Peter Szolovits, Leo Anthony~Celi, and
  Roger~G Mark.
\newblock Mimic-iii, a freely accessible critical care database.
\newblock \emph{Scientific data}, 3\penalty0 (1):\penalty0 1--9, 2016.

\bibitem[Kang et~al.(2023)Kang, Song, Park, Yoo, and
  Pereira]{kang2023benchmarking}
Mingu Kang, Heon Song, Seonwook Park, Donggeun Yoo, and S{\'e}rgio Pereira.
\newblock Benchmarking self-supervised learning on diverse pathology datasets.
\newblock In \emph{Proceedings of the IEEE/CVF Conference on Computer Vision
  and Pattern Recognition}, pages 3344--3354, 2023.

\bibitem[Kapse et~al.(2024)Kapse, Pati, Das, Zhang, Chen, Vakalopoulou, Saltz,
  Samaras, Gupta, and Prasanna]{kapse2024si}
Saarthak Kapse, Pushpak Pati, Srijan Das, Jingwei Zhang, Chao Chen, Maria
  Vakalopoulou, Joel Saltz, Dimitris Samaras, Rajarsi~R Gupta, and Prateek
  Prasanna.
\newblock Si-mil: Taming deep mil for self-interpretability in gigapixel
  histopathology.
\newblock In \emph{Proceedings of the IEEE/CVF Conference on Computer Vision
  and Pattern Recognition}, pages 11226--11237, 2024.

\bibitem[Kather et~al.(2018)Kather, Halama, and Marx]{kather2018100}
Jakob~Nikolas Kather, Niels Halama, and Alexander Marx.
\newblock 100,000 histological images of human colorectal cancer and healthy
  tissue.
\newblock \emph{Zenodo10}, 5281, 2018.

\bibitem[Kim et~al.(2022)Kim, Kang, Song, and Kim]{kim2022application}
Inho Kim, Kyungmin Kang, Youngjae Song, and Tae-Jung Kim.
\newblock Application of artificial intelligence in pathology: Trends and
  challenges.
\newblock \emph{Diagnostics}, 12\penalty0 (11):\penalty0 2794, 2022.

\bibitem[Koh et~al.(2021)Koh, Sagawa, Marklund, Xie, Zhang, Balsubramani, Hu,
  Yasunaga, Phillips, Gao, et~al.]{koh2021wilds}
Pang~Wei Koh, Shiori Sagawa, Henrik Marklund, Sang~Michael Xie, Marvin Zhang,
  Akshay Balsubramani, Weihua Hu, Michihiro Yasunaga, Richard~Lanas Phillips,
  Irena Gao, et~al.
\newblock Wilds: A benchmark of in-the-wild distribution shifts.
\newblock In \emph{International conference on machine learning}, pages
  5637--5664. PMLR, 2021.

\bibitem[Kriegsmann et~al.(2022)Kriegsmann, Lobers, Zgorzelski, Kriegsmann,
  Janssen, Meliss, Muley, Sack, Steinbuss, and Kriegsmann]{kriegsmann2022deep}
Katharina Kriegsmann, Frithjof Lobers, Christiane Zgorzelski, Joerg Kriegsmann,
  Charlotte Janssen, Rolf~Ruedinger Meliss, Thomas Muley, Ulrich Sack, Georg
  Steinbuss, and Mark Kriegsmann.
\newblock Deep learning for the detection of anatomical tissue structures and
  neoplasms of the skin on scanned histopathological tissue sections.
\newblock \emph{Frontiers in Oncology}, 12:\penalty0 1022967, 2022.

\bibitem[Li et~al.(2021{\natexlab{a}})Li, Li, and Eliceiri]{li2021dual}
Bin Li, Yin Li, and Kevin~W Eliceiri.
\newblock Dual-stream multiple instance learning network for whole slide image
  classification with self-supervised contrastive learning.
\newblock In \emph{Proceedings of the IEEE/CVF conference on computer vision
  and pattern recognition}, pages 14318--14328, 2021{\natexlab{a}}.

\bibitem[Li et~al.(2022{\natexlab{a}})Li, Xu, Tian, Wang, Yan, Bi, Ye, Chen,
  Xu, Cao, et~al.]{li2022mplug}
Chenliang Li, Haiyang Xu, Junfeng Tian, Wei Wang, Ming Yan, Bin Bi, Jiabo Ye,
  Hehong Chen, Guohai Xu, Zheng Cao, et~al.
\newblock mplug: Effective and efficient vision-language learning by
  cross-modal skip-connections.
\newblock \emph{arXiv preprint arXiv:2205.12005}, 2022{\natexlab{a}}.

\bibitem[Li et~al.(2022{\natexlab{b}})Li, Li, Li, Niebles, and
  Hoi]{li2022align}
Dongxu Li, Junnan Li, Hongdong Li, Juan~Carlos Niebles, and Steven~CH Hoi.
\newblock Align and prompt: Video-and-language pre-training with entity
  prompts.
\newblock In \emph{Proceedings of the IEEE/CVF Conference on Computer Vision
  and Pattern Recognition}, pages 4953--4963, 2022{\natexlab{b}}.

\bibitem[Li et~al.(2024{\natexlab{a}})Li, Chen, Chen, Yu, Yang, Wang, Ding, and
  Han]{li2024generalizable}
Hao Li, Ying Chen, Yifei Chen, Rongshan Yu, Wenxian Yang, Liansheng Wang, Bowen
  Ding, and Yuchen Han.
\newblock Generalizable whole slide image classification with fine-grained
  visual-semantic interaction.
\newblock In \emph{Proceedings of the IEEE/CVF Conference on Computer Vision
  and Pattern Recognition}, pages 11398--11407, 2024{\natexlab{a}}.

\bibitem[Li et~al.(2021{\natexlab{b}})Li, Li, Sisk, Ye, Wallace, Speier, and
  Arnold]{li2021multi}
Jiayun Li, Wenyuan Li, Anthony Sisk, Huihui Ye, W~Dean Wallace, William Speier,
  and Corey~W Arnold.
\newblock A multi-resolution model for histopathology image classification and
  localization with multiple instance learning.
\newblock \emph{Computers in biology and medicine}, 131:\penalty0 104253,
  2021{\natexlab{b}}.

\bibitem[Li et~al.(2021{\natexlab{c}})Li, Selvaraju, Gotmare, Joty, Xiong, and
  Hoi]{ALBEF}
Junnan Li, Ramprasaath~R. Selvaraju, Akhilesh~Deepak Gotmare, Shafiq Joty,
  Caiming Xiong, and Steven Hoi.
\newblock Align before fuse: Vision and language representation learning with
  momentum distillation.
\newblock In \emph{NeurIPS}, 2021{\natexlab{c}}.

\bibitem[Li et~al.(2022{\natexlab{c}})Li, Li, Xiong, and Hoi]{li2022blip}
Junnan Li, Dongxu Li, Caiming Xiong, and Steven Hoi.
\newblock Blip: Bootstrapping language-image pre-training for unified
  vision-language understanding and generation.
\newblock In \emph{International Conference on Machine Learning}, pages
  12888--12900. PMLR, 2022{\natexlab{c}}.

\bibitem[Li et~al.(2023)Li, Li, Savarese, and Hoi]{li2023blip}
Junnan Li, Dongxu Li, Silvio Savarese, and Steven Hoi.
\newblock Blip-2: Bootstrapping language-image pre-training with frozen image
  encoders and large language models.
\newblock In \emph{International conference on machine learning}, pages
  19730--19742. PMLR, 2023.

\bibitem[Li et~al.(2024{\natexlab{b}})Li, Chen, Chu, Sun, Guan, Han, and
  He]{li2024dynamic}
Jiawen Li, Yuxuan Chen, Hongbo Chu, Qiehe Sun, Tian Guan, Anjia Han, and
  Yonghong He.
\newblock Dynamic graph representation with knowledge-aware attention for
  histopathology whole slide image analysis.
\newblock In \emph{Proceedings of the IEEE/CVF Conference on Computer Vision
  and Pattern Recognition}, pages 11323--11332, 2024{\natexlab{b}}.

\bibitem[Liu et~al.(2021)Liu, Zhang, Hou, Mian, Wang, Zhang, and
  Tang]{liu2021self}
Xiao Liu, Fanjin Zhang, Zhenyu Hou, Li Mian, Zhaoyu Wang, Jing Zhang, and Jie
  Tang.
\newblock Self-supervised learning: Generative or contrastive.
\newblock \emph{IEEE transactions on knowledge and data engineering},
  35\penalty0 (1):\penalty0 857--876, 2021.

\bibitem[Lu et~al.(2023{\natexlab{a}})Lu, Chen, Williamson, Chen, Liang, Ding,
  Jaume, Odintsov, Zhang, Le, et~al.]{lu2023towards}
Ming~Y Lu, Bowen Chen, Drew~FK Williamson, Richard~J Chen, Ivy Liang, Tong
  Ding, Guillaume Jaume, Igor Odintsov, Andrew Zhang, Long~Phi Le, et~al.
\newblock Towards a visual-language foundation model for computational
  pathology.
\newblock \emph{arXiv preprint arXiv:2307.12914}, 2023{\natexlab{a}}.

\bibitem[Lu et~al.(2023{\natexlab{b}})Lu, Chen, Zhang, Williamson, Chen, Ding,
  Le, Chuang, and Mahmood]{lu2023visual}
Ming~Y Lu, Bowen Chen, Andrew Zhang, Drew~FK Williamson, Richard~J Chen, Tong
  Ding, Long~Phi Le, Yung-Sung Chuang, and Faisal Mahmood.
\newblock Visual language pretrained multiple instance zero-shot transfer for
  histopathology images.
\newblock In \emph{Proceedings of the IEEE/CVF Conference on Computer Vision
  and Pattern Recognition}, pages 19764--19775, 2023{\natexlab{b}}.

\bibitem[Lu et~al.(2024)Lu, Chen, Williamson, Chen, Zhao, Chow, Ikemura, Kim,
  Pouli, Patel, et~al.]{lu2024multimodal}
Ming~Y Lu, Bowen Chen, Drew~FK Williamson, Richard~J Chen, Melissa Zhao,
  Aaron~K Chow, Kenji Ikemura, Ahrong Kim, Dimitra Pouli, Ankush Patel, et~al.
\newblock A multimodal generative ai copilot for human pathology.
\newblock \emph{Nature}, pages 1--3, 2024.

\bibitem[Misialek and Allen(2016)]{misialek2016you}
Michael~J Misialek and Timothy~Craig Allen.
\newblock You're on social media! so now what?
\newblock \emph{Archives of Pathology \& Laboratory Medicine}, 140\penalty0
  (5):\penalty0 393--393, 2016.

\bibitem[Otsu(1979)]{otsu1979threshold}
Nobuyuki Otsu.
\newblock A threshold selection method from gray-level histograms.
\newblock \emph{IEEE transactions on systems, man, and cybernetics}, 9\penalty0
  (1):\penalty0 62--66, 1979.

\bibitem[Pantanowitz et~al.(2011)Pantanowitz, Valenstein, Evans, Kaplan,
  Pfeifer, Wilbur, Collins, and Colgan]{pantanowitz2011review}
Liron Pantanowitz, Paul~N Valenstein, Andrew~J Evans, Keith~J Kaplan, John~D
  Pfeifer, David~C Wilbur, Laura~C Collins, and Terence~J Colgan.
\newblock Review of the current state of whole slide imaging in pathology.
\newblock \emph{Journal of pathology informatics}, 2\penalty0 (1):\penalty0 36,
  2011.

\bibitem[Pataki et~al.(2022)Pataki, Olar, Ribli, Pesti, Kontsek,
  Gy{\"o}ngy{\"o}si, Bilecz, Kov{\'a}cs, Kov{\'a}cs, Kramer,
  et~al.]{pataki2022huncrc}
B{\'a}lint~{\'A}rmin Pataki, Alex Olar, Dezs{\H{o}} Ribli, Adri{\'a}n Pesti,
  Endre Kontsek, Benedek Gy{\"o}ngy{\"o}si, {\'A}gnes Bilecz, Tekla Kov{\'a}cs,
  Krist{\'o}f~Attila Kov{\'a}cs, Zs{\'o}fia Kramer, et~al.
\newblock Huncrc: annotated pathological slides to enhance deep learning
  applications in colorectal cancer screening.
\newblock \emph{Scientific Data}, 9\penalty0 (1):\penalty0 370, 2022.

\bibitem[Radford et~al.(2019)Radford, Wu, Child, Luan, Amodei, Sutskever,
  et~al.]{radford2019language}
Alec Radford, Jeffrey Wu, Rewon Child, David Luan, Dario Amodei, Ilya
  Sutskever, et~al.
\newblock Language models are unsupervised multitask learners.
\newblock \emph{OpenAI blog}, 1\penalty0 (8):\penalty0 9, 2019.

\bibitem[Radford et~al.(2021)Radford, Kim, Hallacy, Ramesh, Goh, Agarwal,
  Sastry, Askell, Mishkin, Clark, et~al.]{radford2021learning}
Alec Radford, Jong~Wook Kim, Chris Hallacy, Aditya Ramesh, Gabriel Goh,
  Sandhini Agarwal, Girish Sastry, Amanda Askell, Pamela Mishkin, Jack Clark,
  et~al.
\newblock Learning transferable visual models from natural language
  supervision.
\newblock In \emph{International conference on machine learning}, pages
  8748--8763. PMLR, 2021.

\bibitem[Roetzer-Pejrimovsky et~al.(2022)Roetzer-Pejrimovsky, Moser, Atli,
  Vogel, Mercea, Prihoda, Gelpi, Haberler, H{\"o}ftberger, Hainfellner,
  et~al.]{roetzer2022digital}
Thomas Roetzer-Pejrimovsky, Anna-Christina Moser, Baran Atli, Clemens~Christian
  Vogel, Petra~A Mercea, Romana Prihoda, Ellen Gelpi, Christine Haberler,
  Romana H{\"o}ftberger, Johannes~A Hainfellner, et~al.
\newblock The digital brain tumour atlas, an open histopathology resource.
\newblock \emph{Scientific Data}, 9\penalty0 (1):\penalty0 55, 2022.

\bibitem[Seyfioglu et~al.(2024)Seyfioglu, Ikezogwo, Ghezloo, Krishna, and
  Shapiro]{seyfioglu2024quilt}
Mehmet~Saygin Seyfioglu, Wisdom~O Ikezogwo, Fatemeh Ghezloo, Ranjay Krishna,
  and Linda Shapiro.
\newblock Quilt-llava: Visual instruction tuning by extracting localized
  narratives from open-source histopathology videos.
\newblock In \emph{Proceedings of the IEEE/CVF Conference on Computer Vision
  and Pattern Recognition}, pages 13183--13192, 2024.

\bibitem[Shastry and Sanjay(2022)]{shastry2022cancer}
K~Aditya Shastry and HA Sanjay.
\newblock Cancer diagnosis using artificial intelligence: A review.
\newblock \emph{Artificial Intelligence Review}, pages 1--33, 2022.

\bibitem[Shephard et~al.(2022)Shephard, Jahanifar, Wang, Dawood, Graham,
  Sidlauskas, Khurram, Rajpoot, and Raza]{shephard2022tiager}
Adam Shephard, Mostafa Jahanifar, Ruoyu Wang, Muhammad Dawood, Simon Graham,
  Kastytis Sidlauskas, Syed~Ali Khurram, Nasir Rajpoot, and Shan E~Ahmed Raza.
\newblock Tiager: Tumor-infiltrating lymphocyte scoring in breast cancer for
  the tiger challenge.
\newblock \emph{arXiv preprint arXiv:2206.11943}, 2022.

\bibitem[Shi et~al.(2024)Shi, Li, Gong, Zheng, and Fu]{shi2024vila}
Jiangbo Shi, Chen Li, Tieliang Gong, Yefeng Zheng, and Huazhu Fu.
\newblock Vila-mil: Dual-scale vision-language multiple instance learning for
  whole slide image classification.
\newblock In \emph{Proceedings of the IEEE/CVF Conference on Computer Vision
  and Pattern Recognition}, pages 11248--11258, 2024.

\bibitem[Silva-Rodriguez et~al.(2021)Silva-Rodriguez, Colomer, Dolz, and
  Naranjo]{silva2021self}
Julio Silva-Rodriguez, Adri{\'a}n Colomer, Jose Dolz, and Valery Naranjo.
\newblock Self-learning for weakly supervised gleason grading of local
  patterns.
\newblock \emph{IEEE journal of biomedical and health informatics}, 25\penalty0
  (8):\penalty0 3094--3104, 2021.

\bibitem[Song et~al.(2023)Song, Jaume, Williamson, Lu, Vaidya, Miller, and
  Mahmood]{song2023artificial}
Andrew~H Song, Guillaume Jaume, Drew~FK Williamson, Ming~Y Lu, Anurag Vaidya,
  Tiffany~R Miller, and Faisal Mahmood.
\newblock Artificial intelligence for digital and computational pathology.
\newblock \emph{Nature Reviews Bioengineering}, 1\penalty0 (12):\penalty0
  930--949, 2023.

\bibitem[Song et~al.(2024)Song, Chen, Ding, Williamson, Jaume, and
  Mahmood]{song2024morphological}
Andrew~H Song, Richard~J Chen, Tong Ding, Drew~FK Williamson, Guillaume Jaume,
  and Faisal Mahmood.
\newblock Morphological prototyping for unsupervised slide representation
  learning in computational pathology.
\newblock In \emph{Proceedings of the IEEE/CVF Conference on Computer Vision
  and Pattern Recognition}, pages 11566--11578, 2024.

\bibitem[Srinidhi et~al.(2021)Srinidhi, Ciga, and Martel]{srinidhi2021deep}
Chetan~L Srinidhi, Ozan Ciga, and Anne~L Martel.
\newblock Deep neural network models for computational histopathology: A
  survey.
\newblock \emph{Medical image analysis}, 67:\penalty0 101813, 2021.

\bibitem[Tang et~al.(2024)Tang, Zhou, Huang, Zhu, Zhang, and
  Liu]{tang2024feature}
Wenhao Tang, Fengtao Zhou, Sheng Huang, Xiang Zhu, Yi Zhang, and Bo Liu.
\newblock Feature re-embedding: Towards foundation model-level performance in
  computational pathology.
\newblock In \emph{Proceedings of the IEEE/CVF Conference on Computer Vision
  and Pattern Recognition}, pages 11343--11352, 2024.

\bibitem[Thomas et~al.(2022)Thomas, Anna-Christina, Baran, Vogel, Mercea,
  Romana, Gelpi, Christine, Romana, Hainfellner, et~al.]{thomas2022digital}
Roetzer-Pejrimovsky Thomas, Moser Anna-Christina, Atli Baran, Clemens~Christian
  Vogel, Petra~A Mercea, Prihoda Romana, Ellen Gelpi, Haberler Christine,
  H{\"o}ftberger Romana, Johannes~A Hainfellner, et~al.
\newblock The digital brain tumour atlas, an open histopathology resource.
\newblock \emph{Scientific Data}, 9\penalty0 (1), 2022.

\bibitem[Tokunaga et~al.(2019)Tokunaga, Teramoto, Yoshizawa, and
  Bise]{tokunaga2019adaptive}
Hiroki Tokunaga, Yuki Teramoto, Akihiko Yoshizawa, and Ryoma Bise.
\newblock Adaptive weighting multi-field-of-view cnn for semantic segmentation
  in pathology.
\newblock In \emph{Proceedings of the IEEE/CVF Conference on Computer Vision
  and Pattern Recognition}, pages 12597--12606, 2019.

\bibitem[Van~Rijthoven et~al.(2021)Van~Rijthoven, Balkenhol, Sili{\c{n}}a, Van
  Der~Laak, and Ciompi]{van2021hooknet}
Mart Van~Rijthoven, Maschenka Balkenhol, Karina Sili{\c{n}}a, Jeroen Van
  Der~Laak, and Francesco Ciompi.
\newblock Hooknet: Multi-resolution convolutional neural networks for semantic
  segmentation in histopathology whole-slide images.
\newblock \emph{Medical image analysis}, 68:\penalty0 101890, 2021.

\bibitem[Veeling et~al.(2018)Veeling, Linmans, Winkens, Cohen, and
  Welling]{veeling2018rotation}
Bastiaan~S Veeling, Jasper Linmans, Jim Winkens, Taco Cohen, and Max Welling.
\newblock Rotation equivariant cnns for digital pathology.
\newblock In \emph{Medical Image Computing and Computer Assisted
  Intervention--MICCAI 2018: 21st International Conference, Granada, Spain,
  September 16-20, 2018, Proceedings, Part II 11}, pages 210--218. Springer,
  2018.

\bibitem[Verghese et~al.(2023)Verghese, Lennerz, Ruta, Ng, Thavaraj,
  Siziopikou, Naidoo, Rane, Salgado, Pinder, et~al.]{verghese2023computational}
Gregory Verghese, Jochen~K Lennerz, Danny Ruta, Wen Ng, Selvam Thavaraj,
  Kalliopi~P Siziopikou, Threnesan Naidoo, Swapnil Rane, Roberto Salgado,
  Sarah~E Pinder, et~al.
\newblock Computational pathology in cancer diagnosis, prognosis, and
  prediction--present day and prospects.
\newblock \emph{The Journal of Pathology}, 260\penalty0 (5):\penalty0 551--563,
  2023.

\bibitem[Vorontsov et~al.(2024)Vorontsov, Bozkurt, Casson, Shaikovski,
  Zelechowski, Severson, Zimmermann, Hall, Tenenholtz, Fusi,
  et~al.]{vorontsov2024foundation}
Eugene Vorontsov, Alican Bozkurt, Adam Casson, George Shaikovski, Michal
  Zelechowski, Kristen Severson, Eric Zimmermann, James Hall, Neil Tenenholtz,
  Nicolo Fusi, et~al.
\newblock A foundation model for clinical-grade computational pathology and
  rare cancers detection.
\newblock \emph{Nature Medicine}, pages 1--12, 2024.

\bibitem[Vu et~al.(2023)Vu, Rajpoot, Raza, and Rajpoot]{vu2023handcrafted}
Quoc~Dang Vu, Kashif Rajpoot, Shan E~Ahmed Raza, and Nasir Rajpoot.
\newblock Handcrafted histological transformer (h2t): Unsupervised
  representation of whole slide images.
\newblock \emph{Medical Image Analysis}, 85:\penalty0 102743, 2023.

\bibitem[Wang et~al.(2022)Wang, Yang, Zhang, Wang, Zhang, Yang, Huang, and
  Han]{wang2022transformer}
Xiyue Wang, Sen Yang, Jun Zhang, Minghui Wang, Jing Zhang, Wei Yang, Junzhou
  Huang, and Xiao Han.
\newblock Transformer-based unsupervised contrastive learning for
  histopathological image classification.
\newblock \emph{Medical image analysis}, 81:\penalty0 102559, 2022.

\bibitem[Wang et~al.(2024)Wang, Zhao, Marostica, Yuan, Jin, Zhang, Li, Tang,
  Wang, Li, et~al.]{wang2024pathology}
Xiyue Wang, Junhan Zhao, Eliana Marostica, Wei Yuan, Jietian Jin, Jiayu Zhang,
  Ruijiang Li, Hongping Tang, Kanran Wang, Yu Li, et~al.
\newblock A pathology foundation model for cancer diagnosis and prognosis
  prediction.
\newblock \emph{Nature}, pages 1--9, 2024.

\bibitem[Wei et~al.(2021)Wei, Suriawinata, Ren, Liu, Lisovsky, Vaickus, Brown,
  Baker, Tomita, Torresani, et~al.]{wei2021petri}
Jerry Wei, Arief Suriawinata, Bing Ren, Xiaoying Liu, Mikhail Lisovsky, Louis
  Vaickus, Charles Brown, Michael Baker, Naofumi Tomita, Lorenzo Torresani,
  et~al.
\newblock A petri dish for histopathology image analysis.
\newblock In \emph{Artificial Intelligence in Medicine: 19th International
  Conference on Artificial Intelligence in Medicine, AIME 2021, Virtual Event,
  June 15--18, 2021, Proceedings}, pages 11--24. Springer, 2021.

\bibitem[Weinstein et~al.(2013)Weinstein, Collisson, Mills, Shaw, Ozenberger,
  Ellrott, Shmulevich, Sander, and Stuart]{weinstein2013cancer}
John~N Weinstein, Eric~A Collisson, Gordon~B Mills, Kenna~R Shaw, Brad~A
  Ozenberger, Kyle Ellrott, Ilya Shmulevich, Chris Sander, and Joshua~M Stuart.
\newblock The cancer genome atlas pan-cancer analysis project.
\newblock \emph{Nature genetics}, 45\penalty0 (10):\penalty0 1113--1120, 2013.

\bibitem[Xu et~al.(2024)Xu, Usuyama, Bagga, Zhang, Rao, Naumann, Wong, Gero,
  Gonz{\'a}lez, Gu, et~al.]{xu2024whole}
Hanwen Xu, Naoto Usuyama, Jaspreet Bagga, Sheng Zhang, Rajesh Rao, Tristan
  Naumann, Cliff Wong, Zelalem Gero, Javier Gonz{\'a}lez, Yu Gu, et~al.
\newblock A whole-slide foundation model for digital pathology from real-world
  data.
\newblock \emph{Nature}, pages 1--8, 2024.

\bibitem[Ye et~al.(2023)Ye, Xu, Yan, Xu, Qian, Zhang, and Huang]{Ye_2023_ICCV}
Qinghao Ye, Guohai Xu, Ming Yan, Haiyang Xu, Qi Qian, Ji Zhang, and Fei Huang.
\newblock Hitea: Hierarchical temporal-aware video-language pre-training.
\newblock In \emph{Proceedings of the IEEE/CVF International Conference on
  Computer Vision (ICCV)}, pages 15405--15416, 2023.

\bibitem[Yin et~al.(2024)Yin, Liu, Zhou, Wong, and Yuen]{yin2024prompting}
Chong Yin, Siqi Liu, Kaiyang Zhou, Vincent Wai-Sun Wong, and Pong~C Yuen.
\newblock Prompting vision foundation models for pathology image analysis.
\newblock In \emph{Proceedings of the IEEE/CVF Conference on Computer Vision
  and Pattern Recognition}, pages 11292--11301, 2024.

\bibitem[Zhang et~al.(2024)Zhang, Xu, Usuyama, Xu, Bagga, Tinn, Preston, Rao,
  Wei, Valluri, Wong, Tupini, Wang, Mazzola, Shukla, Liden, Gao, Lungren,
  Naumann, Wang, and Poon]{zhang2024biomedclip}
Sheng Zhang, Yanbo Xu, Naoto Usuyama, Hanwen Xu, Jaspreet Bagga, Robert Tinn,
  Sam Preston, Rajesh Rao, Mu Wei, Naveen Valluri, Cliff Wong, Andrea Tupini,
  Yu Wang, Matt Mazzola, Swadheen Shukla, Lars Liden, Jianfeng Gao, Matthew~P.
  Lungren, Tristan Naumann, Sheng Wang, and Hoifung Poon.
\newblock Biomedclip: a multimodal biomedical foundation model pretrained from
  fifteen million scientific image-text pairs, 2024.

\bibitem[Zhang et~al.(2019)Zhang, Chen, McGough, Xing, Wang, Bui, Xie, Sapkota,
  Cui, Dhillon, et~al.]{zhang2019pathologist}
Zizhao Zhang, Pingjun Chen, Mason McGough, Fuyong Xing, Chunbao Wang, Marilyn
  Bui, Yuanpu Xie, Manish Sapkota, Lei Cui, Jasreman Dhillon, et~al.
\newblock Pathologist-level interpretable whole-slide cancer diagnosis with
  deep learning.
\newblock \emph{Nature Machine Intelligence}, 1\penalty0 (5):\penalty0
  236--245, 2019.

\bibitem[Zhu et~al.(1997)Zhu, Byrd, Lu, and Nocedal]{zhu1997algorithm}
Ciyou Zhu, Richard~H Byrd, Peihuang Lu, and Jorge Nocedal.
\newblock Algorithm 778: L-bfgs-b: Fortran subroutines for large-scale
  bound-constrained optimization.
\newblock \emph{ACM Transactions on mathematical software (TOMS)}, 23\penalty0
  (4):\penalty0 550--560, 1997.

\bibitem[Zhu et~al.(2021)Zhu, Ren, Richards, Suriawinata, Tomita, and
  Hassanpour]{zhu2021development}
Mengdan Zhu, Bing Ren, Ryland Richards, Matthew Suriawinata, Naofumi Tomita,
  and Saeed Hassanpour.
\newblock Development and evaluation of a deep neural network for histologic
  classification of renal cell carcinoma on biopsy and surgical resection
  slides.
\newblock \emph{Scientific reports}, 11\penalty0 (1):\penalty0 7080, 2021.

\end{thebibliography}
}



\twocolumn[{%
	\centering
	\large
	\textbf{Supplementary Material: Multi-Resolution Pathology-Language Pre-training Model with Text-Guided Visual Representation}
	\normalsize
\renewcommand\twocolumn[1][]{#1}%
\captionsetup{type=figure}
\begin{center}
    \centering
    \includegraphics[width=\textwidth]{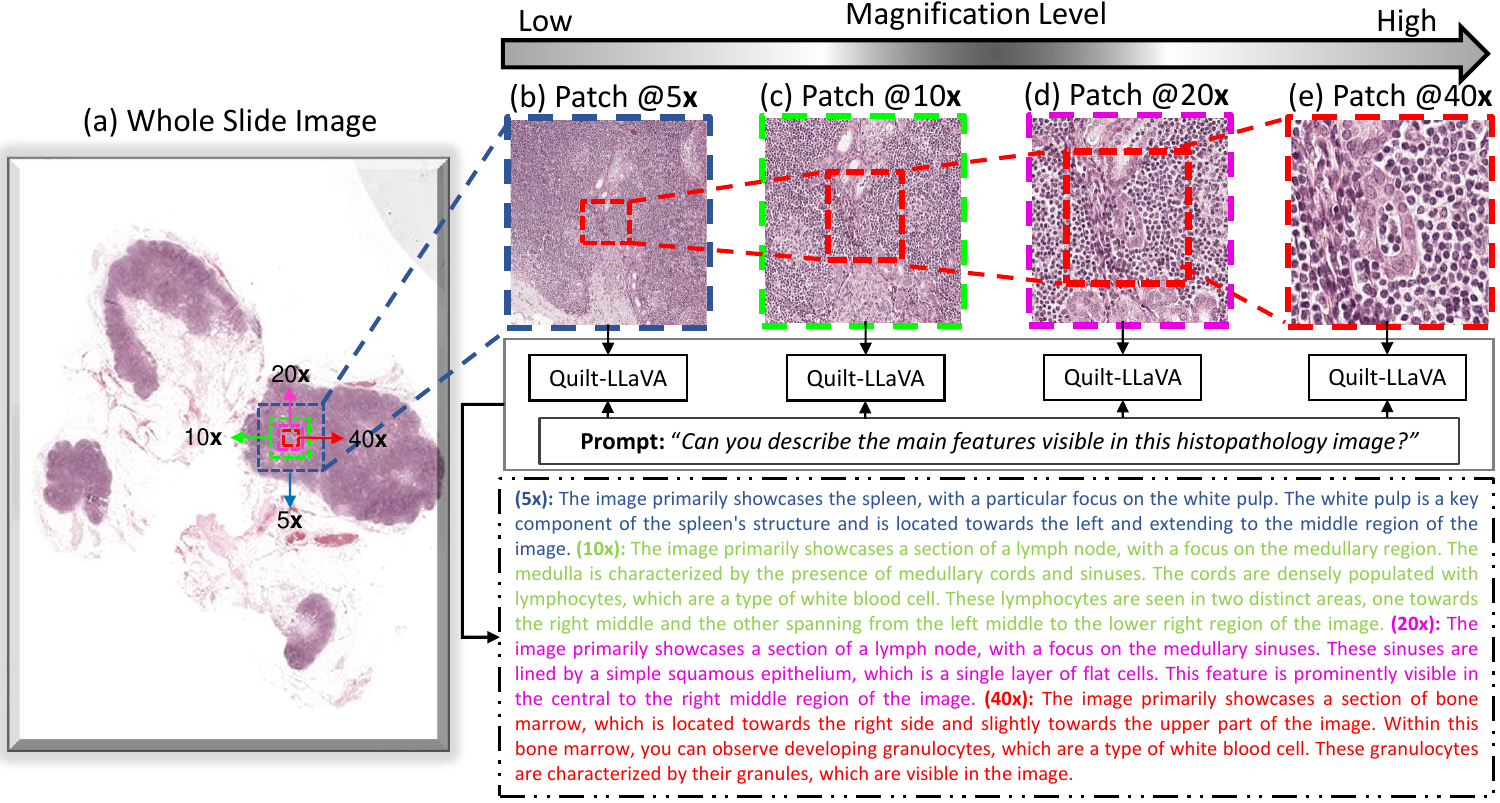} 
    \caption{Example of multi-resolution analysis of a histology image extracted from input WSI {\bf(a)} using the Quilt-LLaVA \cite{seyfioglu2024quilt}.
Exemplar histology patches {\bf(b)-(e)} are shown at different magnifications, demonstrating how higher magnification (5$\times$ to 40$\times$) shifts focus from contextual to detailed information. 
Textual descriptions generated by Quilt-LLaVA vary, reflecting the change in textual details observed at each magnification level from 5$\times$ to 40$\times$.}
\vspace{-1mm}
    \label{fig2_supp}    
\end{center}
}
]

\begin{figure*}[t!]
\centering
\includegraphics[width=\linewidth]{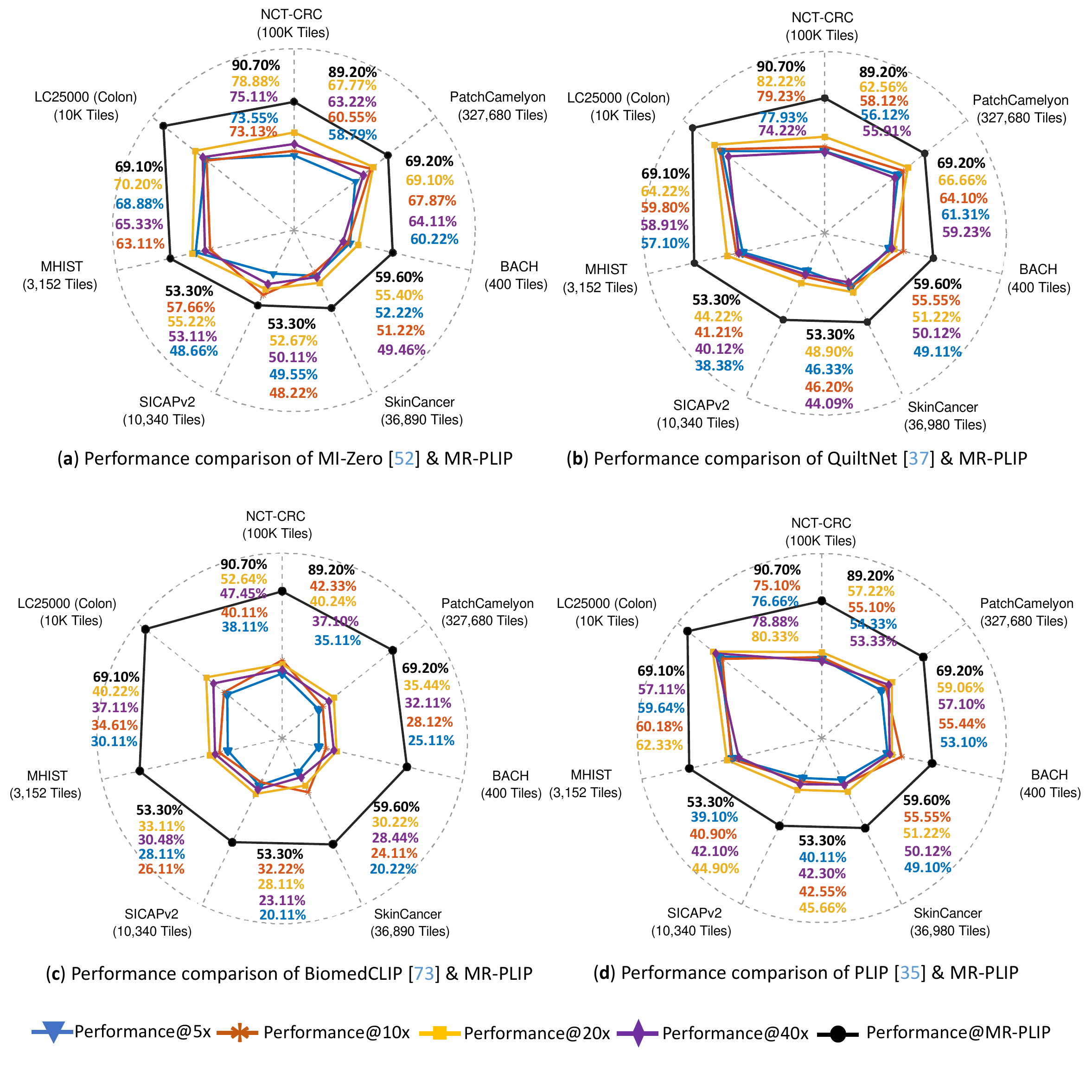}
\caption{Zero-shot tile-based classification performance in terms of accuracy of SOTA VL models including (a) MI-Zero \cite{lu2023visual}, (b) QuiltNet \cite{ikezogwo2024quilt}, (c) BiomedCLIP \cite{zhang2024biomedclip}, and (d) PLIP \cite{huang2023visual} on testing splits of seven independent datasets. 
Both models are pre-trained on the TCGA dataset using patches from 20k WSIs.
In each experiment, the pre-trained vision-language encoders are fine-tuned on a fixed magnification level 5$\times$, 10$\times$, 20$\times$, or 40$\times$. 
Performance variations across different magnification levels show the need for a multi-resolution VL model in computational pathology for improved generalization capabilities.}
\label{fig3}
\end{figure*}

\section*{Table of Contents}
\begin{enumerate}
    \item  \href{insights}{MR-PLIP Model: More Insights (Sec. \ref{insights})}
    \item  \href{parent}{Parent-Child Hierarchy (Sec. \ref{parent})}
    \item  \href{encoder}{Multi-modal Encoder for Text-guided Visual Representation (Sec. \ref{encoder})}
    \item  \href{obj}{Pre-Training Baseline Objectives (Sec. \ref{obj})}
    \item  \href{train}{Additional Training and Implementation Details (Sec. \ref{train})}
     \item \href{metric}{Evaluation Metrics (Sec. \ref{metric})}
    \item  \href{datasets}{Histopathology Datasets (Sec. \ref{datasets})}
    \item  \href{ablation}{More Ablation Studies (Sec. \ref{ablation})}
    \item  \href{zero}{Zero-Shot Experiments (Sec. \ref{zero}).}
    \item  \href{probe}{Linear Probe Experiments (Sec. \ref{probe}).}
    \item  \href{weak}{Weakly-Supervised WSI Classification Results (Sec. \ref{weak}).}
    \item  \href{fine}{Fine-tune Evaluation (Sec. \ref{fine}).}
    \item  \href{more}{More Comparison with Recent MIL-based Methods (Sec. \ref{more}).}  
\end{enumerate}

\section{MR-PLIP Model: More Insights}
\label{insights}
In clinical diagnostics, expert pathologists often analyze the WSI to predict the outcomes of diseases by inspecting it from multiple resolution levels.
The multi-resolution analysis of the WSIs assists expert pathologists to better analyze the tumor micro-environment by looking at the surrounding tissue/cellular structures \cite{ding2020multi, bejnordi2015multi, hashimoto2020multi, abels2019computational, zhang2019pathologist, chen2022fast}.
For example, pathologists look at the global architectural composition of the tissue sample and analyze the context of each tissue component, including cancer, to identify the presence of both healthy and cancerous tissues.
Additionally, they zoom in into each region of interest, where the tissue is examined at a high resolution, to obtain the details of the cancer cells, and characterize the tumor based on its local cellular composition. 
Another example where pathologists take advantage of both context and details is the spatial distribution of immune cells, which may be detected in the presence of inflammation inside the tumor or the stromal compartment of the cancer regions, as well as in specific clustered groups called tertiary lymphoid structures, which may develop in response to cancer as shown in Figs. \ref{fig2_supp}.

The above multi-scale analysis is crucial, as it involves the integration of both overarching (i.e., viewing the WSI at the lowest level of magnification) and detailed (i.e., viewing the WSI at the highest level of magnification) viewpoints \cite{cruz2017accurate,cornish2012whole, hanna2020whole, bolhasani2020histopathological}. 
Such a thorough approach enables pathologists to accurately distinguish between various types of cancer, such as differentiating invasive ductal carcinoma from invasive lobular carcinoma, as well as identifying tumor-infiltrating lymphocytes \cite{aresta2019bach,bolhasani2020histopathological}.

Most contemporary VLMs in histopathology primarily use histology images extracted from WSI of a single resolution, which might restrict their capability to adequately convey the essential broad and detailed perspectives for optimal analysis \cite{lu2023visual, lu2023towards, huang2023visual}. 
An illustration of this, as shown in Figs. \ref{fig2_supp}, is provided by the SOTA VLM, Quilt-LLaVA \cite{seyfioglu2024quilt}, which demonstrates how, with increasing magnification levels, the amount of textual descriptions derived from the input histology patch decreases. 
This decline is attributed to the loss of contextual information at higher magnifications. 
Additionally, pivotal cues, such as those indicating invasive lymph node, may only be visible at specific magnifications, highlighting the Quilt-LLaVA model's considerable dependency on certain magnification levels for generating accurate textual descriptions, a limitation that might be seen as a drawback.

To explore the multi-resolution generalization abilities of the SOTA methods, we fine-tuned SOTA CPath models, including PLIP \cite{huang2023visual}, BiomedCLIP \cite{zhang2024biomedclip}, MI-Zero \cite{lu2023visual}, and QuiltNet \cite{ikezogwo2024quilt}, across magnification levels of 5$\times$, 10$\times$, 20$\times$, and 40$\times$, using 20,000 WSIs (comprising 34 million patches) from the TCGA dataset \cite{weinstein2013cancer}.
These models are assessed through zero-shot settings across seven benchmark datasets for tile-based classification, as depicted in Fig. \ref{fig3}.
Excluding our top-scoring mode MR-PLIP, Fig. \ref{fig3} shows that  20$\times$ is virtually the best, coming first in 13 out of 14 trials. The 10$\times$  consistently ranks either first or second in 8 out of 14 trials, while the extreme magnifications of 5$\times$ and 40$\times$ typically land in the last or third position in 10 out of 14 trials. These statistics highlight that the magnifications  20$\times$ and 10$\times$, striking a balance between detail and context, achieve optimal performance. 
Our intuition is that integrating the 5$\times$ and 40$\times$  alongside the 20$\times$ and 10$\times$ in VL models will further leverage the complementary of the four magnification levels. We suggest, therefore, that synchronizing visual-textual concepts across multiple resolutions enhances their efficacy for diverse CPath applications and their overall generalization.

\begin{figure}[t!]
\centering
\includegraphics[width=\linewidth]{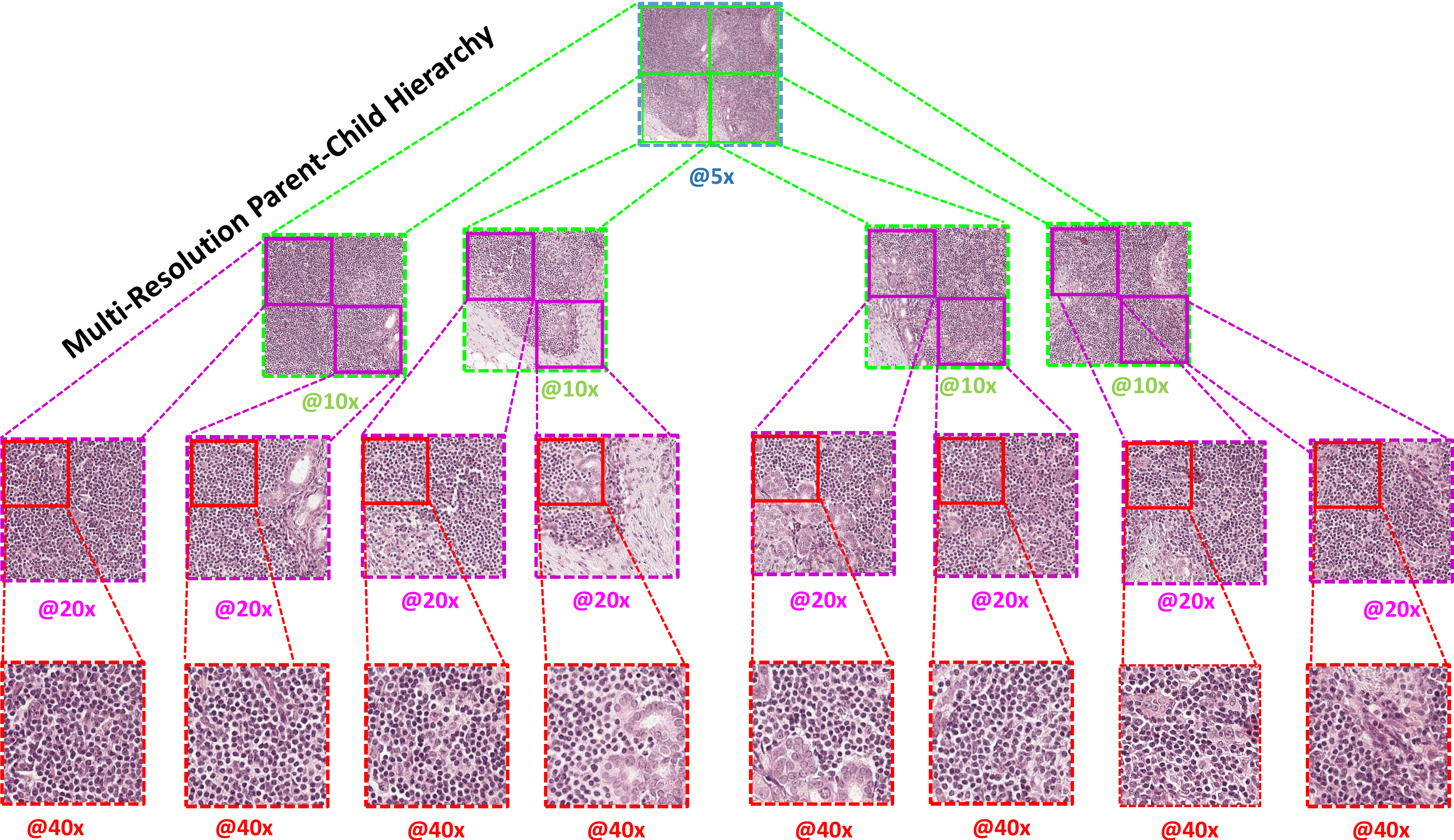}
\caption{Parent-child relationships in the visual bag ($B^{v}_{i,j}$). Our loss ($\mathcal{L}_{MRTVA}$) is minimized while preserving the parent-child relationships.}
\label{fig4}
\end{figure}

\section{Parent-Child Hierarchy}
\label{parent}
To clarify the approach used to preserve the hierarchical structure in our curated histopathology dataset, particularly when aligning text-guided visual features across different resolution levels as outlined in Eq. (\textcolor{red}{3}) of the main manuscript, we focus on maintaining the integrity of the parent-child relationship, as depicted in Fig \ref{fig4}. Specifically, the alignment of text-guided visual features is strictly between parents and their direct offspring. Within the visual bag $B^{v}_{i,j}$, patches lacking a parent-child linkage do not share visual content, rendering their alignment irrelevant. In our MR-PLIP model, alignment is confined to parents with their immediate children and vice versa, enabling the model to assimilate contextual and intricate details across different resolution levels. Alignments between grandparents or grandchildren are omitted to avoid confusion from minimally overlapping content.

\section{Multi-modal Encoder for Text-guided Visual Representation}
\label{encoder}
\begin{figure}[t!]
\centering
\includegraphics[width=\linewidth]{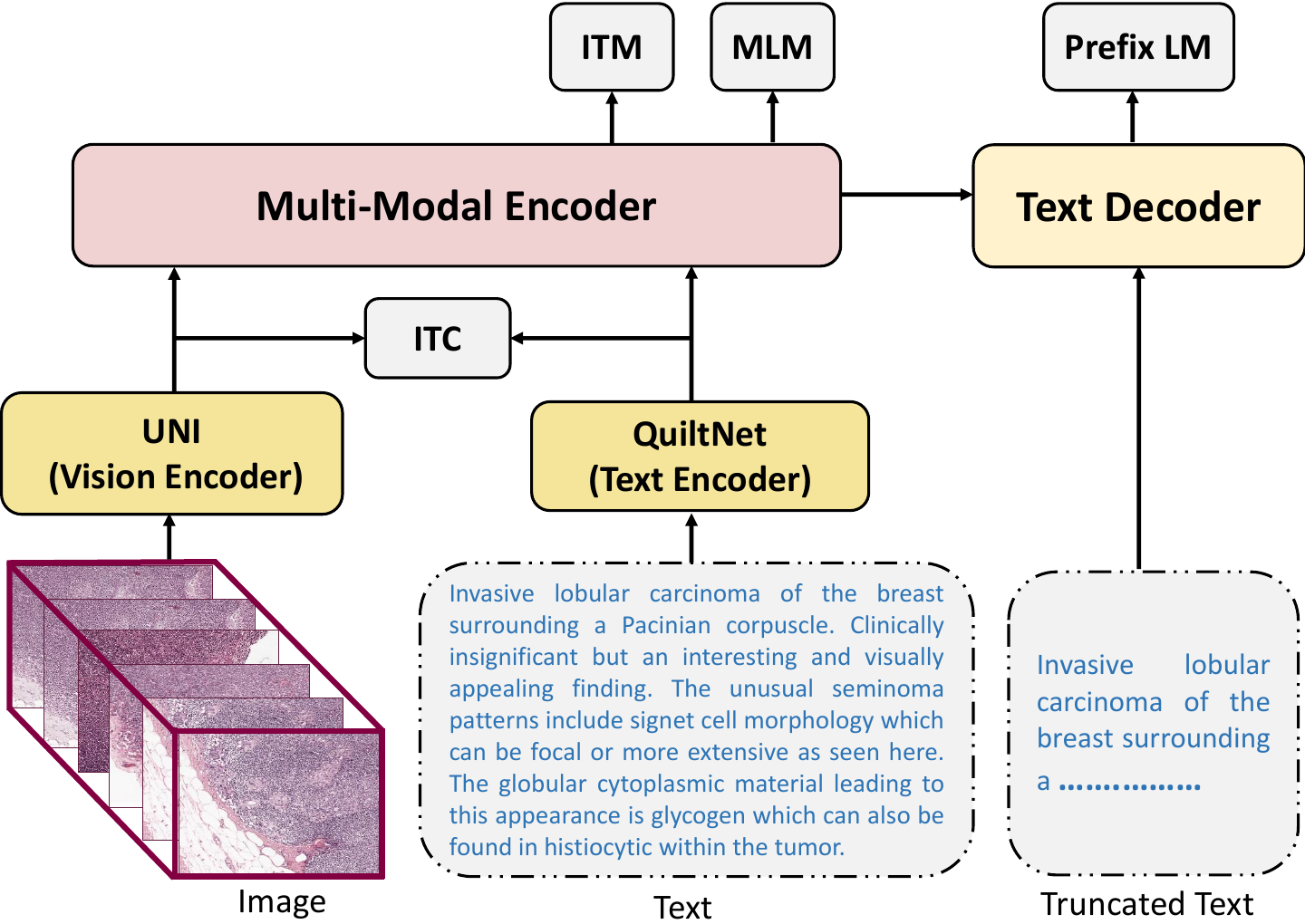}
\caption{The architecture of our multi-modal encoder for estimating text-guided visual feature representation (see Sec. \textcolor{red}{3.4} in the main manuscript).}
\label{fig5}
\end{figure}

Our multi-modal encoder’s architecture incorporates elements from the frameworks presented in~\cite{li2022mplug, ALBEF, Ye_2023_ICCV}, which originally were not applied to Computational Pathology (CPath). Here, we explicitly adapt and merge their methodologies for the first time to suit the specific needs of the CPath domain, as demonstrated in Fig.~\ref{fig5}, showing the encoder’s architecture.
This model incorporates the identical unimodal encoders for text (using the QuiltNet model) and images (using the UNI model), as outlined in the main manuscript, along with a multi-modal encoder for integrating text-guided visual features and a text decoder for generating text.
In particular, the vision encoder processes a multi-resolution histology patch, $p^{r}_{i,j}$, converting it into a visual representation, $v^{r}(i,j)$. Similarly, the text encoder converts the corresponding textual descriptions, $t^{r}_{i,j}$, of $p^{r}_{i,j}$ into textual features, $w^{r}_{i,j}$. To effectively merge multi-modal data while retaining the integrity of single-mode information, we initially combine the image and text features derived from the unimodal encoders as per references~\cite{li2022mplug, ALBEF}. To achieve this, the Image-Text Contrastive (ITC) loss is employed to synchronize the unimodal outputs from both the visual and text encoders. Subsequently, a cross-modal network with skip connections is utilized to merge the visual and textual data efficiently, applying the Image-Text Matching (ITM) and Masked Language Modeling (MLM) losses~\cite{ ALBEF, devlin2018bert} for effective fusion. The decoder, informed by the integrated image and prefix sub-sequence representation, is trained using the Prefix Language Modeling (PLM) loss to complete the caption generation~\cite{bi2020palm}.

%

\section{Pre-Training Baseline Objectives}
\label{obj}
During the pre-training process, we perform four pre-training tasks: Image-Text Contrastive Learning ($\mathcal{L}_{ITC}$), Image-Text Matching ($\mathcal{L}_{ITM}$), Masked Language Modeling ($\mathcal{L}_{MLM}$), and Prefix Language Modeling ($\mathcal{L}_{PLM}$). First, the ITC task is employed to align the unimodal representations of images and texts. Then, the ITM and MLM tasks are used for learning the multi-modal representation. Based on the image-language representations produced by the multi-modal encoder, the decoder is then trained with PLM loss to perform text-completion tasks.

\subsection{Image-Text Contrast (ITC):}
In line with~\cite{li2022align,ALBEF}, this task is used to align the unimodal encoders. Specifically, we calculate the softmax-normalized similarities between image-to-text and text-to-image, incorporating memory queues as described in MoCo~\cite{chenempirical} to expand the pool of negative samples during the learning process. For each patch $p^{r}_{i,j}$, we use its visual feature vector $v^{r}_{i,j}$ along with its corresponding $k_{o}$ positive words to generate the textual feature representation $w^{r}_{i,j}$. Through the application of two projector networks~\cite{ALBEF}, these visual and textual features are then transformed into $v^{r'}_{i,j}$ and ${w}^{r'}_{i,j}$ respectively.

%
%
Formally, the Image-Text Contrastive (ITC) loss is then calculated as:

\begin{equation}
\mathcal{L}_{i2t}=- \frac{1}{M}\sum_{m=1}^{M}\log \frac{\exp(s(v^{r'}_{i,j}, w^{r'}_{i,j})/\tau)}             {\sum_{m=1}^{M}\exp(s(v^{r'}_{i,j}, w^{r'}_{i,m})/\tau)},
\end{equation}

\begin{equation}
\mathcal{L}_{t2i}= -\frac{1}{M}\sum_{m=1}^{M}\log \frac{\exp(s(w^{r'}_{i,j}, v^{r'}_{i,j})/\tau)}                {\sum_{m=1}^{M}\exp(s(w^{r'}_{i,j}, v^{r'}_{i,m})/\tau)},
\end{equation}

\begin{equation}
\mathcal{L}_{ITC}=\frac{1}{2} (\mathcal{L}_{i2t} +\mathcal{L}_{t2i}),
\end{equation}

\noindent where $v^{r'}_{i,m}$ contain both the positive and the negative visual samples for text representation $w_{i,j}^{r}$ and $w^{r'}_{i,m}$ are the positive and negative text samples for visual feature $v_{i,j}^{r}$.

\subsection{Image-Text Matching (ITM):}
This task aims to predict whether an image and a text are paired or not based on the multi-modal representation \cite{li2022mplug, ALBEF}. 
In this loss, we maximize the log-likelihood of predicting a positive and negative pair, given the visual and textual descriptions ${v}^{r'}_{i,j}$ and ${w}^{r'}_{i,j}$.

\begin{equation}
\mathcal{L}_{itm}=-\mathbb{E}_{({v}^{r'}_{i,j}, w^{r'}_{i,j})} \log p(y| {v}^{r'}_{i,j}, w^{r'}_{i,j} )
\end{equation}

\noindent where $y \in \{+, - \}$ is the predicted label during the contrastive learning.

\subsection{Masked Language Modeling (MLM):}
In this task, tokens $w^{r'}_{i,j,b}$ are masked, and the model is tasked with predicting these masked tokens by leveraging the multi-modal representation~\cite{devlin2018bert}. The loss for this masked language modeling is defined as per references~\cite{devlin2018bert, li2022mplug, ALBEF}:

%
\begin{equation}
    \mathcal{L}_{mlm}=-\mathbb{E}_{({v}^{r'}{i,j}, w^{r'}_{i,j,b})}   \log p(w^{r'}_{i,j,b}| v^{r'}_{i,j}, w^{r'}_{i,j,/b})
\end{equation}
\noindent where $w^{r'}_{i,j,/b}$ are the non-masked key words.

\subsection{Prefix Language Modeling (PLM):}
This pretext task requires the model to complete the truncated texts based on the given image and prefix sequence of truncated texts \cite{li2022mplug, ALBEF, li2022blip}.
The model can be trained by maximizing the likelihood of the truncated text in an auto-regressive manner. 
Formally, the prefix language modeling loss is calculated as \cite{li2022blip, li2022mplug}:

\begin{equation}
\begin{split}
   \mathcal{L}_{plm}=-\mathbb{E}_{(v^{r'}_{i,j}w_{b})}\Bigg(\sum_{l=l_{p}}^{L}\log p\Big(w^{r'}_{i,j,l})| w^{r'}_{i,j,[l_{p},l]},  \\ 
w^{r'}_{i,j,< l_{p}}, v^{r'}_{(i,j)}\Big)\Bigg),
\end{split}
\end{equation}

\noindent where $L$ denotes the total number of words and $l_{p}$ is the length of a pre-fix sequence of keywords that is randomly selected.

\section{Additional Training and Implementation Details}
\label{train}
The TCGA dataset is renowned for being one of the largest publicly available histopathology collections, covering a broad spectrum of cancer morphologies and subtypes across key organs~\cite{weinstein2013cancer}. 
We included WSIs from 21 major primary cancer sites in TCGA, covering the adrenal gland, bile duct, bladder, bone marrow, breast, testis, pleura, cervix, eye, head and neck, stomach, uterus, thyroid, pancreas, esophagus, ovary, liver, endometrium, thymus, skin, and larynx. Testing was conducted on novel organs and cancer subtypes—including colorectal, kidney, prostate, and brain—as well as seen types such as breast, lung, bone, skin, and NSCLC, without any overlap with the training data. Results on both seen and unseen cases demonstrated superior performance compared to SOTA methods.

For the customization of the MR-PLIP algorithm to this dataset, we refined it through the creation of 34 million multi-resolution histology patches, each measuring 512 × 512 pixels. Initially, we applied the Otsu method~\cite{otsu1979threshold} for WSI thresholding, then proceeded to extract patches at designated magnification levels (as detailed in Sections \textcolor{red}{3.1} and \textcolor{red}{3.2} in the main manuscript). These patches were chosen based on a criterion ensuring at least 70\% tissue coverage to highlight pertinent histological details.

Our pre-training methodology integrated various architectures, including both domain-specific and general models. The fine-tuning process for our MR-PLIP model involved initializing it with diverse sets of weights for image and text encoders. An ablation study was conducted to evaluate the MR-PLIP model’s performance across different setups (refer to Table~\ref{table4}). 
We initialized the domain-specific vision encoder with UNI (ViT-L/16), pre-trained on histopathology data~\cite{chen2024towards}, and the domain-specific text encoder with the first six layers of the QuiltNet model \cite{ikezogwo2024quilt}, a GPT-2 adaptation with a context length of 77 (GPT-2/77)~\cite{ikezogwo2024quilt}.
The multi-modal encoder was similarly initialized using the latter six layers of the pre-trained QuiltNet model (GPT-2/77). 
The MR-PLIP model was pre-trained over 50 epochs, with batch sizes set to 32 across six NVIDIA A100 GPUs.
The AdamW optimizer~\cite{adam2014method} was used for optimization, featuring a weight decay of 0.02 and beta values of (0.9, 0.98).
The learning rate experienced an initial ramp-up to 5e-5 across the first 1000 iterations, subsequently following a cosine decay pattern. Through empirical testing during pre-training, we established the optimal number of positive keywords ($k_{o}$) at 9, and set the Masked Language Modeling (MLM) mask ratio to 15\%~\cite{ALBEF}.

During the fine-tuning phase of our MR-PLIP model, we used various combinations of image and text encoders to enhance its performance, detailed as follows:

\begin{enumerate}
\item
In alignment with SOTA methods~\cite{huang2023visual, zhang2024biomedclip,lu2023visual,ikezogwo2024quilt}, we fine-tuned the baseline CLIP model~\cite{radford2021learning} using a ViT-B/16-224~\cite{dosovitskiy2020image} as the image encoder and GPT-2/77~\cite{radford2019language} as the text encoder.
\item
Recognizing the baseline CLIP's training on out-of-domain paired data, we also fine-tuned the MR-PLIP model with a pathology domain-specific pre-trained PLIP \cite{huang2023visual} , using PLIP-ViT-B/32-224 as the image encoder and GPT-2/347 as the text encoder.
\item
Following the approach of MI-zero and BiomedCLIP~\cite{zhang2024biomedclip}, we fine-tuned the MR-PLIP model using BioClinicalBert/512~\cite{alsentzer2019publicly} and PubMedBERT/256~\cite{gu2021domain} as text encoders, alongside CTransPath/224~\cite{wang2022transformer} as the image encoder. BioClinicalBert and PubMedBERT, both non-pathology text encoders, are trained on biomedical and clinical corpora, such as PubMed abstracts and MIMIC~\cite{johnson2016mimic}. CTransPath is trained through self-supervised learning on 15.5 million unlabeled histology patches, with both encoders using ViT-B/16.
\item
MR-PLIP was also fine-tuned using BioClinicalBert/512 as the text encoder and PLIP-ViT-B/32-224 as the in-domain image encoder.
\item
Furthermore, we fine-tuned MR-PLIP using CTransPath/224 as the in-domain image encoder and PLIP-GPT/347 as the in-domain text encoder.
\item
Additionally, we fine-tuned MR-PLIP using ViT-B/16 as the in-domain image encoder from the QuiltNet and GPT-2/77 as the in-domain text encoder of the QuiltNet model \cite{ikezogwo2024quilt}.
\end{enumerate}

\noindent These experiments were conducted using inference time prompts similar to \cite{lu2023towards} for a fair comparison. 
Throughout this paper, our reported experiments predominantly used the UNI (ViT-L/16)~\cite{chen2024towards} as the image encoder and GPT-2/77 as the text encoder from the QuiltNet model~\cite{ikezogwo2024quilt}.
Please see Table~\ref{table4} for comparison.

\section{Evaluation Metrics}
\label{metric}
For evaluating performance across different CPath tasks~\cite{huang2023visual, lu2023towards}, we use a variety of metrics. These include the weighted average $F_{1}$ score, \textcolor{black}{balanced accuracy}, dice score, precision, recall, multi-class Panoptic Quality (mPQ), Recall$@$1 (R$@$1), Recall$@$50 (R$@$50), and Recall$@$200 (R$@$200). 
Consistent with prior VLMs studies in CPath, the weighted average $F_{1}$ score and \textcolor{black}{balanced accuracy} are applied to gauge performance in tile-level and WSI-level classification tasks. For segmentation tasks, we measure using dice score, precision, and recall. The mPQ metric is specifically used for nuclei instance segmentation tasks, while R$@$1, R$@$50, and R$@$200 metrics are dedicated to evaluating the efficacy of cross-modal retrieval tasks.

\section{Downstream Histopathology Datasets}
\label{datasets}
We performed five different computational pathology (CPath) tasks, including \textit{\textbf{tile-level}} classification, \textit{\textbf{WSI-level}} classification, \textit{\textbf{cross-modal retrieval}}, \textit{\textbf{WSI segmentation}}, and \textit{\textbf{nuclei segmentation}}.
To evaluate our MR-PLIP model on these tasks, we used 26 independent datasets.
For fair comparisons with SOTA methods \cite{huang2023visual, lu2023visual, ikezogwo2024quilt, kang2023benchmarking}, we employed the official testing splits of the datasets for zero-shot evaluation.
For linear probing and fine-tuning experiments, we employed the official training and testing splits of each dataset.
The details of each dataset are outlined below:

\subsection{Tile-level Classification Datasets}
We used 15 independent datasets, consisting of a wide range of tissue images from various resolution levels across different cancer types and tissues.
These datasets include Databiox~\cite{bolhasani2020histopathological}, which focuses on invasive ductal carcinoma from 124 patients, BACH~\cite{iciar2018grand} for breast cancer analysis from 500 WSIs, PatchCamelyon~\cite{veeling2018rotation} for identifying normal and metastatic tumor tissues from 400 WSIs, WILDS-CAM17 \cite{bandi2018detection,koh2021wilds} for classifying breast metastasis, UniToPatho \cite{barbano2021unitopatho} for colorectal polyp classification, and Osteo~\cite{arunachalam2019viable} for osteosarcoma from 40 WSIs. 
Additionally, SkinCancer dataset~\cite{kriegsmann2022deep} provides 36,890 skin tissue patches for identifying various skin conditions, MHIST\cite{wei2021petri} for colorectal polyps analysis, RenalCell~\cite{brummer2022integrative} for studying clear-cell renal cell carcinoma, and several others focusing on specific cancers like lung and colon cancer, as well as datasets like DigestPath~\cite{da2022digestpath} for colonoscopy analysis, SICAP~\cite{silva2021self} for prostate cancer Gleason pattern classification, and WSSS4LUAD~\cite{han2022wsss4luad} for lung adenocarcinoma. These datasets, with their specific focuses, resolutions, and classifications, provide a comprehensive resource for validating and fine-tuning the MR-PLIP model's capability in accurately classifying a wide array of histopathological images, demonstrating its adaptability and precision across different domains within pathology. 
\begin{itemize}
\item \textbf{Databiox (3 Classes):} is an invasive ductal carcinoma dataset collected from pathological biopsy samples of 124 patients. 
This dataset comprises 922 samples, corresponding to $2100 \times 1574$ and $1276 \times 956$ pixels. 
Each sample is captured at four different levels of magnification, including 4$\times$, 10$\times$, 20$\times$, and 40$\times$. 
The samples are annotated into Grade I (well-differentiated), Grade II (moderately differentiated), and Grade III (poorly differentiated). 
\item  \textbf{BACH (4 Classes):} is a breast cancer dataset containing 500 large tiles, each with $2048 \times 1536$ pixels captured at 40$\times$ magnification and sampled from 500 WSIs. 
The dataset is classified into four different tissue types, including normal, benign, in-situ carcinoma, and invasive carcinoma. 
The official training (320 tiles) and testing (80 tiles) splits are provided. 

\item \textbf{PatchCamelyon (2 Classes):} is a breast cancer dataset containing normal and metastatic tumor tissues.
This dataset is collected from 400 WSIs, containing 327,680 H\&E stained histology images with 96$\times$96 pixel tiles.
The samples are extracted from lymph node sections at 10$\times$ magnification level to provide an increased field of view.
The official training, validation, and testing splits contain 262,144, 32,768, and 32,768 histology images.

\item \textbf{WILDS-CAM17 (2 classes):} is a patch-based breast metastasis detection dataset based on CAM17 dataset, with folds created by WILDS for testing the models’ robustness under distribution shift. 
The dataset consists of 417,894 patches, each with $96 \times 96$ pixels extracted from WSIs of breast cancer metastases in lymph node sections.
The patch label refers to whether the patch contains a tumor or is normal.
For training and evaluation, we used the official train–validation–test folds provided by WILDS. 
The training set contains 302,436 patches from three hospitals, and the model is evaluated on 34,904 validation patches and 80,554 testing patches.
We resized images at $224 \times 224$ pixels and evaluated SOTA methods.

\item \textbf{UniToPatho (6 classes):} is a colorectal polyp classification dataset containing 9,536 each with $1,812 \times 1,812$ pixels at 0.44 $\mu m$/pixel annotated and extracted from 292 WSIs. 
The dataset contains six tissue types, including normal (950 patches), hyperplastic polyp (545 patches), tubular adenoma with high-grade dysplasia (454 patches), tubular adenoma with low-grade dysplasia (3,618 patches), tubulovillous adenoma with high-grade dysplasia (916 patches), and tubulovillous adenoma with low-grade dysplasia (2,186 patches). 
The official train set consists of 6,270 patches, while the testing set contains 2,399  patches.
We resized images at $224 \times 224$ pixels and evaluated SOTA methods.

\item \textbf{Osteo (3 Classes):} dataset focuses on osteosarcoma and contains 1,144 patches of size $1024 \times 1024$ pixels.
The dataset is collected from 40 heterogeneous WSIs at 10$\times$ magnification level.
The dataset contains three distinct classes, including tumor, non-tumor, and necrotic tumor. 
The official train and test splits are provided with a ratio of 80:20.

\item \textbf{SkinCancer (16 Classes):} comprises 36,890 skin tissue patches extracted from 386 patients, each with $395 \times 395$ pixels.
The patches are captured at 10$\times$ magnification from patients with basal cell carcinoma, squamous cell carcinoma, naevi, and melanoma. 
The tiles are categorized into 16 distinct categories, including chondral tissue, dermis, elastosis, epidermis, hair follicle, skeletal muscle, necrosis, nerves, sebaceous glands, subcutis, eccrine glands, vessels, BCC, SqCC, naevi, melanoma. 
The official train (88971), validation (72348), and testing (28039) splits are provided. 

\item \textbf{MHIST (2 Classes):} is a colorectal polyps dataset that contains 3,152 tissue patches of size $224 \times 224$ pixels extracted at 40$\times$ magnification level from 328 WSIs.
The tiles are annotated into two classes, including hyperplastic polyps and sessile serrated adenomas. 
For training, 2,175 tiles are utilized, while 977 tiles are reserved for testing. 

\item \textbf{RenalCell (5 Classes):} dataset contains histology patterns of clear-cell renal cell carcinoma. 
The dataset consists of 52,713 H\&E-stained images with $300 \times 300$ pixels captured at 40$\times$.
The dataset is annotated into five distinct classes, including red blood cells, renal cancer, normal tissue, torn adipose necrotic tissue, and muscle fibrous stroma blood vessels. 

\item \textbf{NCT-CRC (9 Classes):} is a colorectal cancer dataset comprising H\&E stained images encompassing nine distinct classes including Adipose, background, debris, lymphocytes, mucus, smooth muscle, normal colon mucosa, cancer-associated stroma, and colorectal adenocarcinoma epithelium. 
Tissue patches, corresponding to $224 \times 224$ pixels, are extracted at 20$\times$ magnification level.
The training dataset comprises 100K patches extracted from 86 patients, while the testing set consists of 7,180 images extracted from 50 patients. 

\item \textbf{LC25000Lung (3 Classes):} is a lung cancer dataset comprising 25K H\&E stained images corresponding to $768 \times 768$ pixels extracted from 750 patients.
The dataset encompasses three classes, including lung adenocarcinoma, benign lung, and lung squamous cell carcinoma. 
The official training and testing splits are provided with a ratio of 80:20.

\item \textbf{LC25000Colon (2 Classes):} is a colon cancer dataset containing H\&E stained images each of size $768 \times 768$ pixels extracted from 500 patients. 
This dataset contains two classes: benign colon tissue and colon adenocarcinomas. 
The official training and testing splits are provided with a ratio of 80:20.

\item \textbf{DigestPath (2 Classes):} is a colonoscopy dataset containing H\&E 660 tissue images.
Similar to PLIP, we performed tile-based zero-shot classification for Tumor Vs. Normal on the testing split containing 18814 images.

\item \textbf{SICAP (4 Classes):} is a prostate cancer dataset tailored for Gleason pattern classification.
It encompasses $512 \times 512$ pixels tiles extracted from 155 WSIs. 
The official training split comprises 9,959 images sourced from 124 WSIs, while the testing split includes 2,122 images from 31 WSIs. 
The dataset encompasses four labels, indicating the primary Gleason pattern (3, 4, or 5) or noncancerous (NC). 

\item \textbf{WSSS4LUAD (2 Classes):} is a lung adenocarcinoma dataset containing tiles of $200 \times 500$ pixels.
It encompasses three distinct classes: tumor, tumor-associated stroma, and normal. 
We conducted a binary classification of tumor vs. normal. 
The training dataset comprises 7,063 images, while the testing set comprises 3,028 images (2,015 tumors, 1,013 normal). 

\end{itemize}

\subsection{Tumor Subtype Classification Datasets} 
We used eight distinct datasets for tumour subtyping at the WSI level: 
CAMELYON16 (CAM16)~\cite{bejnordi2017diagnostic}, 
CAMELYON-17 (CAM17)~\cite{bandi2018detection}, 
RCC-DHMC \cite{zhu2021development}, 
BRCA-BRACS \cite{brancati2022bracs}, 
HunCRC \cite{pataki2022huncrc}, 
PANDA \cite{bulten2022artificial}, 
EBRAINS \cite{roetzer2022digital, thomas2022digital}
and NSCLC-CPTAC \cite{edwards2015cptac}, all of which provide only slide-level labels. 
Following the approach of CONCH~\cite{lu2023towards} and MI-Zero~\cite{lu2023visual}, we employed a top-$K$ pooling operator for WSI classification task.
CAM16 focuses on breast cancer, specifically the detection of lymph node metastasis from 400 gigapixel WSIs, with a training split of 270 WSIs and a testing split of 130 WSIs, where 159 are normal, and 111 contain tumor regions. 
CAM17, another breast cancer dataset, originates from sentinel lymph node sections of 200 patients, comprising 1000 WSIs with 5 slides per patient, labeled for metastasis vs. normal, with the model evaluated on 500 WSIs for both training and testing. 
CPTAC dataset, consisting of 1091 WSIs of different patients, focuses on LUng ADenocarcinoma (LUAD) and LUng Squamous cell Carcinoma (LUSC)  and offers images at different resolutions, providing a comprehensive set of data for validating WSI-level classification performance.
EBRAINS is a dataset for 30-way fine-grained brain tumor subtyping.
RCC-DHMC is a Renal Cell Carcinoma (RCC) subtyping dataset.
HunCRC is a colorectal cancer screening dataset.
BRCA-BRACS is a breast cancer subtyping dataset while PANDA is a prostate cancer dataset for gleason scoring. 

\begin{itemize}
    \item \textbf{CAMELYON16 (CAM16) (2 Classes)} is a breast cancer dataset designed for detecting lymph node metastasis using gigapixel WSIs. 
It comprises a total of 400 WSIs with only slide-level labels provided. 
The official training split consists of 270 WSIs, while the testing split contains 130 WSIs. 
Within the training set, there are 159 normal WSIs, and 111 WSIs contain tumor regions indicating breast cancer metastasis. 

\item \textbf{CAMELYON17 (CAM17) (2 Classes)} is a breast cancer dataset generated from a sentinel lymph node section of the Breast from 200 patients. 
The Camelyon17 dataset contains 1000 WSIs with 5 slides per patient. 
The WSI is labeled as metastasis vs. normal. The proposed model is evaluated on the official train (500 WSIs) and test set (500 WSIs) splits.

\item \textbf{NSCLC-CPTAC (2 Classes)} is a Non-Small Cell Lung Carcinoma (NSCLC) subtyping based on CPTAC contains two classes including LUAD and LUSC.
We excluded slides that were frozen tissue, nontumor tissue, or were not labeled as having acceptable tumor segments, which resulted in 1,091 slides (578 LUAD and 513 LUSC). 
For training and evaluation, we divided datasets into train-validation-test folds with 80:10:10 ratio and 872:109:109 slides.
For zero-shot classification, we used 109 testing WSIs.

\item \textbf{EBRAINS \cite{roetzer2022digital, thomas2022digital} (30 classes)} is a dataset consists of H\&E histopathology WSIs of brain tissue selected from The Digital Brain Tumour Atlas an open histopathology resource. 
Similar to the CONCH \cite{lu2023visual}, we used a subset of 2,319 WSIs out of 3,114 WSIs and defined a 30-way fine-grained brain tumor subtyping task limited to diagnostic labels that have at least 30 slides. 
For the supervised dataset, we performed a 50–25–25 split for training (1,151 slides), validation (595 slides) and testing (573 slides).
For the zero-shot test set, we used the testing split of 573 slides. 
The WSI counts for each class in the dataset were also set according to the CONCH \cite{lu2023visual}.

\item \textbf{RCC-DHMC (5 classes)} is a Renal Cell Carcinoma (RCC) subtyping dataset consisting of 563 RCC H\&E diagnostic histopathology WSIs.
The dataset contains six cancer subtypes, including primary Clear Cell Renal Cell Carcinoma (CCRCC) (344 slides), Papillary Renal Cell Carcinoma (PRCC) (101 slides), CHromophobe RCC (CHRCC) (23 slides), Renal Oncocytomas (ROCY)(66 slides), and benign cases (29 slides).
Similar to UNI \cite{chen2024towards}, for training and evaluation, we used a modified configuration of the train–validation–test folds with a 70:4:26 ratio (393:23:147 slides), with eight CHRCC cases moved from the test to the training fold due to CHRCC being absent in the training fold.

\item \textbf{HunCRC (4 Classes)} is colorectal cancer screening dataset containing of 200 H\&E diagnostic histopathology WSIs of
colorectal biopsies. 
Similar to UNI \cite{chen2024towards}, we employed a  4-way coarse-grained subtyping task using the categories of normal (10 slides), non-neoplastic lesion (38 slides), CRC (46 slides), and adenoma (106 slides), in which the ground-truth label was set by the study’s pathologist. 
For training and evaluation, we split the dataset into 50:25:25 train–validation–test folds (158:21:21 slides).

\item \textbf{BRCA-BRACS (3 classes)} is a BReast CAncer (BRCA) dataset containing 547 breast carcinoma H\&E WSIs from 187 patients sourced from the breast carcinoma subtyping task.
The dataset contains 3 classes, including benign, atypical, and malignant tumor labels.
For training and evaluation, we used the official train–validation–test folds with a 72:12:16 ratio (395:65:87 WSIs).

\item \textbf{PANDA (6 classes)} is the International Society of Urological Pathology (ISUP) grading task derived from the PANDA challenge.
It consists of 10,616 prostate cancer core needle biopsies of the prostate. 
Each slide is assigned an ISUP score that defines prostate cancer grade (6-class grading task). 
We removed the noisy labels and considered 9,555 slides only in which Grade 0 (G0) contains 2,603 WSIs, G1 contains 2,399 WSIs, G2 contains 1,209WSIs, G3 contains 1,118 WSIs, G4 contains 1,124 WSIs, and G5 contains 1,102). 
For training and evaluation, we employed 80:10:10 train–validation–test folds (7,647:954:954 WSIs). 

\end{itemize}

\begin{table*}[t!]
\centering
\caption{Ablation study of MR-PLIP's zero-shot classification performance (weighted $F_{1}$ score) using different multi-resolution image-text pairs. 
The best resolution combinations for each dataset are highlighted.}
\begin{tabular}{ |c|c|c|c|c|c|c|}
\hline
Resolutions&CAM16&CPTAC&SICAP&DigestPath&Databiox&NCT-CRC\\
\hline
5$\times$,10$\times$&0.566&0.706&0.421&0.792&0.421&0.732 \\
20$\times$,40$\times$&\underline{0.631}&0.805&\underline{0.529}&0.887&0.472&\underline{0.835} \\
5$\times$,10$\times$, 20$\times$&0.599&0.771&0.469&0.883&\underline{0.518}&0.807\\
10$\times$,20$\times$,40$\times$&0.621&\underline{0.843}&0.502&\underline{0.891}&0.506&0.834 \\
5$\times$,10$\times$,20$\times$,40$\times$&\textbf{0.664}&\textbf{0.875}&\textbf{0.546}&\textbf{0.935}&\textbf{0.532}&\textbf{0.871} \\
\hline
\end{tabular}        
\label{table_newmag}
\end{table*}

\subsection{Histology Image Segmentation Datasets}  
We used three distinct datasets for histolog image segmentation tasks, including DigestPath, SICAP, and TIGER~\cite{shephard2022tiager}. 
\begin{itemize}
    \item \textbf{DigestPath:} focuses on colonoscopy H\&E tissue sections, comprising 660 images with pixel-level lesion annotations for colorectal cancer provided for 250 testing images from 93 patients. 
    \item \textbf{SICAP:} This dataset aimed at prostate cancer, facilitates Gleason pattern classification with 31 WSIs in the testing split for tumor vs. normal tissue segmentation, and 124 WSIs in the training split.
    \item \textbf{TIGER:} dataset, dedicated to tumor-infiltrating lymphocytes (TILs) score prediction in H\&E stained breast cancer images, includes 195 WSIs from as many patients. 
These WSIs are annotated for semantic segmentation to identify regions such as invasive tumor, tumor-associated stroma, in-situ tumor, and others. 
For our study, we simplified the classification into three classes: tumor, stroma (merging tumor-associated and inflamed stroma), and background, by combining invasive and in-situ tumors to denote the tumor region, with all other regions marked as background.
\end{itemize}

\subsection{Nuclei Segmentation Datasets} 
For the nuclei segmentation task, we follow the same settings as proposed in DINOSSLPath \cite{kang2023benchmarking} by employing a HoverNet \cite{graham2019hover} baseline model.
We used two publicly available datasets: PanNuKe~\cite{gamper2019pannuke} and CONSEP~\cite{graham2019hover}. 

\begin{itemize}
    \item \textbf{CONSEP:} This dataset focuses on diverse nuclei segmentation and classification across six nuclear types, containing 41 image tiles of $1000 \times 1000$ pixels each, captured at 40$\times$ magnification, with 26 images designated for training and 14 for testing. 
    \item \textbf{PanNuKe:} This dataset offers a broad diversity with 19 distinct tissue types for nuclei segmentation and classification. 
It includes 4,346 images for training and 1,888 images for testing, each measuring $256 \times 256$ pixels, showcasing a wide variety of tissues and nuclei types for comprehensive segmentation analysis.
\end{itemize}

\subsection{Cross-modal Retrieval Datasets} 
Mirroring the approaches of PLIP and QuiltNet, we assessed the effectiveness of cross-modal retrieval through zero-shot text-to-image and image-to-text retrieval tasks, using the Twitter validation~\cite{huang2023visual} and ARCH~\cite{gamper2021multiple} datasets.

\begin{itemize}
    \item \textbf{ARCH:} dataset, designed specifically for computational pathology, consists of 25,028 vision-language pairs from PubMed articles and pathology textbooks, narrowed down to 8,176 pairs after filtering, providing detailed diagnostic and morphological information across various stains, tissue types, and pathologies. 
    \item \textbf{Twitter:} dataset, derived from pathology-related hashtags~\cite{huang2023visual,misialek2016you,allen2014social}, offers a comprehensive collection of 243,375 public pathology images, including H\&E stained cases, along with image-text pairs from tweets and replies. For our analysis, we utilized a validation subset similar to PLIP, comprising 2,023 image-text pairs filtered from over 200,000 pairs, to evaluate our model's cross-modal retrieval performance on both common and rare pathology cases.
\end{itemize}

\begin{table*}[t!]
\centering
\caption{Ablation study examining the zero-shot classification performance of the MR-PLIP model in terms of weighted $F_{1}$ score, while varying the number of positive keywords ($k_{0}$).
The table highlights the effect of different $k_{0}$ values on the classification results across six datasets, with the optimal performance marked in bold for each dataset.}
\begin{tabular}{ |c|c|c|c|c|c|c|}
\hline
$k_{0}$ value&CAM16&CPTAC&SICAP&DigestPath&Databiox&NCT-CRC \\
\hline
$k_{0}=3$&0.602&0.816&0.514&0.918&0.476&0.813\\
$k_{0}=6$&0.636&0.858&0.529&0.926&0.484&0.847\\
$k_{0}=9$&\textbf{0.664}&\textbf{0.875}&\textbf{0.546}&0.935&\textbf{0.532}&\textbf{0.871}\\
$k_{0}=12$&0.662&0.859&0.532&\textbf{0.938}&0.498&0.863\\
$k_{0}=15$&0.642&0.824&0.528&0.929&0.481&0.859\\
$k_{0}=18$&0.611&0.821&0.505&0.901&0.470&0.850\\
\hline
\end{tabular}        
\label{table_positive}
\end{table*}

\begin{table}[t!]
\centering
\caption{Ablation study comparing the zero-shot classification performance of MR-PLIP in terms of weighted $F_{1}$ score using different captioning models to generate textual descriptions of histology images at the multi-resolution level. 
The table highlights the performance of Quilt-LLaVA, QuiltNet, BLIP2, and GPT4V across six datasets, with the best performance for each dataset marked in bold.}
\scalebox{0.90}{
\begin{tabular}{ |c|c|c|c|c|}
\hline
Models&Quilt-LLaVA&QuiltNet&BLIP2&GPT4V\\
\hline
CAM16&\textbf{0.664}&0.621&0.577&\underline{0.634} \\
CPTAC&\textbf{0.875}&\underline{0.834}&0.788&0.817 \\
SICAP&\textbf{0.546}&\underline{0.507}&.468&0.433 \\
DigestPath&\textbf{0.935}&\underline{0.881}&0.821&0.846 \\
Databiox&\textbf{0.532}&\underline{0.510}&0.401&0.476 \\
NCT-CRC&\textbf{0.871}&\underline{0.837}&0.703&0.752 \\
\hline
\end{tabular}   }     
\label{table_caption}
\end{table}

\begin{table*}[t!]
\caption{Zero-shot classification performance comparison in terms of weighted $F_{1}$ score using different pre-trained vision and text encoders.}
\begin{center}
\makebox[\linewidth]{
\scalebox{0.80}{
\begin{tabu}{|c|c|c|c|c|c|c|c|c|}
\tabucline[0.5pt]{-}
Ablation Study&Vision Encoder&Text Encoder&CAM16&CPTAC&SICAP&DigestPath&Databiox&NCT-CRC\\\tabucline[0.5pt]{-}
MR-PLIP&CLIP&CLIP&0.483&0.702&0.398&0.761&0.251&0.721\\
&(ViT-B/16-224)&(GPT-2/77)&&&&&&\\\tabucline[0.5pt]{-}
MR-PLIP&PLIP&PLIP&0.541&0.729&0.461&0.834&0.452&0.749\\
&(ViT-B/32-224)&(GPT-2/347)&&&&&&\\\tabucline[0.5pt]{-}
MR-PLIP&CTransPath&BioClinicalBert&0.584&0.752&0.531&0.883&0.471&0.751\\
&(ViT-B/16-224)&(BioClinicalBert/512)&&&&&&\\\tabucline[0.5pt]{-}
MR-PLIP&CTransPath&PubMedBERT&0.581&0.799&0.528&0.881&0.471&\underline{0.761}\\
&(ViT-B/16-224)&(PubMedBERT/256)&&&&&&\\\tabucline[0.5pt]{-}
MR-PLIP&PLIP&PubMedBERT&0.556&0.765&0.500&0.841&0.440&0.769\\
&(ViT-B/32-224)&(PubMedBERT/256)&&&&&&\\\tabucline[0.5pt]{-}
MR-PLIP&PLIP&BioClinicalBert&0.534&0.756&0.491&0.804&0.450&0.781\\
&(ViT-B/32-224)&(BioClinicalBert/512)&&&&&&\\\tabucline[0.5pt]{-}
MR-PLIP&CTransPath&PLIP&0.563&0.771&0.503&\underline{0.924}&0.479&\underline{0.806}\\
&(ViT-B/16-224)&(GPT/347)&&&&&&\\\tabucline[0.5pt]{-}
MR-PLIP&QuiltNet&QuiltNet&0.591&0.773&0.498&0.871&0.451&0.774\\
&(ViT-B/16-224)&(GPT-2/77)&&&&&&\\\tabucline[0.5pt]{-}
MR-PLIP&QuiltNet&PubMedBERT&0.611&0.785&0.519&0.909&0.485&0.821\\
&(ViT-B/16-224)&(PubMedBERT/256)&&&&&&\\\tabucline[0.5pt]{-}
MR-PLIP&DinoSSLPath&PubMedBERT&0.634&0.802&0.538&0.914&0.486&0.835\\
&(ViT-B/16-224)&(PubMedBERT/256)&&&&&&\\\tabucline[0.5pt]{-}
MR-PLIP&DinoSSLPath&QuiltNet&\underline{0.622}&\underline{0.841}&\underline{0.541}&\underline{0.931}&\underline{0.491}&\underline{0.842}\\
&(ViT-B/16-224)&(GPT-2/77)&&&&&&\\\tabucline[0.5pt]{-}
MR-PLIP&DinoSSLPath&BioClinicalBert&0.589&0.783&0.528&0.901&0.472&0.841\\
&(ViT-B/16-224)&(BioClinicalBert/512)&&&&&&\\\tabucline[0.5pt]{-}
\textbf{MR-PLIP}&\textbf{UNI}&\textbf{QuiltNet}&\textbf{0.664}&\textbf{0.875}&\textbf{0.546}&\textbf{0.935}&\textbf{0.532}&\textbf{0.871}\\
&\textbf{(ViT-L/16-224)}&\textbf{(GPT-2/77)}&&&&&&\\\tabucline[0.5pt]{-}
\end{tabu}
}}
\end{center}
\label{table4}
\end{table*}

\section{More Ablation Studies}
\label{ablation}
\subsection{Impact of Magnification Levels (Table \ref{table_newmag})}
In Sec. \ref{insights}, we presented the argument that multi-resolution pre-training is advantageous, considering that downstream CPath tasks are executed at various magnifications. Our experimental results indeed confirm that incorporating multiple resolutions in the pre-training dataset improves the performance of downstream tasks (see Table~\ref{table_newmag} for details).

To substantiate this finding, we carried out an ablation study by performing zero-shot classification tasks using combinations of different resolutions. We noted that the combination of 5$\times$,10$\times$,20$\times$, and 40$\times$ resolutions yielded the highest performance. 
On average, using 10$\times$, 20$\times$, and 40$\times$ resolutions was the second most effective, outperforming the use of just 20$\times$ and 40$\times$. 

The rationale is that images and their paired textual descriptions at lower magnifications, like 5$\times$, provide sufficient contextual information, whereas higher magnifications, like 40$\times$, offer the necessary cellular detail for high performance. Current SOTA methods fail to capture either the broader context or the intricate details, leading to a decrease in performance.

\begin{table*}[t!]
\centering
\caption{Zero-shot transfer for histology image classification performance comparison in terms of weighted average $F_{1}$ score. 
The results are reported using the zero-shot inference protocol used in the SOTA methods \cite{huang2023visual, ikezogwo2024quilt, lu2023visual} and our proposed zero-shot transfer protocol.}
\begin{tabular}{ |c|c|c|c|c|c|c|}
\hline
Zero-Shot&CAM16&CPTAC&SICAP&DigestPath&Databiox&NCT-CRC\\
\hline
Classical Zero-shot&\underline{0.636}&\underline{0.812}&\underline{0.526}&\underline{0.902}&\underline{0.472}&\underline{0.841}\\
Proposed Zero-shot&\textbf{0.664}&\textbf{0.875}&\textbf{0.546}&\textbf{0.935}&\textbf{0.532}&\textbf{0.871}\\
\hline
\end{tabular}
\label{table5}
\end{table*}

\subsection{Optimal Number of Positive Keywords ($k_{o}$) for the CVTA Module (Table \ref{table_positive})} We explore how different numbers of positive keywords ($k_{o}$), as detailed in Sec. (\textcolor{blue}{3.3}), impact performance. 
According to the results shown in Table~\ref{table_positive}, enhancing $k_{o}$ from 3 to 9 leads to performance gains across all datasets, attributed to the enrichment of information with more keywords. However, a further increment to $k_{o}=12$ only marginally diminishes performance across most datasets, with the exception of DigestPath, which still shows improvement. Beyond this point, performance suffers due to the addition of noisy keywords that dilute the textual description's relevance.

\subsection{Performance Comparison of Different Captioning Models (Table \ref{table_caption})}
Table \ref{table_caption} presents the zero-shot classification performance of MR-PLIP across six datasets using four different captioning methods: Quilt-LLaVA \cite{seyfioglu2024quilt}, QuiltNet \cite{ikezogwo2024quilt}, BLIP2 \cite{li2023blip}, and GPT4V \cite{achiam2023gpt}.
The best performance is observed with Quilt-LLaVA. 
Therefore, all results in this work are reported using Quilt-LLaVA as the captioning model.

\subsection{Generalization to other Image-Text Encoders (Table \ref{table4})} 
In this experiment, we compared the performance of our proposed MR-PLIP model in terms of initializing different image-text encoders including CLIP (out-of-domain pre-trained encoders), PLIP (in-domain pre-trained encoders), CTransPath (in-domain pre-trained image encoder), PubMedBERT (out-of-domain pre-trained text encoders), BioclinicalBert (out-of-domain pre-trained text encoders), DinoSSLPath (in-domain pre-trained image encoder), QuiltNet (in-domain pre-trained encoders) as shown in Table \ref{table1}.
The best results on six datasets are reported using UNI (ViT-L/16-224) as an image encoder and QuiltNet (GPT-2/77) to initialize the text encoder.
This is because UNI is pre-trained on unlabeled large histology images, and the QuiltNet text encoder is trained on 1M histology image-text pairs.
The in-domain MR-PLIP variants also showed comparable performance compared to the best-performing MR-PLIP (in-domain) variant.

\begin{figure}
    \centering
    \includegraphics[width=1\linewidth]{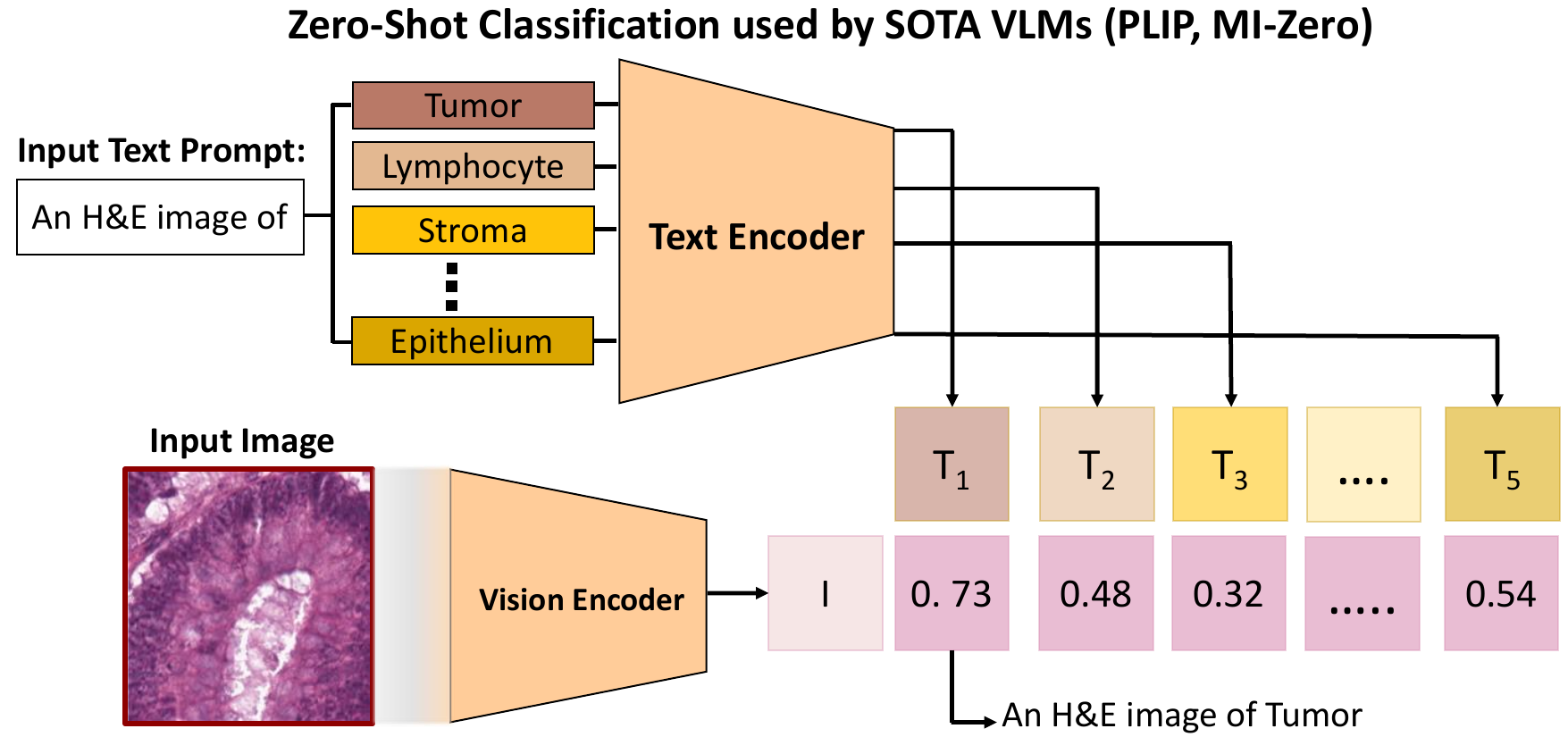}
    \caption{Illustration of the zero-shot classification process used by SOTA VLMs, like PLIP \cite{huang2023visual} and MI-Zero \cite{lu2023visual}. An input image, in this case, an H\&E-stained tissue image indicating a tumor, is processed in parallel by a vision encoder and a text encoder. 
    The vision encoder extracts features from the image, while the text encoder processes a set of predefined text prompts relating to possible classifications (e.g., Tumor, Lymphocyte, Stroma, Epithelium). Each class has a corresponding token ($T_1, T_2, T_3, ..., T_5$), and the model outputs a probability score for each class, predicting the likelihood that the input image corresponds to each class based on visual-textual feature matching. The highest probability score indicates the model's classification of the image.}
    \label{fig6}
\end{figure}

\begin{figure}
    \centering
    \includegraphics[width=1\linewidth]{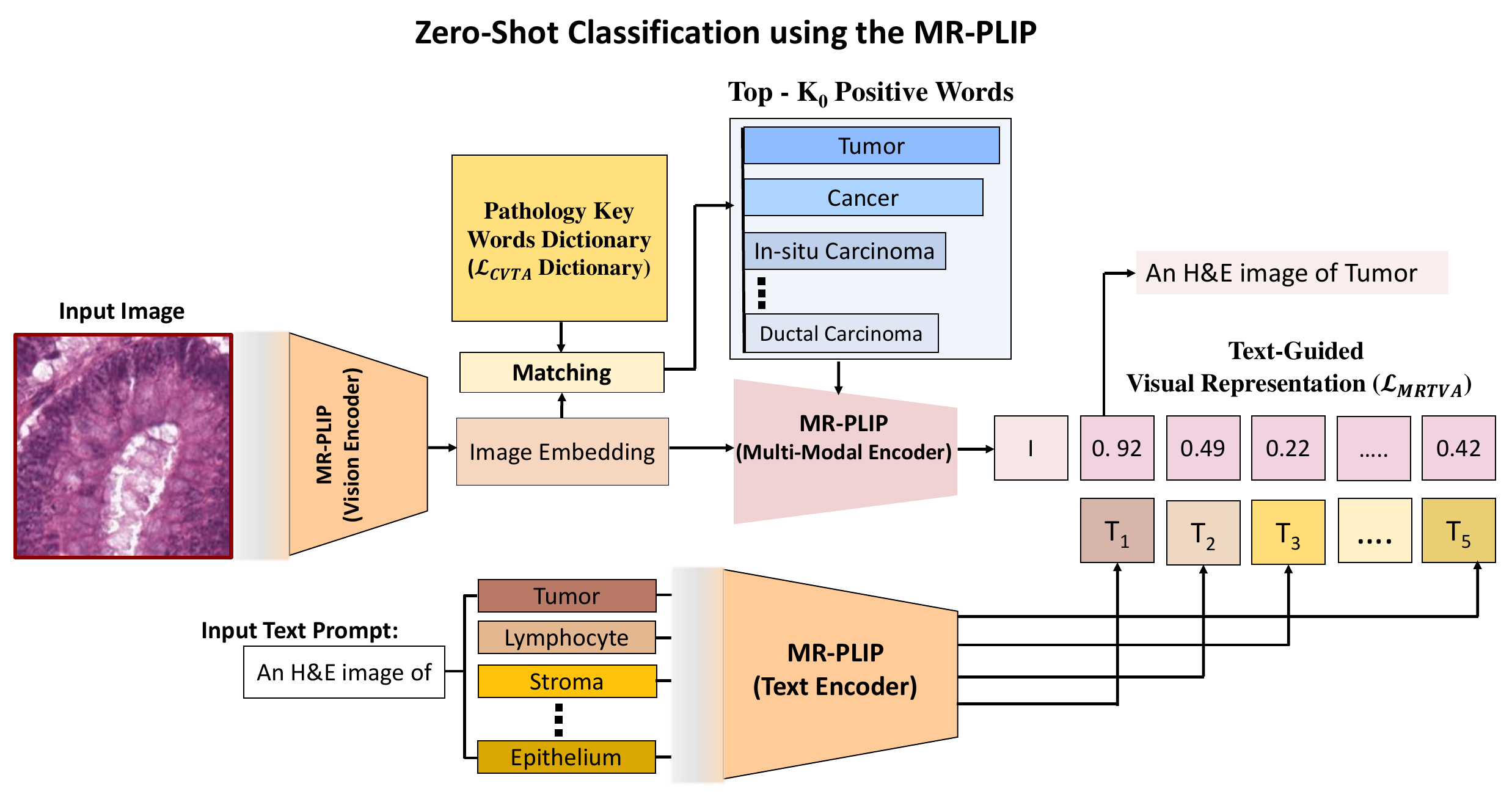}
    \caption{Illustration of the zero-shot classification process using the proposed MR-PLIP model.
    An input H\&E (Hematoxylin and Eosin) stained tissue image is input to the vision encoder and a set of testing prompts are input to the text encoder.
The vision encoder generates an image embedding, while the text encoder processes input text prompts such as ``An H\&E image of Tumor'' and generates embedding.
The image embeddings are matched against a pathology keyword dictionary, leading to a list of top $k_{0}$ positive words associated with various types of histology prompts.
 The multi-modal encoder (MR-PLIP) then combines these text and image embeddings to produce a text-guided visual representation. This yields a set of probability scores (such as 0.92, 0.49, etc.) for each possible classification token $T_1, T_2, T_3, ...T_5$, that represent different pathological features or diagnoses, predicting the most likely classification for the input image.
 The MR-PLIP model presents a significant enhancement through its integration of a pathology keywords dictionary, which augments the classification process. By comparing input text prompts with an extensive collection of relevant medical terminology provided by the dictionary, MR-PLIP can produce more precise and contextually detailed visual representations guided by text. As a result, it can determine more accurate classification probabilities for various pathology-specific tokens drawn from both the image and associated textual data. This approach advances beyond prior models by embedding specialized knowledge specific to the field of medicine into the algorithm's framework.}
    \label{fig7}
\end{figure}

\subsection{Impact of Zero-Shot Inference (Figs. \ref{fig6}-\ref{fig7} $\&$ Table \ref{table5})} 
In this experiment, we evaluate the performance of the MR-PLIP algorithm by employing two different zero-shot inference protocols.
Echoing the SOTA approaches such as PLIP \cite{huang2023visual}, QuiltNet \cite{ikezogwo2024quilt}, and MI-Zero \cite{lu2023visual}, we extract visual features for a given histology patch using our vision encoder and compare them against a predefined set of testing prompts to determine its class label as shown in Fig. \ref{fig6}.

Since our VLM is based on fine-tuning two uni-modal encoders (one vision encoder and one text encoder) and one multi-modal encoder, we introduce a novel zero-shot approach (Fig. \ref{fig7}) where, for a specific test histology patch, we select $k_{o}$ positive keywords from a dictionary of textual descriptions collected during the training process of the CVTA module (Sec. \textcolor{red}{3.3} in the main manuscript).
These positive keywords and the visual features are input to the multi-modal encoder to obtain text-guided visual features which are then used to match with the testing prompts to predict the class label as shown in Fig. \ref{fig7}.

Table \ref{table5} shows the performance comparison of using both zero-shot inference protocols on six independent datasets in terms of weighted $F_{1}$ score.
Following \cite{ikezogwo2024quilt} and \cite{lu2023visual}, we used similar testing prompts to compare the performances of both zero-shot evaluation protocols. 
Our proposed zero-shot inference protocol outperformed the classical approach across six datasets.
This demonstrates that the learned text-guided visual representations using the proposed multi-modal encoder are more effective compared to the unimodal encoder representations.

\subsection{Motivation of using Positive Keywords vs. Full Text (Table \ref{table_keyword})} 
Synthetically generated textual descriptions may contain hallucinations, noise, and irrelevant words, which the proposed CVTA module (Sec. \textcolor{cyan}{3.2}) removes.
The top-$k_{o}$ well-aligned words with visual features are retained as positive keywords. 
\textit{By using positive keywords, we capture fine-grained tissue morphology in multi-resolution histology images, enhancing zero-shot classification}. 
In contrast, full-text descriptions may obscure such details. 
Moreover, aligning multi-resolution positive keyword representations with histology images enables MR-PLIP to localize meaningful tissue structures more effectively. 
Our positive keyword-based alignment approach improves generalization to novel keyword-tissue structures.
\textit{As shown in our ablation study (Table \ref{table_keyword}), positive keyword alignment outperforms full-text and all-word alignments across six datasets}.

\begin{table}[t!]
\centering
\caption{Zero-shot weighted $F_{1}$ scores for MR-PLIP using full text, all words, and positive keywords alignment.}
\makebox[\linewidth]{
\scalebox{0.65}{
\begin{tabular}{ |c|c|c|c|c|c|c|}
\hline
Alignment&NCT-CRC&SICAP&Databiox&CAM16&CPTAC&EBRAINS\\
\hline
Full Text&0.834&0.446&0.481&0.611&0.833&0.351\\
All Words&\underline{0.846}&\underline{0.458}&\underline{0.501}&\underline{0.629}&\underline{0.846}&\underline{0.376}\\
Positive Keywords&\textbf{0.871}&\textbf{0.546}&\textbf{0.532}&\textbf{0.664}&\textbf{0.875}&\textbf{0.398}\\
\hline
\end{tabular} 
}}
\label{table_keyword}
\end{table}

\begin{table*}[t!]
    \centering
    \caption{
\textcolor{black}{Zero-shot segmentation performance comparison of gigapixel images in terms of dice score, precision, and recall with existing
VLMs in CPath on three independent datasets. 
The MR-PLIP algorithm outperforms existing models.}}
    \begin{tabular}{ |c|c|c|c|c|c|c|c|c|c|}
    \hline
   \textbf{Methods}&\multicolumn{3}{c|}{DigestPath\cite{da2022digestpath}}&\multicolumn{3}{c|}{SICAP \cite{silva2021self}}&\multicolumn{3}{c|}{TIGER \cite{shephard2022tiager}} \\
    \hline
CLIP \cite{radford2021learning}&0.367&0.492&0.511&0.367&0.599&0.605&0.210&0.261&0.278 \\
BioCLIP  \cite{zhang2024biomedclip}&0.446&0.581&0.601&0.484&0.536&0.557&0.255&0.281&0.302 \\
PLIP \cite{huang2023visual}&0.426&0.526&0.541&0.549&0.605&0.644&0.311&0.341&0.331 \\
MI-Zero \cite{lu2023visual}&0.599&0.648&0.691&0.587&0.651&0.726&0.371&0.402&0.398 \\
CONCH \cite{lu2023towards}&0.615&0.663&0.709&0.601&0.672&0.751&0.424&0.447&0.406 \\
QuiltNet \cite{ikezogwo2024quilt}&0.581&0.621&0.681&0.595&0.661&0.706&0.386&0.433&\underline{0.418} \\
CPLIP \cite{javed2024cplip}&\underline{0.687}&\underline{0.722}&\underline{0.761}&\underline{0.651}&\underline{0.715}&\underline{0.806}&\underline{0.420}&\underline{0.454}&0.413 \\
MR-PLIP&\textbf{0.706}&\textbf{0.741}&\textbf{0.785}&\textbf{0.664}&\textbf{0.745}&\textbf{0.823}&\textbf{0.459}&\textbf{0.489}&\textbf{0.436} \\
\hline
    \end{tabular}        
    \label{table_segmentation}
\end{table*}

\begin{table*}[t!]
    \centering
    \caption{\textcolor{black}{Zero-shot cross-modal retrieval (text-to-image and image-to-text) results on two datasets. In each cell, the results are displayed in the format (\%$|$\%), with text-to-image retrieval results on the left and image-to-text retrieval results on the right.}}
    \begin{tabular}{|c | c c| c c| c c|cc|cc|cc|}
    \hline
    Methods & &\multicolumn{3}{c}{ARCH}& &&&\multicolumn{3}{c}{Twitter}&&  \\ \hline
     & \multicolumn{2}{c}{R@1} & \multicolumn{2}{c}{R@50} &\multicolumn{2}{c|}{R@200} & \multicolumn{2}{c}{R@1} & \multicolumn{2}{c}{R@50} &\multicolumn{2}{c|}{R@200} \\
    CLIP&0.07&0.05&2.42&2.52&7.21&7.22&0.09&0.08&1.28&1.23&6.61&6.97\\
    BioCLIP&8.89&\underline{9.97}&53.24&52.13&71.43&68.47&9.11&10.56&40.30&39.23&52.33&51.66\\
    PLIP&0.56&0.74&43.10&42.71&29.85&29.46&2.33&2.42&52.76&53.25&62.33&64.40\\
   MI-Zero&6.87&7.71&52.10&54.14&60.96&61.21&5.77&6.98&70.21&68.83&75.66&74.10\\
   QuiltNet&8.77&9.85&55.14&53.06&77.64&73.43&7.89&8.66&69.81&70.44&73.44&72.11\\
CONCH&8.16&9.11&\underline{58.91}&\underline{59.10}&75.16&76.90&\underline{9.90}&\underline{10.19}&\underline{75.91}&\underline{76.80}&\underline{80.91}&\underline{81.68}\\
   CPLIP&\underline{9.10}&9.06&56.77&57.12&\underline{79.10}&\underline{80.19}&6.17&7.90&72.89&73.09&79.10&80.17\\
   MR-PLIP&\textbf{11.17}&\textbf{12.56}&\textbf{61.31}&\textbf{62.23}&\textbf{83.98}&\textbf{84.20}&\textbf{10.21}&\textbf{11.33}&\textbf{78.84}&\textbf{76.91}&\textbf{83.21}&\textbf{82.97}\\  
    \hline
   \end{tabular}
    \label{table_crossmodal}
\end{table*}

\begin{table*}[t!]
\centering
\caption{Performance comparison of the proposed MR-PLIP with existing SOTA foundation models, including both VLMs and vision-only models. The tile-level classification performance is reported using linear probe evaluations, while WSI-level classification results are reported using weakly supervised learning in which the ABMIL method is employed for both feature aggregation and MIL classification.
BA represents balanced accuracy and $F_{1}$ is the weighted $F_{1}$ score.}
\makebox[\linewidth]{
\scalebox{0.70}{
\begin{tabular}{ c| c c| c c| c c| c c |c c |c c|cc|cc|cc|}
\hline
\textbf{Datasets} & \multicolumn{2}{c|}{CONCH} & \multicolumn{2}{c|}{QuiltNet} & \multicolumn{2}{c|}{UNI}& \multicolumn{2}{c|}{REMEDIS}&\multicolumn{2}{c|}{Virchow} &\multicolumn{2}{c|}{CHIEF}&\multicolumn{2}{c|}{CTransPath}&\multicolumn{2}{c|}{GigaPath}&\multicolumn{2}{c|}{MR-PLIP}\\
(Tile-level)& BA & $F_{1}$ & BA & $F_{1}$ & BA & $F_{1}$ & BA & $F_{1}$ & BA & $F_{1}$& BA & $F_{1}$&BA&$F_{1}$&BA&$F_{1}$&BA&$F_{1}$\\
\hline
NCT-CRC&0.938&0.955&0.922&0.947&0.874&0.875&0.787&0.802&\underline{0.960}&\underline{0.968}&0.844&0.856&0.845&0.867&0.929&0.942&\textbf{0.965}&\textbf{0.976}\\
PatchCamelyon&0.866&0.869&0.822&0.831&0.901&0.930&0.805&0.822&\underline{0.933}&\underline{0.933}&0.833&0.851&0.911&0.935&0.925&0.931&\textbf{0.955}&\textbf{0.961}\\
WILDS-CAM17&0.911&0.925&0.861&0.877&\textbf{0.983}&\textbf{0.983}&0.926&0.926&0.971&0.971&0.901&0.922&0.960&0.960&0.951&0.962&\underline{0.975}&\underline{0.980}\\
MHIST&0.791&0.807&0.802&0.823&\underline{0.856}&\underline{0.881}&0.781&0.807&0.831&0.836&0.791&0.813&0.811&0.826&0.851&0.879&\textbf{0.876}&\textbf{0.915}\\
SICAP&0.711&0.745&0.722&0.767&0.826&0.841&0.806&0.811&\underline{0.855}&\underline{0.873}&0.771&0.783&0.678&0.747&0.845&0.861&\textbf{0.886}&\textbf{0.905}\\
WSSS4LUAD&0.811&0.825&0.805&0.812&0.831&0.835&0.769&0.782&\underline{0.866}&\underline{0.873}&0.812&0.828&0.844&0.857&0.860&0.872&\textbf{0.887}&\textbf{0.896}\\
BACH&0.856&0.871&0.833&0.861&0.925&0.926&0.863&0.864&0.915&0.920&0.847&0.863&0.875&0.872&\underline{0.933}&\underline{0.947}&\textbf{0.945}&\textbf{0.966}\\
UniToPatho&0.451&0.467&0.446&0.457&0.504&0.533&0.446&0.473&\underline{0.557}&\underline{0.574}&0.405&0.416&0.432&0.481&0.535&0.540&\textbf{0.605}&\textbf{0.622}\\
\hline
\textbf{Datasets} & \multicolumn{2}{c|}{CONCH} & \multicolumn{2}{c|}{QuiltNet} & \multicolumn{2}{c|}{UNI}& \multicolumn{2}{c|}{REMEDIS}&\multicolumn{2}{c|}{Virchow} &\multicolumn{2}{c|}{CHIEF}&\multicolumn{2}{c|}{CTransPath}&\multicolumn{2}{c|}{GigaPath}&\multicolumn{2}{c|}{MR-PLIP}\\    
(WSI-level)& BA & $F_{1}$ & BA & $F_{1}$ & BA & $F_{1}$ & BA & $F_{1}$ & BA & $F_{1}$& BA & $F_{1}$&BA&$F_{1}$&BA&$F_{1}$&BA&$F_{1}$\\
\hline
CAM16&0.881&0.902&0.902&0.922&\underline{0.957}&0.961&0.930&0.923&0.951&0.913&0.944&0.952&0.897&0.907&\textbf{0.967}&\underline{0.960}&0.950&\textbf{0.966}\\
RCC-DHMC&0.856&0.866&0.851&0.864&0.919&0.926&0.865&0.877&\underline{0.922}&0.931&0.897&0.901&0.804&0.883&0.921&\underline{0.936}&\textbf{0.941}&\textbf{0.952}\\
HunCRC&\underline{0.681}&0.721&0.702&0.722&0.643&\textbf{0.824}&0.604&\underline{0.787}&0.621&0.667&0.651&0.667&0.556&0.728&0.641&0.667&\textbf{0.688}&0.701\\
BRCA-BRACS&\underline{0.723}&\underline{0.748}&0.718&0.725&0.687&0.691&0.676&0.696&0.708&0.722&0.656&0.666&0.639&0.648&0.704&0.715&\textbf{0.741}&\textbf{0.767}\\
PANDA&0.702&0.733&0.722&0.744&\underline{0.757}&\underline{0.809}&0.711&0.766&0.728&0.741&0.724&0.745&0.691&0.752&0.744&0.789&\textbf{0.786}&\textbf{0.816}\\
EBRAINS&0.687&0.717&0.655&0.666&0.675&\underline{0.746}&0.382&0.471&\underline{0.701}&0.723&0.688&0.706&0.514&0.597&0.687&0.704&\textbf{0.745}&\textbf{0.763}\\
NSCLC-CPTAC&0.881&0.902&0.877&0.900&0.904&0.935&0.841&0.866&\underline{0.923}&\underline{0.936}&0.922&0.934&0.877&0.895&0.900&0.915&\textbf{0.930}&\textbf{0.955}\\
\hline     
\end{tabular}
}}
\label{table1}
\end{table*}

\section{Zero-Shot Experiments}
\label{zero}
Zero-shot learning refers to the capability of models to accurately perform tasks on new, unseen data without direct training on those specific tasks, utilizing pre-learned representations from image-text pairs.
\subsection{Zero-shot Segmentation Results (Table~\ref{table_segmentation})}
We conduct zero-shot WSI-level segmentation using a process akin to the one for tile-based classification previously mentioned. Rather than compiling scores from tiles into a singular WSI-level prediction, we map tile-level scores back to their respective spatial locations within the WSI, averaging scores in overlapping areas. The highest-scoring class at each location is used to determine the pixel-level segmentation mask. 
Table~\ref{table_segmentation} presents the zero-shot WSI-level segmentation outcomes on three datasets, setting them against six SOTA methods. 
Our MR-PLIP \textcolor{black}{model} outperforms all other methods in terms of performance across all three datasets.
Overall, \textcolor{black}{CPLIP} secures its position as the runner-up in performance on the DigestPath and SICAP datasets, while QuiltNet ranks as the second-best on the TIGER dataset.

\subsection{Zero-shot Cross-modal Retrieval Results (Table~\ref{table_crossmodal})}
The zero-shot text-to-image and image-to-text retrieval tasks are evaluated by locating the closest matches for each modality and verifying whether the correct ground-truth pair falls within the top {1, 50, 200} closest matches. 
Table~\ref{table_crossmodal} shows the zero-shot cross-modal retrieval performance on two separate datasets, alongside a comparison with five SOTA VLMs in CPath. On both datasets, \textcolor{black}{the MR-PLIP model} outperforms all other methods by a significant margin, indicating its robust ability to align cross-resolution features across diverse textual and visual domains. \textcolor{black}{CONCH and CPLIP} also show strong performance, ranking as the second-best in terms of recall metrics.

\section{Linear Probe Experiments (Table \ref{table1})}
\label{probe}
In the context of deep learning, linear probing refers to a technique used to evaluate the quality of features learned by a deep neural network. 
Specifically, it involves training a simple linear classifier (e.g., logistic regression) on top of the features extracted from a pre-trained neural network. 
This process is performed without fine-tuning the original network's weights; only the weights of the linear classifier are updated during the training process.
The primary goal of linear probing is to assess how well the pre-trained network has captured valuable data representations. 
If a simple linear classifier can achieve high performance using the features extracted by the neural network, it suggests that the network has learned a rich and informative representation of the data.

Following the SOTA VLMs in CPath, we conducted a downstream analysis by freezing the weights of our proposed model and subsequently training linear layers for supervised classification tasks. We obtained text-guided visual features by inputting an image alongside its most closely matching text description from a predefined prompt set. A downstream linear classifier was then trained on these features for tile-level assessments to evaluate the quality of the representations learned by our MR-PLIP model. 

\begin{table}[t!]
\centering
\caption{WSI-level segmentation performance comparison in terms of dice score, precision, and recall of the MR-PLIP with other SOTA methods using \textbf{linear probe} evaluation protocols on three datasets.}
\makebox[\linewidth]{
\scalebox{0.60}{
\begin{tabular}{ c| c| c| c| c| c|c|c|c|c|c|}
\hline
\textbf{Datasets} & \multicolumn{3}{c|}{DigestPath} & \multicolumn{3}{c|}{SICAP} &\multicolumn{3}{c|}{TIGER}\\    
\hline     
PLIP&0.426&0.526&0.541&0.549&0.605&0.644&0.463&0.478&0.493\\
QuiltNet&0.521&0.545&0.564&0.592&0.603&0.621&0.602&0.619&0.627\\
DINOSSLPath&0.551&0.588&0.603&0.634&0.667&0.683&0.601&0.628&0.636\\
CTransPath&0.503&0.516&0.526&0.534&0.567&0.582&0.533&0.561&0.587\\
CONCH&0.615&0.663&0.709&0.601&\underline{0.672}&\textbf{0.751}&0.433&0.457&0.461 \\
UNI&0.804&0.811&0.826&\underline{0.645}&0.662&0.603&0.687&0.702&0.724 \\
Virchow&\underline{0.833}&\underline{0.865}&\underline{0.889}&0.641&0.652&0.687&\underline{0.707}&\textbf{0.732}&0.755 \\
MR-PLIP&\textbf{0.851}&\textbf{0.873}&\textbf{0.903}&\textbf{0.655}&\textbf{0.688}&\underline{0.708}&\textbf{0.726}&\underline{0.730}&\textbf{0.778}\\
\hline
\end{tabular}
}}
\label{table2}
\end{table}

\begin{table}[t!]
\centering
\caption{Nuclei segmentation results in terms of mPQ measure of the MR-PLIP with other SOTA methods using both \textbf{linear probing} and \textbf{full fine-tuning} evaluation protocols on two datasets}.
\begin{tabular}{ c| c c| c c|}
\hline
\textbf{Datasets} & \multicolumn{2}{c|}{CONSEP} & \multicolumn{2}{c|}{PanNuKe}\\    
(Segmentation)& Linear & Fine-tune & Linear & Fine-tune\\
\hline     
QuiltNet&41.11&45.56&48.11&52.21\\
MI-Zero&40.91&43.33&\underline{49.54}&\underline{54.90}\\
CONCH&39.85&\underline{51.75}&44.31&50.21\\
DinoSSLPath&\underline{42.71}&46.70&48.88&54.41\\
MR-PLIP&\textbf{46.61}&\textbf{52.25}&\textbf{51.64}&\textbf{58.61}\\
\hline
\end{tabular}
\label{table3}
\end{table}

\noindent Tables \ref{table1}-\ref{table3} displays the results of linear probe evaluations on three downstream histopathology tasks including tile-level classification, WSI-level segmentation, and nuclei segmentation across 21 datasets, evaluating the weighted average $F_{1}$ score and comparing it against SOTA CPath models including PLIP \cite{huang2023visual}, MI-Zero \cite{lu2023visual}, QuiltNet \cite{ikezogwo2024quilt}, CTransPath \cite{wang2022transformer}, and DINOSSLPath \cite{kang2023benchmarking}.
Across all datasets, our MR-PLIP algorithm consistently outperforms other methods by a substantial margin, affirming the advanced performance of the MR-PLIP model over the second-best performing methods GigaPath, Virchow, CONCH, and UNI.

\section{Weakly-Supervised WSI Classification Results (Table \ref{table1})}
\label{weak}
We performed weakly-supervised WSI classification to evaluate the text-guided visual representations learned by MR-PLIP across seven diverse WSI classification datasets. 
MR-PLIP was used to extract text-guided visual features from each patch, after which the ABMIL method \cite{ilse2018attention} was employed for feature aggregation and MIL-based classification, as done in other SOTA methods \cite{wang2022transformer, wang2024pathology, azizi2023robust, vorontsov2024foundation, chen2024towards}.
For training, we used the AdamW optimizer with a cosine learning rate scheduler, a learning rate of $1 \times 10^{-4}$, cross-entropy loss, and a maximum of 20 epochs. 
To ensure fair comparisons, we followed the experimental protocols of existing SOTA methods for WSI classification tasks \cite{chen2024towards}.
If official data folds were not available, the WSI datasets were case-stratified and label-stratified into train-validation-test splits as suggested by UNI \cite{chen2024towards}. 

Table \ref{table1} compares MR-PLIP with SOTA foundation models based on balanced accuracy and $F_{1}$ score. 
Our results show that MR-PLIP outperforms existing models by a significant margin, highlighting the benefits of explicitly incorporating multi-resolution image-text features.

\begin{table*}[t!]
\centering
\caption{\textbf{Weakly-Supervised WSI Classification Results Compared with SOTA Methods}. 
Accuracy (Acc), weighted average $F_{1}$, and AUC scores for weakly-supervised WSI classification across multiple datasets. 
The proposed MR-PLIP model achieves the highest performance across most metrics, demonstrating its effectiveness over other SOTA methods.}
\scalebox{0.90}{
\begin{tabular}{ |c|ccc|ccc|ccc|}
\hline
Methods&\multicolumn{3}{c|}{CAM16}&\multicolumn{3}{c|}{NSCLC}&\multicolumn{3}{c|}{PANDA}\\
&Acc&$F_{1}$&AUC&Acc&$F_{1}$&AUC&Acc&$F_{1}$&AUC\\
\cline{2-10}
PANTHER \hfill(CVPR24)\cite{song2024morphological}&0.933&0.949&\underline{0.981}&\underline{0.881}&0.867&0.906&0.751&0.796&\underline{0.946}\\
R$^{2}$T \hfill(CVPR24)\cite{tang2024feature}&\underline{0.954}&0.942&0.980&0.856&\underline{0.883}&0.901&\underline{0.768}&\underline{0.806}&0.903\\
SI-MIL \hfill(CVPR24)\cite{kapse2024si}&0.944&\underline{0.951}&0.963&0.844&0.830&0.881&0.733&0.771&0.886\\
ViLa-MIL \hfill(CVPR24)\cite{shi2024vila}&0.913&0.928&0.966&0.821&0.829&0.876&0.750&0.761&0.854\\
FiVE \hfill(CVPR24)\cite{li2024generalizable}&0.942&0.932&0.975&0.842&0.853&\underline{0.935}&0.743&0.788&0.873\\
MR-PLIP &\textbf{0.950}&\textbf{0.966}&\textbf{0.993}&\textbf{0.930}&\textbf{0.955}&\textbf{0.964}&\textbf{0.786}&\textbf{0.816}&\textbf{0.977}\\
\hline
\end{tabular}}     
\label{table_new}
\end{table*}

\section{Fine-tune Evaluation (Table \ref{table3})}
\label{fine}
In the context of deep learning, fine-tuning refers to a technique used to assess the adaptability and transfer potential of the learned weights by a deep neural network. 
Specifically, it involves fine-tuning the original network's weights. 
The primary goal of full fine-tuning is to assess how well the network weighted are transferred to the downstream analysis task. 

We performed full fine-tuning of our model in conjunction with the linear layers for classification. 
This approach assesses the adaptability and transfer potential of the learned weights within the MR-PLIP framework. 

Tables \ref{table3} display the results of full fine-tuning across nuclei segmentation datasets, evaluating the mPQ scores and comparing them against the SOTA methods.
 Across all datasets, our MR-PLIP algorithm consistently outperforms other methods by a substantial margin, affirming the advanced performance of the MR-PLIP model over the second-best performing methods QuiltNet, MI-Zero, and DINOSSLPath.

\section{More Comparisons with SOTA MIL-based Methods (Table \ref{table_new})}
\label{more}
In this experiment, we compared the performance of our MR-PLIP model with recently proposed MIL-based methods including FiVE \cite{li2024generalizable}, R$^{2}$T \cite{tang2024feature}, SI-MIL \cite{kapse2024si}, ViLa-MIL \cite{shi2024vila}, and PANTHER \cite{song2024morphological}.
For a fair comparison, we employed the same ABMIL method for WSI-level feature aggregation and classification as discussed in Sec. \ref{weak}. 

Table \ref{table_new} shows the results of the weakly supervised WSI classification and comparison with recently proposed SOTA methods on three tumor subtype classification datasets including CAM16, NSCL, and PANDA.
The performance is reported in terms of balanced accuracy, $F_{1}$ score, and AUC. 
The MR-PLIP model consistently achieved superior performance across most evaluation metrics.

\end{document}